\documentclass[10pt,twocolumn,letterpaper]{article}

\usepackage[pagenumbers]{cvpr} %
\usepackage[accsupp]{axessibility}

\usepackage{microtype}
\makeatletter
\renewcommand\paragraph{\@startsection{paragraph}{4}{\z@}%
	{0.75ex \@plus.5ex \@minus.2ex}%
	{-1em}%
	{\normalfont\normalsize\bfseries\maybe@addperiod}}
\newcommand{\maybe@addperiod}[1]{#1\@addpunct{.}}
\makeatother

\usepackage[dvipsnames]{xcolor}
\usepackage{tabularx}
\usepackage{multirow}
\usepackage{makecell}
\usepackage{pifont}
\usepackage{ragged2e}
\newcolumntype{Y}{>{\centering\arraybackslash}X}

\newcommand{\delete}[1]{}

\newcommand{\norm}[1]{\left\lVert#1\right\rVert}

\newcommand{\col}{{\mathbf{c}}}
\newcommand{\dir}{{\mathbf{d}}}
\newcommand{\feat}{{\mathbf{\boldsymbol\varphi}}}
\newcommand{\normal}{{\mathbf{n}}}
\newcommand{\param}{{\boldsymbol\theta}}
\newcommand{\pos}{{\mathbf{x}}}
\newcommand{\quat}{{\mathbf{q}}}
\newcommand{\rough}{{\rho}}
\newcommand{\tint}{{\mathbf{s}}}
\newcommand{\origin}{{\mathbf{o}}}
\newcommand{\grad}{\mathbf{g}}
\newcommand{\eps}{\boldsymbol\epsilon}
\newcommand{\mean}{{\mathbf{\boldsymbol\mu}}}
\newcommand{\scale}{{\mathbf{\boldsymbol\sigma}}}
\newcommand{\invscale}{{\mathbf{\boldsymbol\psi}}}
\newcommand{\encoding}{encoding\xspace}
\newcommand{\Encoding}{Encoding\xspace}
\DeclareMathOperator{\sign}{sign}

\definecolor{yesgreen}{rgb}{0, 0.55, 0}
\definecolor{nored}{RGB}{180,0,0}
\newcommand{\yesmark}{\textcolor{yesgreen}{\checkmark}}
\newcommand{\nomark}{\textcolor{nored}{\ding{55}}}

\definecolor{cvprblue}{rgb}{0.21,0.49,0.74}
\usepackage[pagebackref,breaklinks,colorlinks,citecolor=cvprblue]{hyperref}

\def\paperID{6769} %
\def\confName{CVPR}
\def\confYear{2024}

\title{%
SpecNeRF: Gaussian Directional \Encoding~for Specular Reflections
\vspace{-2mm}
}

\author{Li Ma\textsuperscript{1,2}\quad%
	Vasu Agrawal\textsuperscript{2}\quad%
	Haithem Turki\textsuperscript{2,3}\quad%
	Changil Kim\textsuperscript{2}\\%
	Chen Gao\textsuperscript{2}\quad%
	Pedro Sander\textsuperscript{1}\quad%
	Michael Zollhöfer\textsuperscript{2}\quad%
	Christian Richardt\textsuperscript{2}\\[0.5em]
	\textsuperscript{1}The Hong Kong University of Science and Technology\\%
	\textsuperscript{2}Meta Reality Labs\quad%
	\textsuperscript{3}Carnegie Mellon University
    \vspace{-3mm}
}

\begin{document}
\maketitle
\begin{abstract}	
Neural radiance fields have achieved remarkable performance in modeling the appearance of 3D scenes.
However, existing approaches still struggle with the view-dependent appearance of glossy surfaces, especially under complex lighting of indoor environments.
Unlike existing methods, which typically assume distant lighting like an environment map, we propose a learnable Gaussian directional encoding to better model the view-dependent effects under near-field lighting conditions.
Importantly, our new directional encoding captures the spatially-varying nature of near-field lighting and emulates the behavior of prefiltered environment maps.
As a result, it enables the efficient evaluation of preconvolved specular color at any 3D location with varying roughness coefficients.
We further introduce a data-driven geometry prior that helps alleviate the shape radiance ambiguity in reflection modeling.
We show that our Gaussian directional encoding and geometry prior significantly improve the modeling of challenging specular reflections in neural radiance fields, which helps decompose appearance into more physically meaningful components.
\end{abstract}

\section{Introduction}
\label{sec:introduction}

\begin{figure}
    \centering
    \includegraphics[width=\linewidth]{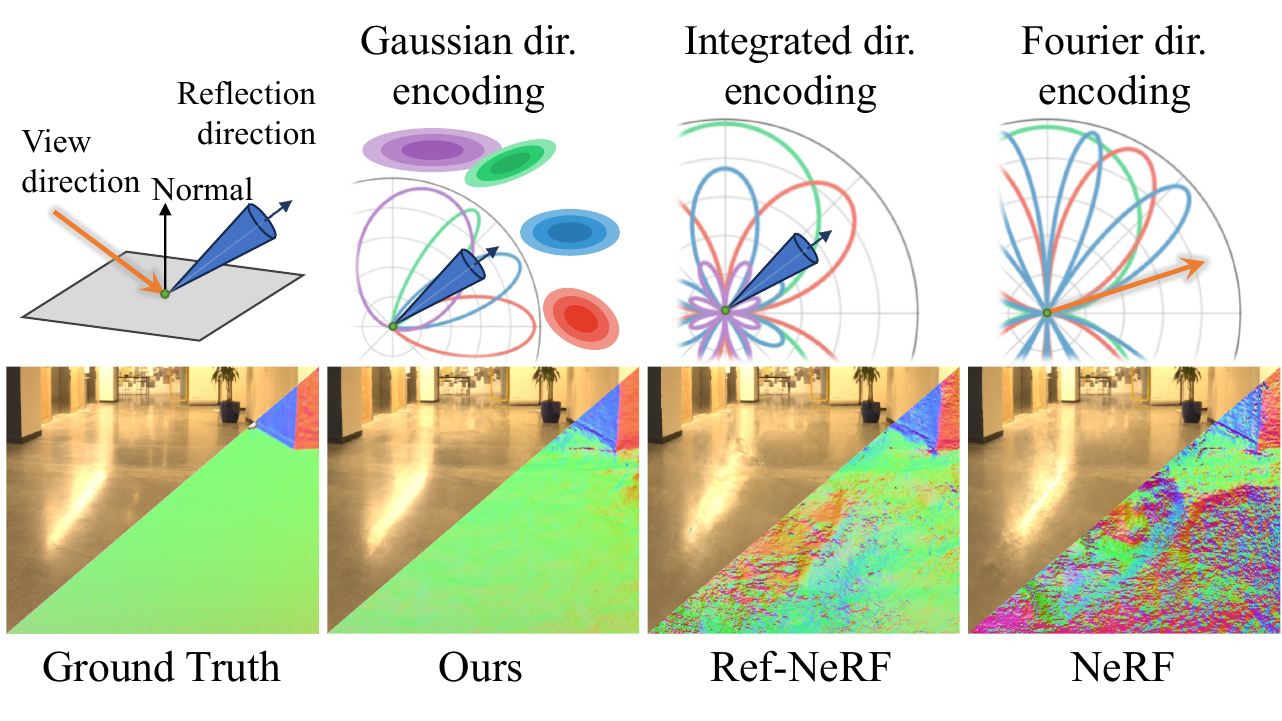}
    \vspace{-6mm}
    \caption{\label{fig:teaser}%
    	We propose a Gaussian directional \encoding that leads to better modeling of specular reflections under near-field lighting conditions.
    	In contrast, the integrated directional encoding utilized in Ref-NeRF \cite{VerbiHMZBS2022} and Fourier directional encoding in NeRF \cite{MildeSTBRN2020} exhibit suboptimal performance under similar conditions.}
    \vspace{-4mm}
\end{figure}

Neural radiance fields (NeRFs) have emerged as a popular scene representation for novel-view synthesis \cite{MildeSTBRN2020,TewarTMSTWLSMLSTNBWZG2022,XieTSLYKTTSS2022}.
By training a neural network based on sparse observations of a 3D scene, NeRF-like representations are able to synthesize novel views with photorealistic visual quality. 
In particular, with a scalable model design, such as InstantNGP \cite{MuelleESK2022}, NeRFs are able to model room-scale 3D scenes with extraordinary detail \cite{XuALGBKRPKBLZR2023}.
However, existing approaches typically only manage to model mild view-dependent effects like those seen on nearly diffuse surfaces.
When encountering highly view-dependent glossy surfaces, NeRFs struggle to model the high-frequency changes when the viewpoint changes.
Instead, they tend to ``fake'' specular reflections by placing them behind surfaces, which may result in poor view interpolation and ``foggy'' geometry \cite{VerbiHMZBS2022}.
Moreover, fake reflections are not viable if one can look behind the surface, as NeRF can no longer hide the reflections there.

Accurately modeling and reconstructing specular reflections presents notable challenges, especially for room-scale scenes.
Physically correct reflection modeling involves path-tracing many rays for every single pixel, which is impractical for NeRF-like volumetric scene representations, primarily due to the large computational requirements to shade a single pixel.
Consequently, an efficient approximation of the reflection shading is needed for a feasible modeling of reflections. 
Existing works \cite{VerbiHMZBS2022,GeHZLC2023} address this challenge by incorporating heuristic modules inspired by real-time image-based lighting (IBL) \cite{IBL_sig84} techniques, such as explicit ray bounce computations to enhance NeRF's capability to simulate reflections, and integrated directional encoding to simulate appearance change under varying surface roughness.

While these improvements have shown to be effective in modeling specular reflections for NeRFs, they are limited to object-level reconstruction under distant lighting, which assumes the object is lit by a 2D environment map.
They work poorly for modeling near-field lighting, where the corresponding environment map varies spatially.
The issue is that existing methods rely on directional encodings to embed ray directions for generating view-dependent reflections. 
These encodings, such as Fourier encoding or spherical harmonics, are spatially invariant.
\Cref{fig:teaser} demonstrates one example of NeRF \cite{MildeSTBRN2020} and Ref-NeRF \cite{VerbiHMZBS2022} reconstructions of an indoor scene with spatially-varying lighting.
NeRF produces extremely noisy geometry, resulting in artifacts in the rendering result.
Ref-NeRF offers a slight improvement, but still struggles with noisy geometry and view interpolation.
This illustrates that the spatial invariance in the directional encodings of existing methods presents challenges under spatially-varying lighting conditions.

In this work, we propose a novel Gaussian directional \encoding\ that is tailored for spatially varying lighting conditions.
Instead of only encoding a 2D ray direction, we use a set of learnable 3D Gaussians as the basis to embed a 5D ray space including both ray origin and ray direction.
We show that, with appropriately optimized Gaussian parameters, this encoding introduces an important inductive bias towards near-field lighting, which enhances the model's ability to capture the characteristics of specular surfaces, leading to photorealistic reconstructions of shiny reflections.
We further demonstrate that by changing the scale of the 3D Gaussians, we can edit the apparent roughness of a surface.

While our proposed Gaussian directional encoding improves the reflection modeling of NeRF, high-quality reflection reconstruction also requires an accurate surface geometry and normal in order to compute accurate reflection rays.
However, the geometry within NeRFs is often noisy in the early phases of training, which presents challenges in simultaneously optimizing for good geometry and reflections.
To better address this challenge, we introduce a data-driven prior to direct the NeRF model towards the desired solution.
We deploy a monocular normal estimation network to supervise the normal of the geometry at the beginning of the training stage, and show that this bootstrapping strategy improves the reconstruction of normals, and further leads to successful modeling of specular reflections.
We conduct experiments on several public datasets and show that the proposed method outperforms existing methods, achieving higher-quality photorealistic rendering of reflective scenes while also providing more meaningful and accurate color component decomposition.
Our contributions can be summarized as follows:
\begin{itemize}
    \item We propose a novel Gaussian directional encoding that is more effective in modeling view-dependent effects under near-field lighting conditions.
    
    \item We propose to use monocular normal estimation to resolve shape-radiance ambiguity in the early training stages.
    
    \item Our full NeRF pipeline achieves state-of-the-art novel-view synthesis performance for specular reflections.
\end{itemize}

\section{Related Work}
\label{sec:relatedwork}

\begin{figure*}
    \centering
    \includegraphics[width=\textwidth]{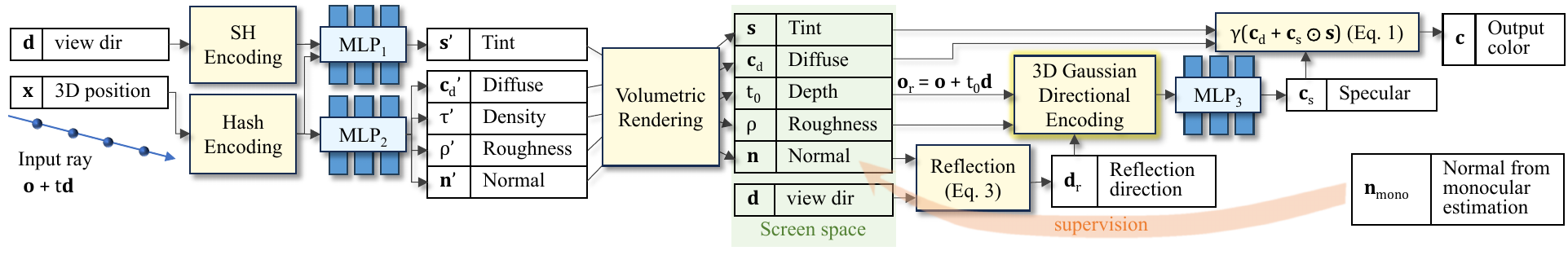}
    \vspace{-6mm}
    \caption{An overview of our
        model.
        The key enabler for specular reflections is our novel 3D Gaussian directional \encoding module that converts the reflected ray into a spatially-varying embedding, which is further decoded into specular color.
    }
    \vspace{-5mm}
    \label{fig:overview}
\end{figure*}

\paragraph{Reflection-aware NeRFs}

Successfully modeling view-dependent effects, such as specular reflections, can greatly enhance the photorealism of the reconstructed NeRF.
NeRF models view-dependency by conditioning the radiance on the positional encoding \cite{TanciSMFRSRBN2020} of the input ray direction, which is only capable of mild view-dependent effects.
Ref-NeRF \cite{VerbiHMZBS2022} improves NeRF's capability for modeling reflections by conditioning the view-dependent appearance on the reflection ray direction instead of incident ray direction, and by modulating the directional encoding based on surface roughness.
This reparameterization of outgoing radiance makes the underlying scene function simpler, leading to a better geometry and view interpolation quality for glossy objects.
Ref-NeuS \cite{GeHZLC2023} further extends these concepts to a surface-based representation.
However, these are primarily designed for object-level reconstruction under environment map lighting conditions.
Modeling large-scale scenes with near-field lighting remains a problem.
Clean-NeRF \cite{LiuTT2023} decomposes the radiance into diffuse and specular colors, and supervises the two components by least-square estimations of multiple input rays.
This alleviates the ambiguity of highly specular regions;
yet, it does not change the view-dependent structure of the NeRF model, thus limiting its ability to model reflections.
NeRF-DS \cite{YanLL2023} models specularities in dynamic scenes and considers the variations in reflections caused by dynamic geometry through the use of a dynamic normal field, but requires additional object masks for accurate specular reconstruction.

\paragraph{NeRF-based Inverse Rendering}

Inverse rendering goes beyond simple reflection modeling and aims to jointly recover one or more of scene geometry, material appearance and the lighting condition.
In practice, the material appearance is typically modeled using physically-based rendering assets such as albedo, roughness and glossiness.
Mesh-based inverse rendering methods \cite{AzinoLKN2019,PhiliMGD2021,RT_FIPT,ZhuanZWZFLSC2024} try to recover materials using differentiable path tracing \cite{LiADL2018}.
However, they typically assume a given geometry, since optimizing mesh geometry is challenging.
On the contrary, NeRF-based inverse rendering approaches \cite{BiXSMSHHKR2020,SriniDZTMB2021,ZhangSDDFB2021,INV_invrender} make it easier to optimize geometry jointly by modeling material properties and density continuously in a volumetric 3D space.
The lighting is usually represented as point or directional lights \cite{BiXSMSHHKR2020,ZengCDPWT2023,LiL2022},
an environment texture map \cite{SriniDZTMB2021,LiuWLLWLKW2023,ZhangSDDFB2021,INV_NRTF,MaiVKF2023}, or an implicit texture map modeled by spherical Gaussians \cite{ZhangLWBS2021,INV_invrender,DengLLY2024,ZhangXYYWJYY2023,BossBJBLL2021,INV_invrender,INV_TensoIR} or MLPs \cite{BossJBLBL2021,LiangZLYGV2022}.
Most methods are limited to object-level reconstruction and assume the lighting is spatially invariant (i.e. distant).
Several light estimation techniques \cite{lighthouse,indoorlight,lightediting,svlight} explore using 3D light primitives or spatially-varying spherical Gaussians to model spatially varying lighting.
However, these methods focus on data-driven approaches to estimate lighting for image editing.
NeILF \cite{YaoZLQFMTQ2022} and NeILF++ \cite{ZhangYLLFMTQ2023} model lighting as a 5D light field using another MLP, but still focus mainly on small-scale reconstruction.
Several works apply inverse rendering for relighting outdoor scenes \cite{INV_Outdoor_MILO,WangSGHMHGCF2023,INV_Outdoor_NeRFRelit,LiGFLG2022}.
However, they focus more on diffuse materials with correct shadow modeling instead of reflections.
In this work, we have a different goal compared to inverse rendering, focusing only on correctly modeling reflections for better novel-view synthesis, rather than trying to discern material properties for standalone use.

\paragraph{NeRF with mirror reflections}

One special case of reflection is mirror reflection.
One approach represents the reflected scene as a separate NeRF \cite{GuoKBHZ2022}, and composites the two NeRF results in image space.
This is also deployed in image-based rendering \cite{XuWZHYBX2021} and large-scale NeRF reconstruction \cite{WuXZBHTX2022}.
Given a multi-mirror scene, the idea can be further extended to multi-space NeRFs \cite{YinQCR2023}.
An alternate approach is to explicitly model the mirror geometry, and to render the mirrored scene by path tracing \cite{HollaBMSK2023,ZengBCDZBC2023}.
However, since estimating the mirror geometry is highly ill-posed, manual annotation is usually needed.
Curved reflectors need even more careful handling \cite{KopanLRJD2022,TiwarDBKVR2023}.

\section{Preliminaries}
\label{sec:preliminary}

We first review Ref-NeRF \cite{VerbiHMZBS2022} for decomposing view-dependent appearance.
Similar to NeRF, Ref-NeRF models the scene as a function that maps the position $\pos$ and view direction $\dir$ to the final color $\col$ and density $\tau$.
The difference is that Ref-NeRF predicts the color as a combination of diffuse color $\col_\text{d}$ and specular color $\col_\text{s}$:
\begin{align}
    \col = \gamma(\col_\text{d} + \col_\text{s} \odot \tint) \text{,}
    \label{eq:composite}
\end{align}
where $\tint$ is the specular tint, `$\odot$' the element-wise product, and $\gamma(\cdot)$ a tone-mapping function.
To predict the specular color $\col_\text{s}$, Ref-NeRF first predicts the surface normal $\normal$, roughness $\rough$, and features $\feat$ at location~$\mathbf{x}$ using an MLP.
Then, the specular color $\col_\text{s}$ is parameterized as a function of the reflection direction $\dir_\text{r}$:
\begin{align}
    \col_\text{s} = F_{\param}(\lambda_\text{IDE}(\dir_\text{r}, \rough), \feat) \text{,}
\end{align}
where $\lambda_\text{IDE}(\cdot)$ is the integrated directional encoding introduced by Ref-NeRF, $F_{\param}(\cdot)$ represents an MLP with parameters $\param$, and the reflection direction $\dir_\text{r}$ is the input direction $\dir$ reflected at the predicted surface normal $\normal$:
\begin{align}
    \dir_\text{r} = \dir - 2(\dir \cdot \normal) \normal \text{.}
    \label{eq:reflection}
\end{align}
By conditioning the specular color on reflection direction and roughness, the function $F_{\param}$ needs to fit is much simpler.

\section{Method}

Our goal is to enhance NeRF's capabilities for modeling specular reflections under near-field lighting conditions.
\Cref{fig:overview} presents an overview of our pipeline.
A key contribution is the 3D Gaussian directional \encoding that maps a ray and surface roughness to a ray embedding.

To render a pixel, we sample points along an input ray $\origin + t\dir$, and predict volume density $\tau'$, diffuse color $\col_\text{d}'$, tint $\tint'$, roughness $\rough'$, and normal direction $\normal'$ at each sample point
(we denote per-sample properties using a prime).
Given that reflections occur only at the surface, we evaluate the specular component once per ray on the surface obtained from the NeRF depth.
This also results in less computation than per-sample-point specular shading.
Consequently, we calculate volumetric depth $t_0$ by rendering the ray marching distance at each sample point.
We also volumetrically render all attributes to synthesize screen-space attributes $(\col_\text{d}, \tint, \rough, \normal)$.
Note that the rendered normal must be normalized to yield the final screen-space normal $\normal$.
We then evaluate the specular component by first computing the reflected ray using origin $\origin_\text{r} = \origin + t_0 \dir$, and the reflection direction $\dir_\text{r}$ derived using \cref{eq:reflection}.
The reflected ray $\origin_\text{r} + t \dir_\text{r}$ and surface roughness $\rough$ are then encoded using our novel 3D Gaussian directional encoding.
After a tiny MLP, we compute the specular color $\col_\text{s}$, and the final rendering result using \cref{eq:composite}.

From a physically based rendering perspective, \cref{eq:composite} is analogous to the Cook--Torrance approximation \cite{cooktorrance} of the rendering equation \cite{Kajiy1986}.
The term $\col_\text{s} \!\odot\! \tint$ can be interpreted as the split-sum approximation of the specular part of the Cook--Torrance model, with the specular color $\col_\text{s}$ corresponding to the preconvolved incident light, and the tint $\tint$ to the pre-integrated bidirectional reflectance distribution function (BRDF).

\begin{figure}
    \centering
    \includegraphics[width=\linewidth]{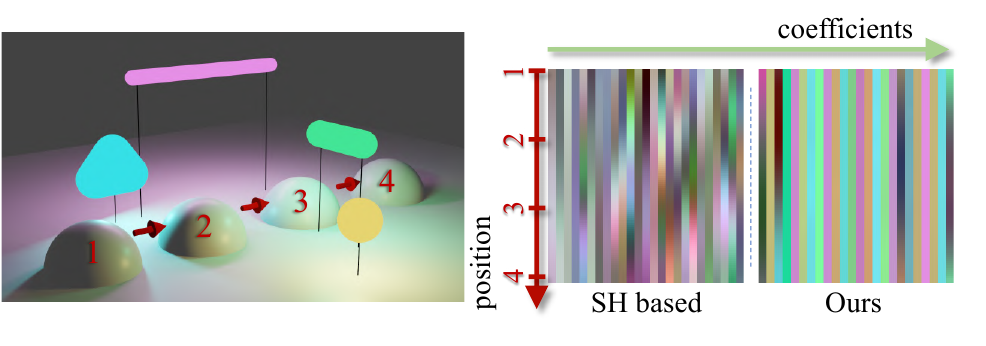}
    \vspace{-6mm}
    \caption{\label{fig:toy1}%
    	\textbf{Toy example of 3D Gaussian \encoding.}
    	\textbf{Left:}
    	A hemisphere probe translates underneath 4 lights along positions numbered 1 to 4.
            Note that we dilate the lights for better visualization.
    	\textbf{Right:}
		Representation of the probe's specular components using spherical harmonics and our 3D Gaussian directional \encoding.
        The SH encoding shows a more complex pattern under position change, while ours has spatially largely invariant coefficients.
        This suggests a simpler function for the specular prediction MLP to fit using Gaussian directional encoding.
  }
  \vspace{-4mm}
\end{figure}

\subsection{Gaussian Directional \Encoding}

Existing works parameterize view-dependent appearance by first encoding view or reflection direction into Fourier or spherical harmonics (SH) features, which results in a spatially invariant encoding of the view direction.
Therefore, it becomes challenging for the NeRF to model spatially varying view-dependent effects, such as near-field lighting.
We illustrate this via a toy example in \cref{fig:toy1}, where we place a hemispherical specular probe in a simple scene with four lights of different shapes and colors.
Then, we represent the specular component of the toy example by linearly combining the directional encoding features.
We find the optimal coefficients for each encoding type that best fit the ground-truth specular component using stochastic gradient descent, and visualize them in \cref{fig:toy1}.
We can see that even for this simple toy setup, the SH-based encoding requires complex, spatially varying coefficients, which complicates the underlying function for the NeRF to fit and interpolate.

We propose to spatially vary the encoding function by defining the basis functions via several learnable 3D Gaussians.
Specifically, we parameterize 3D Gaussians using their
position $\boldsymbol\mean_i \!\in\! \mathbb{R}^3$, scale $\scale_i \!\in\! \mathbb{R}^3$, and quaternion rotation $\quat_i \!\in\! \mathbb{H}$: %
\begin{equation}
    \mathcal{G}_i(\pos) = \exp \!\left( -\norm{
    	\mathcal{Q}(\pos - \mean_i; \quat_i) \odot \scale_i^{-1}
    }^2_2 \right) \text{,}
\end{equation}
where $\mathcal{Q}(\mathbf{v}; \quat_i)$ represents applying quaternion rotation $\quat_i$ to the vector $\mathbf{v}$.
To compute the $i$-th dimension of the encoding for a ray $\origin + t \dir$, we need to define a basis function $\mathcal{P}_i(\origin, \dir) \!\in\! \mathbb{R}$ that maps the ray to a scalar value given the Gaussian parameters.
While there are many ways to define the mapping, we find one that is efficient and has a closed-form solution by defining the projection as the maximum value of the Gaussian along the ray:
\begin{align}
    \mathcal{P}_i(\mathbf{o}, \mathbf{d}) = \max_{t \geq 0} \mathcal{G}_i(\mathbf{o} + t \dir) \text{.}
    \label{eq:maxG}
\end{align}
In the supplement,
we derive the closed-form solution:
\begin{align}
    \mathcal{P}_i(\mathbf{o}, \dir) = 
    \begin{cases}
	    \exp\!\left(\frac{(\mathbf{o}_i^\top \dir_i)^2}{\dir_i^\top \dir_i} - \mathbf{o}_i^\top \mathbf{o}_i\right)
	    & \mathbf{o}_i^\top \dir_i < 0 \\
	    \mathcal{G}_i(\mathbf{o})
	    & \text{otherwise,}
    \end{cases}
    \label{eq:gau_embed}
\end{align}
where $\mathbf{o}_i$ and $\dir_i$ are the ray origin and direction transformed into Gaussian-local space:
\begin{align}    
    \origin_i &= \mathcal{Q}(\origin - \mean_i; \quat_i) \odot \scale_i^{-1} \text{,} \\
    \dir_i &= \mathcal{Q}(\dir; \quat_i) \odot \scale_i^{-1} \text{.}
\end{align}
By applying \cref{eq:gau_embed} for every 3D Gaussian, we obtain a vector of projected values $\{\mathcal{P}_i\}$, which forms our final \encoding features.
Similar to existing NeRF-based representations \cite{MildeSTBRN2020,ChenXGYS2022,MuelleESK2022}, we rely on a small MLP to convert the \encoding to a specular color $\col_\text{s}$.

\begin{figure}
	\centering
	\includegraphics[width=\linewidth]{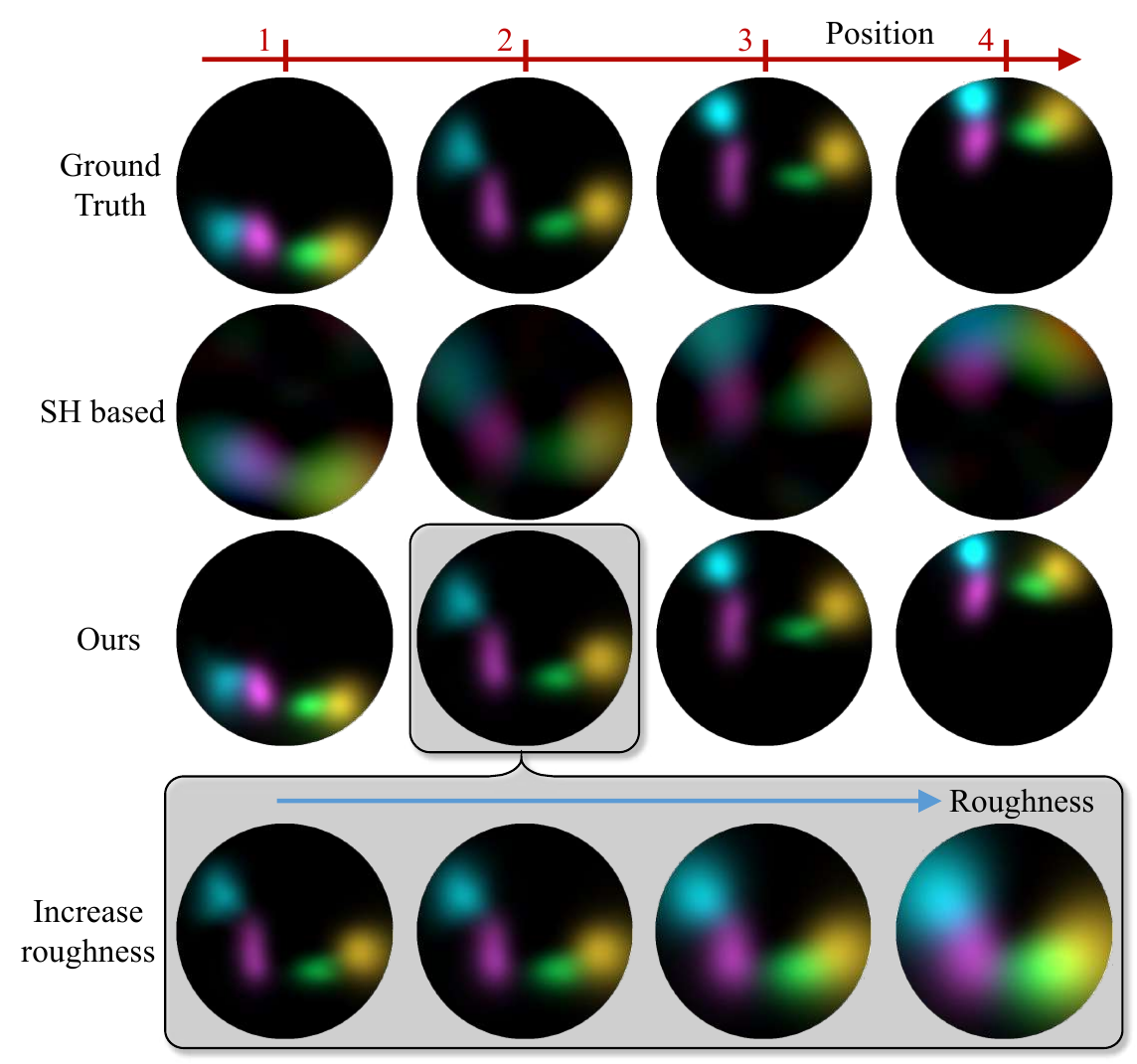}
	\vspace{-6mm}
	\caption{\label{fig:toy2}%
		Stereographic projections of the specular fitting results for the toy example in \cref{fig:toy1}.
		Both encodings produce 25 coefficients for each color channel, which are then summed to produce the final color.
		Note that the GT shows soft boundaries because it is preconvolved.
		The 3D Gaussian-based encoding demonstrates superior performance in representing the specular change with positional changes, and is also capable of smoothly varying roughness.
	}
	\vspace{-6mm}
\end{figure}

As illustrated by
the toy example in \cref{fig:toy1}, our Gaussian directional \encoding\ exhibits more constant coefficients in response to the position changes, suggesting a smoother mapping from the embedding features to the specular color.
This smoothness is due to the Gaussian basis function producing spatially varying features that mimic the behavior of how the specular component would change under near-field lighting conditions.
As a result, the underlying functions that model the specular reflections are easier to learn.

We also visualize the fitted specular color of both approaches in \cref{fig:toy2}.
Our 3D Gaussian directional encoding more accurately captures the spatial variations of the specular components.

Similar to Ref-NeRF, we use an additional ``roughness'' value $\rough$ to control the maximum frequency of the specular color.
We achieve this in our Gaussian embedding
by multiplying each Gaussian's scale $\scale_i$ with the roughness $\rough$.
Intuitively, a larger Gaussian results in a smoother function with varying direction $\dir$.
Substituting the $\scale_i$ with $\rough\scale_i$ in \cref{eq:gau_embed} leads to the complete equation of our 3D Gaussian encoding.
\Cref{fig:toy2} demonstrates the ability of our 3D Gaussian-based encoding to modify roughness on the fly.

\subsection{Optimizing the Gaussian Directional Encoding}

It is worth noting that our proposed Gaussian encoding correctly models spatially varying specular reflections only when the Gaussians are positioned properly in  3D space.
We thus jointly optimize the Gaussian parameters together with the NeRF during training, to ensure the Gaussians are in the optimal state for modeling reflections.
However, there is no direct supervision for the Gaussian parameters.

Our experiments show that without proper initial Gaussian parameters, the optimization may lead to suboptimal local minima, resulting in inconsistent quality of specular reconstruction.
To address this, we propose an initialization stage for the Gaussian parameters and to bootstrap the specular color prediction.
As mentioned earlier, the specular color is essentially the preconvolved incident light, which can be directly deduced from input images.

Motivated by this observation, we train the 3D Gaussians and the specular decoder (MLP\textsubscript{3} in \cref{fig:overview})
in the initialization stage using the input images.
We train an incident light field that accommodates a diversity of rays and roughness values.
Therefore, we apply a range of Gaussian blurs to all input images using a series of standard deviations, generating Gaussian pyramids.
These pyramids of input images provide a pseudo target for incident light under different degrees of surface roughness.
In each iteration of the training, we sample pixels from the pyramids and trace rays to these pixels.
The traced rays are also associated with a roughness value that is equivalent to the blur's standard deviation.
We encode each ray with roughness using our Gaussian directional encoding, and predict the specular color $\col_\text{s}$ using the decoder.
By minimizing the errors between $\col_\text{s}$ and the pseudo ground truth, we refine the Gaussian parameters and the specular decoder, which then serve as initialization for the subsequent joint optimization stage.

\begin{figure}
  \centering
  
  \footnotesize 
  \begin{tabular}{@{}c@{\hspace{1pt}}c@{\hspace{1pt}}c@{\hspace{1pt}}c@{}}
   	Target &
   	Full &
   	w/o $\mathcal{L}_\text{mono}$ &
   	w/o early stop \\
    \includegraphics[width=0.245\linewidth]{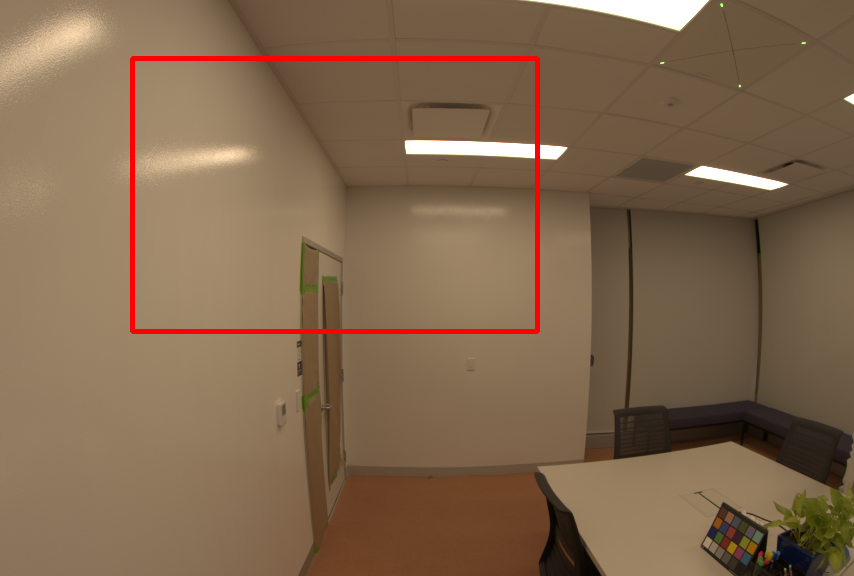} &
    \includegraphics[width=0.245\linewidth]{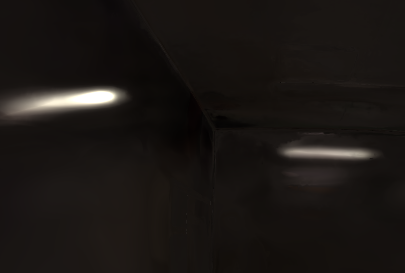} &
    \includegraphics[width=0.245\linewidth]{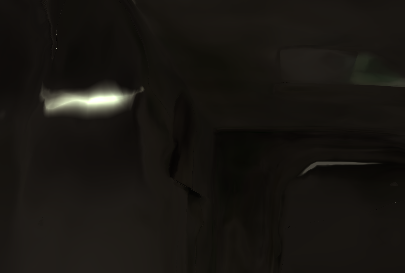} &
    \includegraphics[width=0.245\linewidth]{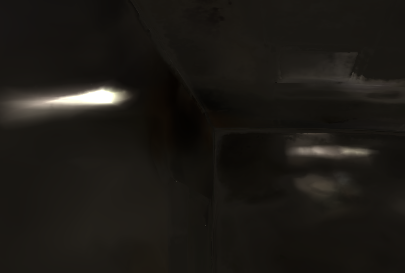} \\[-2pt]
    \includegraphics[width=0.245\linewidth]{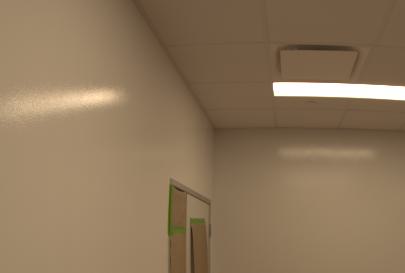} &
    \includegraphics[width=0.245\linewidth]{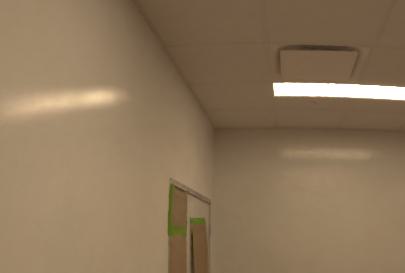} &
    \includegraphics[width=0.245\linewidth]{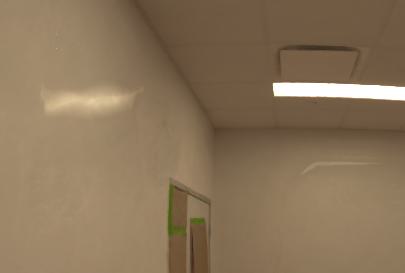} &
    \includegraphics[width=0.245\linewidth]{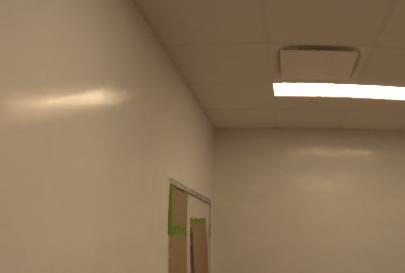} \\[-2pt]
    \includegraphics[width=0.245\linewidth]{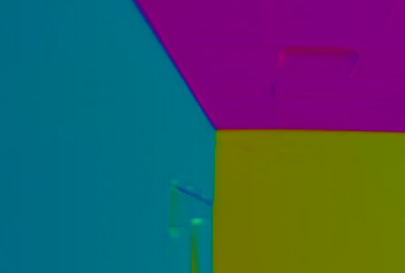} &
    \includegraphics[width=0.245\linewidth]{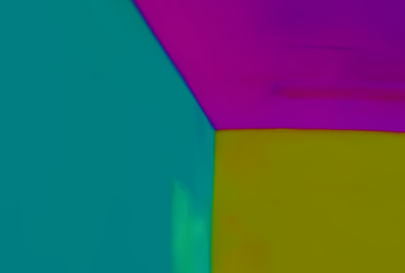} &
    \includegraphics[width=0.245\linewidth]{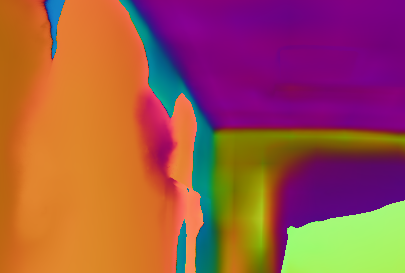} &
    \includegraphics[width=0.245\linewidth]{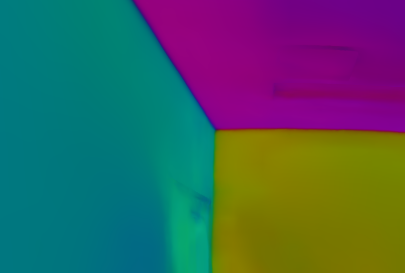}
  \end{tabular}
  \vspace{-4mm}
  \caption{The specular component reconstruction (first row, except the first image), novel-view synthesis results (second row) and normal visualizations (third row) under varying monocular normal supervision.
    The target normal visualizes the monocular normal prediction.
    Without $\mathcal{L}_\text{mono}$, the predicted normal exhibits enormous error, leading to poor specular reconstruction.
    Without early stopping $\mathcal{L}_\text{mono}$, minor errors in the predicted normals lead to a slight degradation in the reflection quality compared to our full model.
  }
  \vspace{-4mm}
  \label{fig:ab_mono}
\end{figure}

\subsection{Resolving the Shape--Radiance Ambiguity}

Regardless of any view-dependent parameterization, there remains a fundamental ambiguity between shape and radiance in NeRFs.
For example, consider a perfect mirror reflection.
Without any prior knowledge, it is nearly impossible for the NeRF model to tell whether the mirror is a flat surface with perfect reflection, or a window to a (virtual) scene behind the surface.
Therefore, prior information is needed to guide the model to learn the correct geometry.
Inspired by recent progress in monocular geometry estimation \cite{EftekSMZ2021,RanftLHSK2022,BhatBWWM2023,YuPNSG2022}, we propose to supervise the predicted normal $\normal$ using monocular normal estimation $\normal_\text{mono}$ \cite{EftekSMZ2021}:
\begin{align}
    \mathcal{L}_\text{mono} = \sum_{j} \norm{\normal^j - \mathbf{R}^j \; \normal^j_\text{mono}}^2_2 \text{,}
    \label{eq:mono_loss}
\end{align}
where the superscript $j$ is a ray index, and $\mathbf{R}^j$ is the corresponding camera rotation matrix that converts normals from view space to world space.

However, monocular normals are prone to error.
We therefore %
use them primarily as initialization and apply $\mathcal{L}_\text{mono}$ only at the beginning of the training, so that the errors in the normals do not overwhelm the geometry of the NeRF.
\Cref{fig:ab_mono} and \cref{tab:ablation} show results with different configurations of $\mathcal{L}_\text{mono}$.
We can see that without monocular normal as supervision (`w/o $\mathcal{L}_\text{mono}$'), the predicted normals have catastrophic errors, such as
those pointing inwards \textcolor{orange}{(orange)} or lying parallel \textcolor{violet}{(violet)} to the surface.
Consequently, the learned specular component is less accurate due to the incorrect normals.
Despite this, a somewhat plausible specular reflection can still be learned as the Gaussian \encoding can ``cheat'' the reflections even with erroneous normals.
On the other hand, without early stopping of the loss (`w/o early stop'), minor inaccuracies from the monocular normals permeate into predicted normals, leading to a degradation of the reflection quality. %

\begin{table*}[t]
\caption{\label{tab:comparison}%
    Quantitative comparisons of novel-view synthesis on three datasets. We highlight the best numbers in \textbf{bold}.
}
\vspace{-2mm}
\begin{tabularx}{\textwidth}{p{2.2cm}|YYY|YYY|YYY}
\toprule
Methods & \multicolumn{3}{c|}{Eyeful Tower dataset \cite{XuALGBKRPKBLZR2023}} & \multicolumn{3}{c|}{NISR \cite{WuXZBHTX2022} + Inria \cite{PhiliMGD2021} dataset} & \multicolumn{3}{c}{Shiny dataset \cite{VerbiHMZBS2022}} \\
 &
  \multicolumn{1}{c}{PSNR $\uparrow$} &
  \multicolumn{1}{c}{SSIM $\uparrow$} &
  \multicolumn{1}{c|}{LPIPS $\downarrow$} &
  \multicolumn{1}{c}{PSNR $\uparrow$} &
  \multicolumn{1}{c}{SSIM $\uparrow$} &
  \multicolumn{1}{c|}{LPIPS $\downarrow$} &
  \multicolumn{1}{c}{PSNR $\uparrow$} &
  \multicolumn{1}{c}{SSIM $\uparrow$} &
  \multicolumn{1}{c}{LPIPS $\downarrow$} \\
\midrule
Ours                           & \bf 32.583 & \bf 0.9328 & \bf 0.1445 & \bf 30.771 & \bf 0.8909 & \bf 0.1655 & \bf 26.564 & \bf 0.7277 & \bf 0.2776 \\
NeRF~\cite{MildeSTBRN2020}     &     31.854 &     0.9254 &     0.1626 &     30.748 &     0.8873 &     0.1728 &     26.469 &     0.7235   &   0.2852  \\
Ref-NeRF~\cite{VerbiHMZBS2022} &     31.652 &     0.9258 &     0.1570 &     30.654 &     0.8903 &     0.1669 &  26.502 & 0.7242   &  0.2827 \\
MS-NeRF~\cite{YinQCR2023}      &     31.715 &     0.9311 &     0.1561 &     30.224 &     0.8840 &     0.1816 &  26.466 &  0.7070   &  0.3225 \\
\bottomrule
\end{tabularx}
\vspace{-4mm}
\end{table*}

\begin{figure*}
	\rotatebox[origin=c]{90}{\vspace{-3mm} \footnotesize Eyeful \textit{Workshop}}
	\begin{minipage}{0.183\textwidth}
		\begin{subfigure}{\textwidth}
			\caption*{GT Test Image}
			\includegraphics[width=\linewidth]{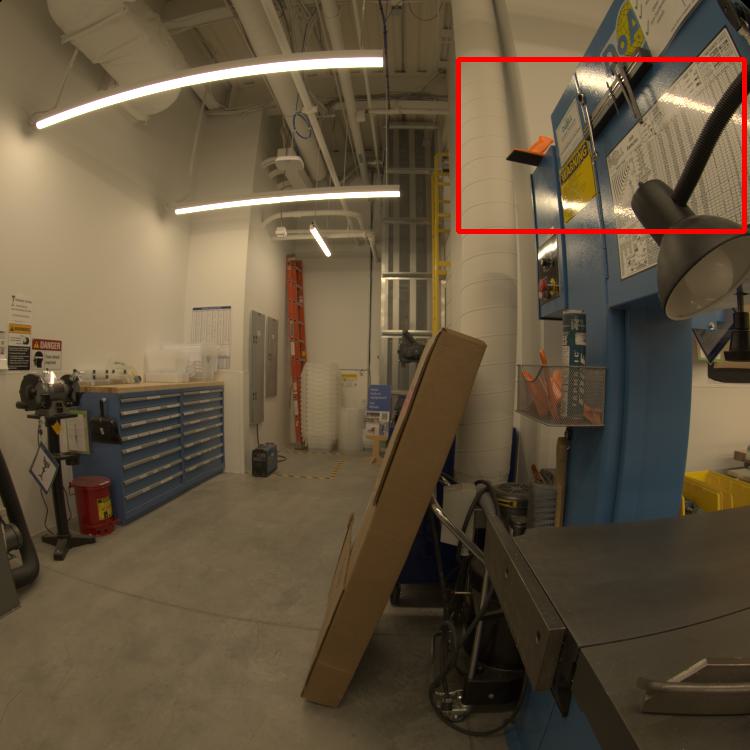}
		\end{subfigure}
	\end{minipage}
	\begin{minipage}{0.793\textwidth}
		\begin{subfigure}{0.19\linewidth}
			\caption*{GT \& SfM Normal}
			\includegraphics[width=\linewidth]{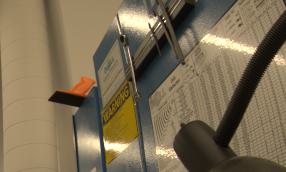}
		\end{subfigure}
		\begin{subfigure}{0.19\linewidth}
			\caption*{Ours}
			\includegraphics[width=\linewidth]{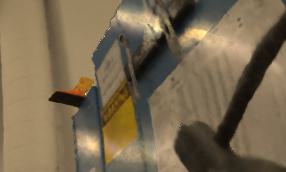}
		\end{subfigure}
		\begin{subfigure}{0.19\linewidth}
			\caption*{Ref-NeRF \cite{VerbiHMZBS2022}}
			\includegraphics[width=\linewidth]{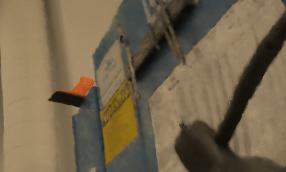}
		\end{subfigure}
		\begin{subfigure}{0.19\linewidth}
			\caption*{MS-NeRF \cite{YinQCR2023}}
			\includegraphics[width=\linewidth]{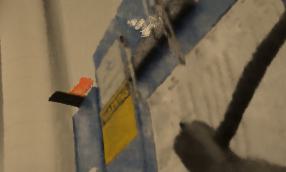}
		\end{subfigure}
		\begin{subfigure}{0.19\linewidth}
			\caption*{NeRF \cite{MildeSTBRN2020}}
			\includegraphics[width=\linewidth]{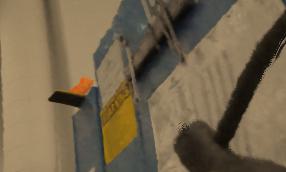}
		\end{subfigure} \\
		\begin{subfigure}{0.19\linewidth}
			\includegraphics[width=\linewidth]{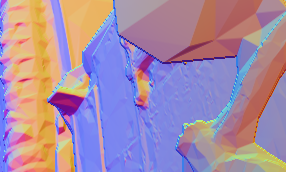}
		\end{subfigure}
		\begin{subfigure}{0.19\linewidth}
			\includegraphics[width=\linewidth]{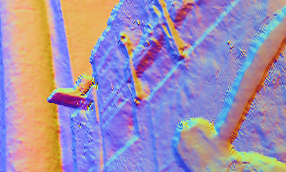}
		\end{subfigure}
		\begin{subfigure}{0.19\linewidth}
			\includegraphics[width=\linewidth]{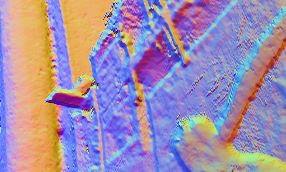}
		\end{subfigure}
		\begin{subfigure}{0.19\linewidth}
			\includegraphics[width=\linewidth]{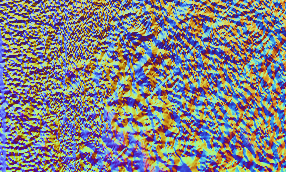}
		\end{subfigure}
		\begin{subfigure}{0.19\linewidth}
			\includegraphics[width=\linewidth]{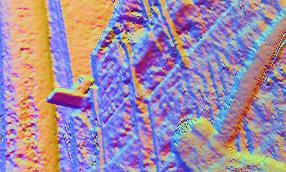}
		\end{subfigure}
	\end{minipage}
	
	\rotatebox[origin=c]{90}{\footnotesize NISR \textit{LivingRoom2}}
	\begin{minipage}{0.183\textwidth}
		\begin{subfigure}{\textwidth}
			\includegraphics[width=\linewidth]{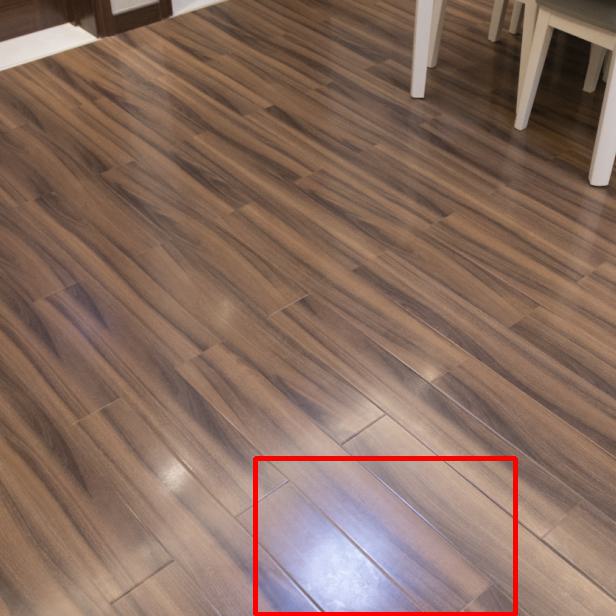}
		\end{subfigure}
	\end{minipage}
	\begin{minipage}{0.793\textwidth}
		\begin{subfigure}{0.19\linewidth}
			\includegraphics[width=\linewidth]{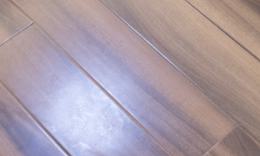}
		\end{subfigure}
		\begin{subfigure}{0.19\linewidth}
			\includegraphics[width=\linewidth]{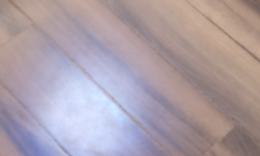}
		\end{subfigure}
		\begin{subfigure}{0.19\linewidth}
			\includegraphics[width=\linewidth]{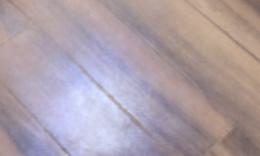}
		\end{subfigure}
		\begin{subfigure}{0.19\linewidth}
			\includegraphics[width=\linewidth]{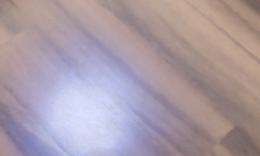}
		\end{subfigure}
		\begin{subfigure}{0.19\linewidth}
			\includegraphics[width=\linewidth]{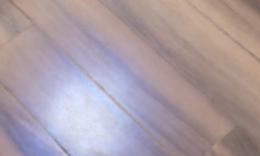}
		\end{subfigure} \\
		\begin{subfigure}{0.19\linewidth}
			\includegraphics[width=\linewidth]{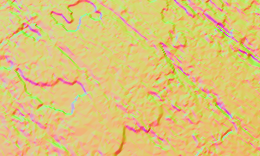}
		\end{subfigure}
		\begin{subfigure}{0.19\linewidth}
			\includegraphics[width=\linewidth]{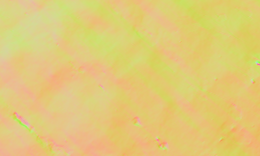}
		\end{subfigure}
		\begin{subfigure}{0.19\linewidth}
			\includegraphics[width=\linewidth]{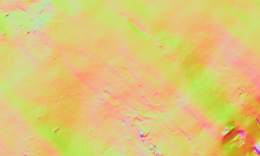}
		\end{subfigure}
		\begin{subfigure}{0.19\linewidth}
			\includegraphics[width=\linewidth]{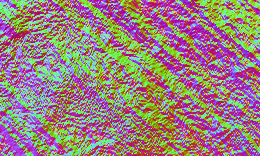}
		\end{subfigure}
		\begin{subfigure}{0.19\linewidth}
			\includegraphics[width=\linewidth]{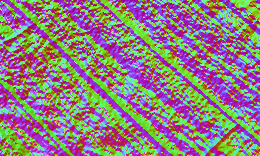}
		\end{subfigure}
	\end{minipage}
	
	\rotatebox[origin=c]{90}{\footnotesize Eyeful \textit{Office2}}
	\begin{minipage}{0.183\textwidth}
		\begin{subfigure}{\textwidth}
			\includegraphics[width=\linewidth]{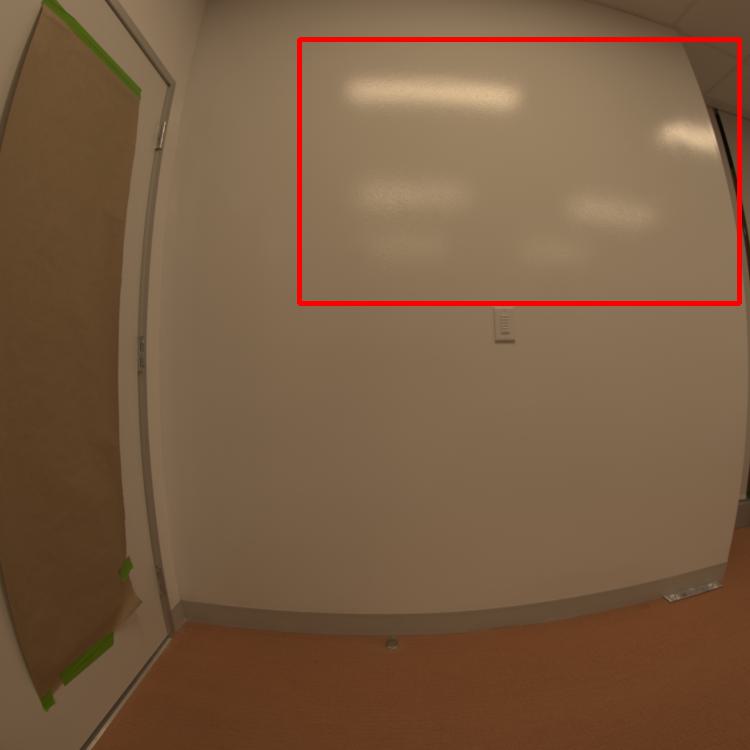}
		\end{subfigure}
	\end{minipage}
	\begin{minipage}{0.793\textwidth}
		\begin{subfigure}{0.19\linewidth}
			\includegraphics[width=\linewidth]{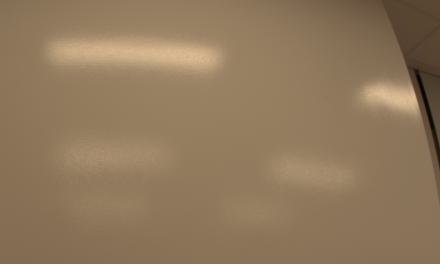}
		\end{subfigure}
		\begin{subfigure}{0.19\linewidth}
			\includegraphics[width=\linewidth]{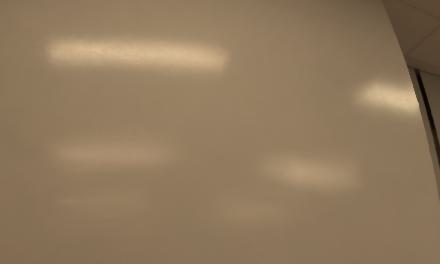}
		\end{subfigure}
		\begin{subfigure}{0.19\linewidth}
			\includegraphics[width=\linewidth]{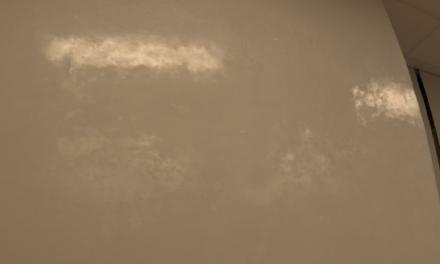}
		\end{subfigure}
		\begin{subfigure}{0.19\linewidth}
			\includegraphics[width=\linewidth]{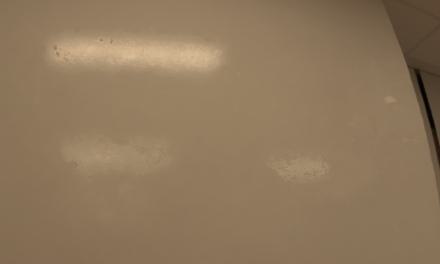}
		\end{subfigure}
		\begin{subfigure}{0.19\linewidth}
			\includegraphics[width=\linewidth]{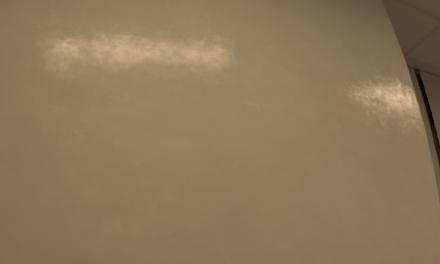}
		\end{subfigure} \\
		\begin{subfigure}{0.19\linewidth}
			\includegraphics[width=\linewidth]{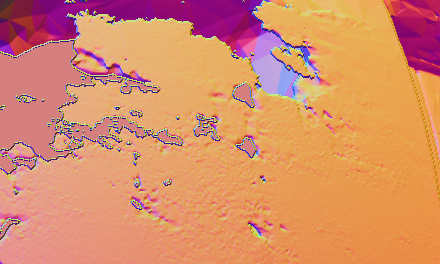}
		\end{subfigure}
		\begin{subfigure}{0.19\linewidth}
			\includegraphics[width=\linewidth]{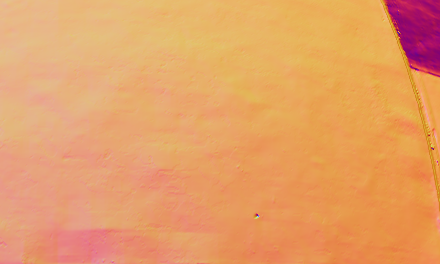}
		\end{subfigure}
		\begin{subfigure}{0.19\linewidth}
			\includegraphics[width=\linewidth]{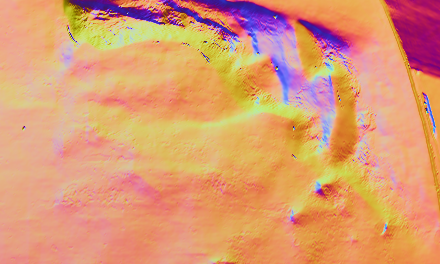}
		\end{subfigure}
		\begin{subfigure}{0.19\linewidth}
			\includegraphics[width=\linewidth]{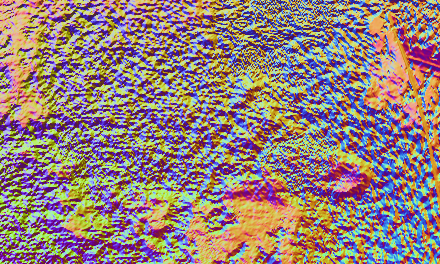}
		\end{subfigure}
		\begin{subfigure}{0.19\linewidth}
			\includegraphics[width=\linewidth]{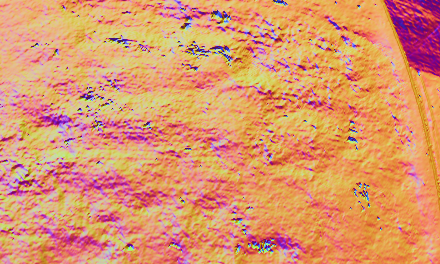}
		\end{subfigure}
	\end{minipage}
 
    \vspace{-2mm}
    \caption{\label{fig:comparisons}%
        Comparisons of novel-view synthesis quality and normal map visualizations.
        Our method consistently reconstructs reflections while other methods either produce `faked' reflections, resulting in incorrect normals, or fail to model reflections entirely.
    }
    \vspace{-5mm}
\end{figure*}

\subsection{Losses}

To jointly optimize all parameters within our proposed pipeline, we use a combination of loss terms:
\begin{equation}
    \mathcal{L} = \mathcal{L}_\text{c} + \mathcal{L}_\text{prop} + \lambda_\text{dist} \mathcal{L}_\text{dist} + \lambda_\text{mono} \mathcal{L}_\text{mono} + \lambda_\text{norm} \mathcal{L}_\text{norm} \text{.}
\end{equation}
In this equation, $\mathcal{L}_\text{c}$ is the L1 reconstruction loss between the predicted and ground-truth colors.
The terms
$\mathcal{L}_\text{prop}$ and $\mathcal{L}_\text{dist}$ are adopted from mip-NeRF 360 \cite{mipnerf360}, where $\mathcal{L}_\text{prop}$ supervises the density proposal networks, and $\mathcal{L}_\text{dist}$ is the distortion loss encouraging density sparsity.
To tie predicted normals to the density field, we use Ref-NeRF's normal prediction loss $\mathcal{L}_\text{norm}$ \cite{VerbiHMZBS2022}, which guides the predicted normal $\normal$ with the density gradient direction.
Further elaboration on these loss components can be found in the supplementary material.

In our experiments, we set $\lambda_\text{dist} \!=\! 0.002$, aligning with the settings of the ``nerfacto'' model in Nerfstudio \cite{TanciWNLYWKASAMKK2023}.
For $\lambda_\text{mono}$, we choose a value of $1$ in the first 4K iterations, and reduce to $0$ thereafter to cease its effect, as described earlier.
We find that our method is robust to the value of $\lambda_\text{mono}$ as it only serves as initialization.
We also assign $\lambda_\text{norm} = 10^{-3}$, which is slightly higher than the weight in Ref-NeRF \cite{VerbiHMZBS2022}, as we find that this produces slightly smoother normals without substantially compromising the rendering quality.

\section{Experiments}
\label{sec:experiments}

\paragraph{Implementation}
To model room-scale scenes, we employ a network architecture similar to the ``nerfacto'' model presented in Nerfstudio \cite{TanciWNLYWKASAMKK2023}.
We use two small density networks as proposal networks, supervised via $\mathcal{L}_\text{prop}$.
We sample 256 and 96 points for each proposal network, and 48 points for the final NeRF model.
These three networks all use hash-based positional encodings.
When querying the hash features in the final NeRF model, we incorporate the LOD-aware scheme proposed in VR-NeRF \cite{XuALGBKRPKBLZR2023}.
We train our model for 100,000 iterations and randomly sample 12,800 rays in each iteration.
This process takes around 8\,GB of GPU memory and approximately 3.5 hours to train a model using an NVIDIA A100 GPU.
Further details regarding the model's structure can be found in the supplementary materials.

\begin{figure*}
	\centering
	\begin{minipage}{0.14\textwidth}
		\centering
		\begin{subfigure}{2.34cm}
			\caption*{Test Image}
			\includegraphics[width=\linewidth]{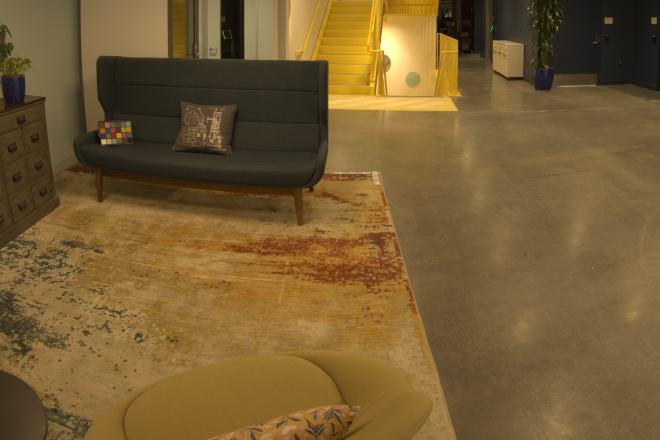}
		\end{subfigure}
	\end{minipage}
	\begin{minipage}{0.855\textwidth}
		\rotatebox{90}{\hspace{10pt} \centering\footnotesize Ours}
		\begin{subfigure}{2.34cm}
			\caption*{Final}
			\includegraphics[width=\linewidth]{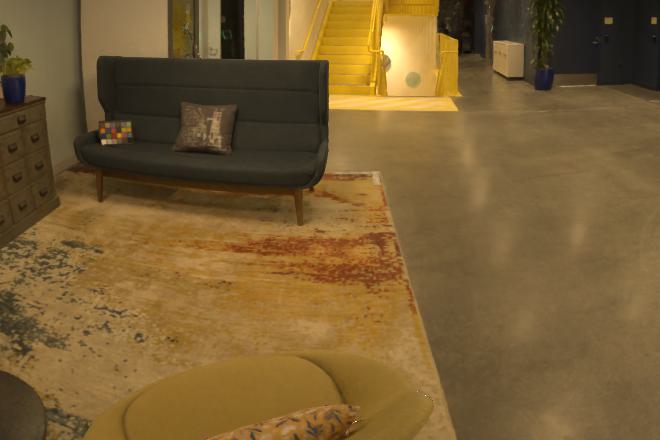}
		\end{subfigure}
		\hfill
		\begin{subfigure}{2.34cm}
			\caption*{Diffuse}
			\includegraphics[width=\linewidth]{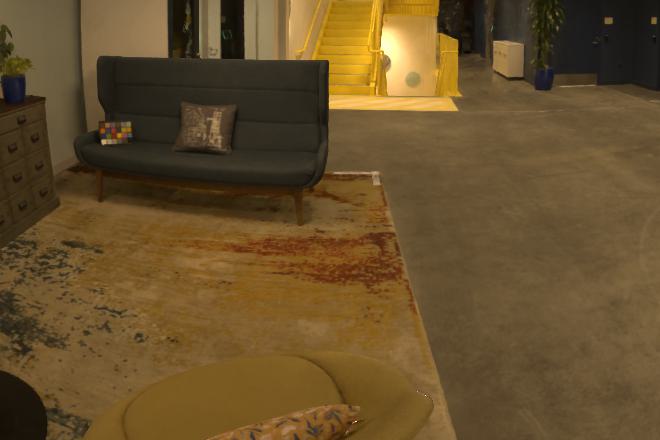}
		\end{subfigure}
		\hfill
		\begin{subfigure}{2.34cm}
			\caption*{Specular}
			\includegraphics[width=\linewidth]{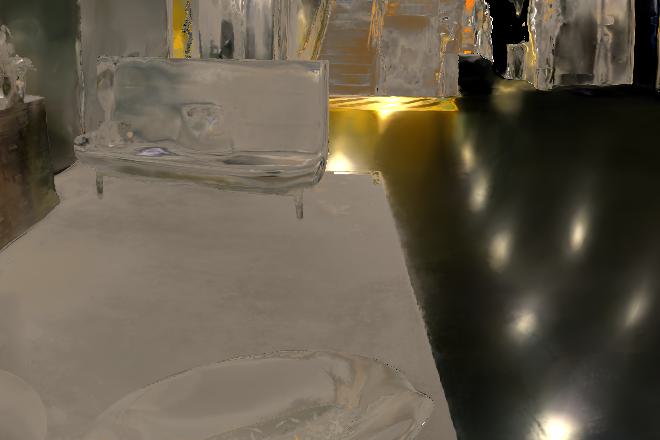}
		\end{subfigure}
		\hfill
		\begin{subfigure}{2.34cm}
			\caption*{Tint}
			\includegraphics[width=\linewidth]{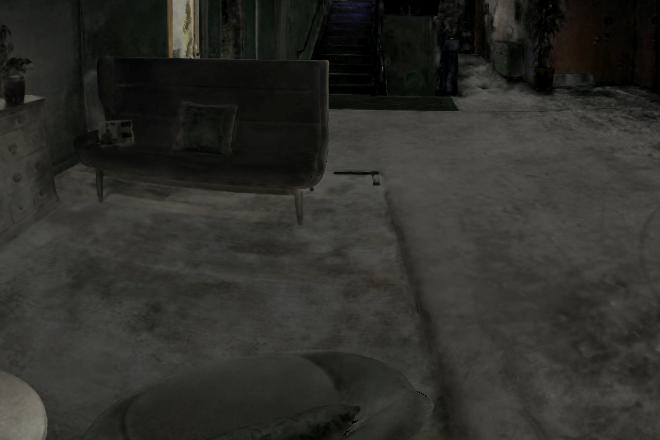}
		\end{subfigure}
		\hfill
		\begin{subfigure}{2.34cm}
			\caption*{Roughness}
			\includegraphics[width=\linewidth]{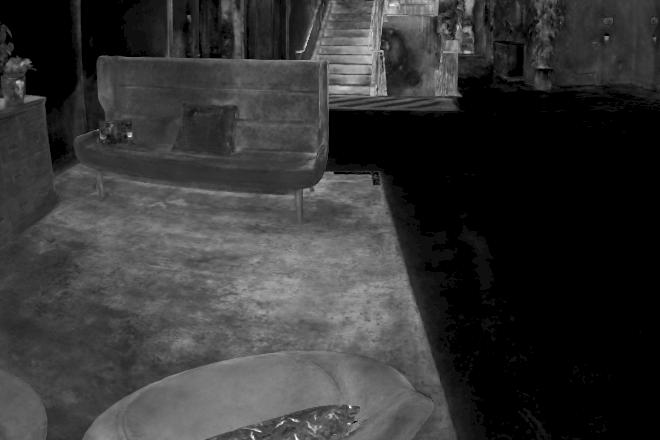}
		\end{subfigure}
		\hfill
		\begin{subfigure}{2.34cm}
			\caption*{Normal}
			\includegraphics[width=\linewidth]{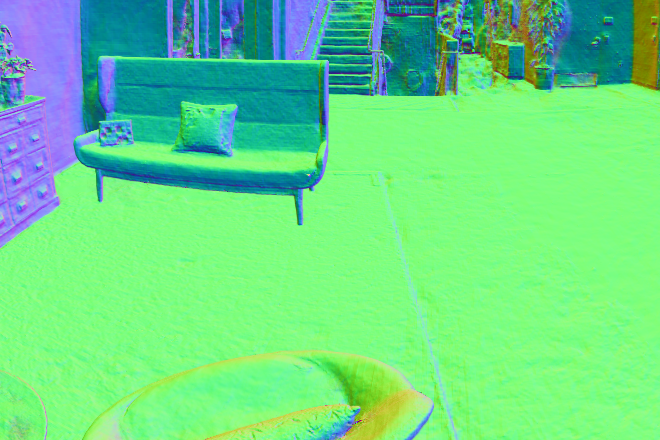}
		\end{subfigure} \\
		\rotatebox{90}{\hspace{3pt} \centering\footnotesize Ref-NeRF}
		\begin{subfigure}{2.34cm}
			\includegraphics[width=\linewidth]{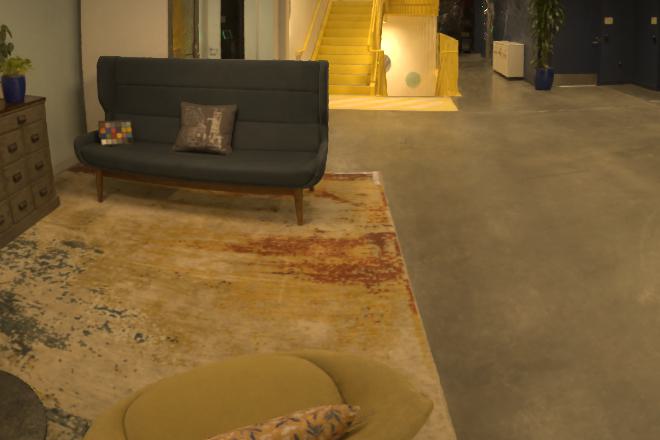}
		\end{subfigure}
		\hfill
		\begin{subfigure}{2.34cm}
			\includegraphics[width=\linewidth]{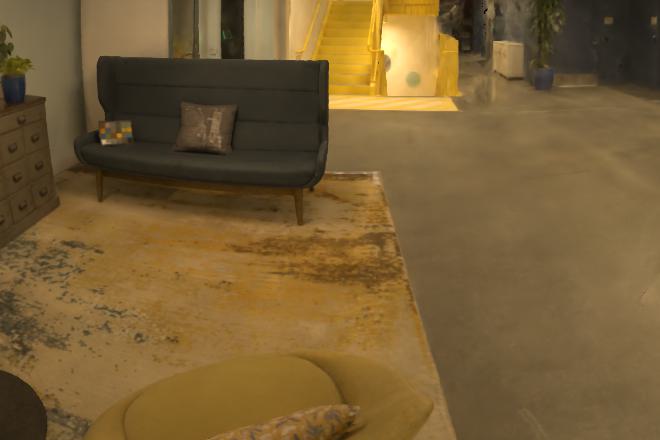}
		\end{subfigure}
		\hfill
		\begin{subfigure}{2.34cm}
			\includegraphics[width=\linewidth]{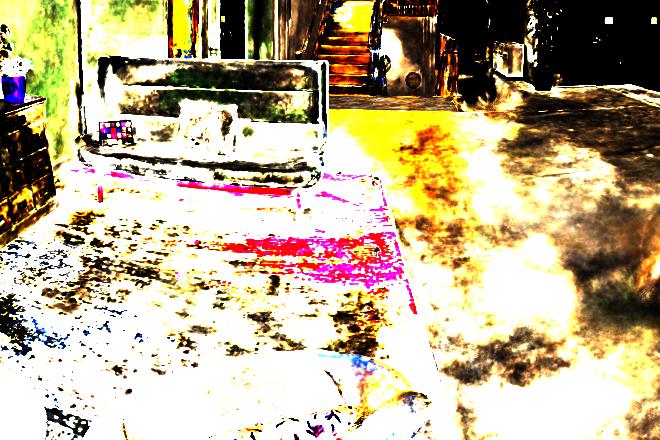}
		\end{subfigure}
		\hfill
		\begin{subfigure}{2.34cm}
			\includegraphics[width=\linewidth]{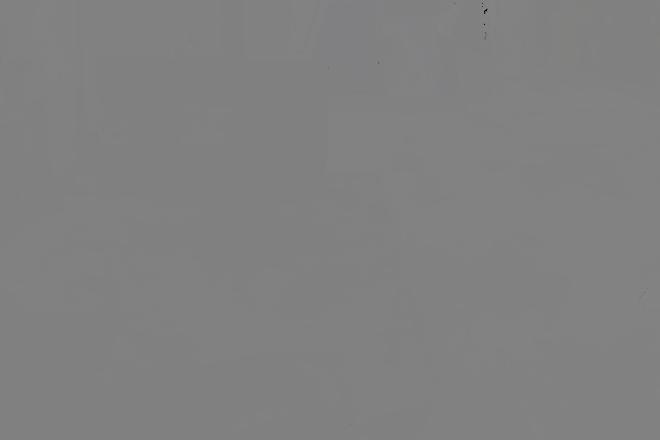}
		\end{subfigure}
		\hfill
		\begin{subfigure}{2.34cm}
			\includegraphics[width=\linewidth]{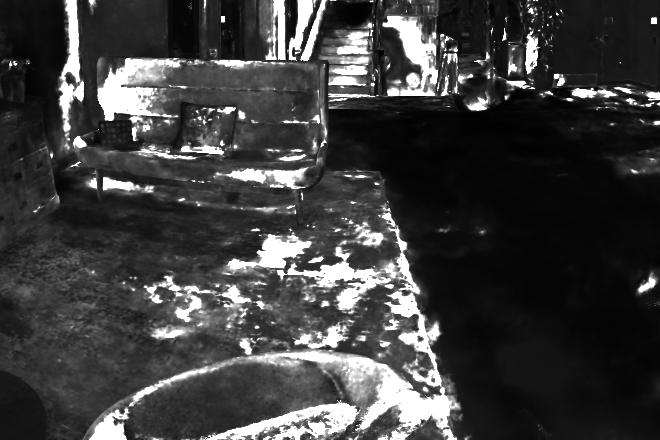}
		\end{subfigure}
		\hfill
		\begin{subfigure}{2.34cm}
			\includegraphics[width=\linewidth]{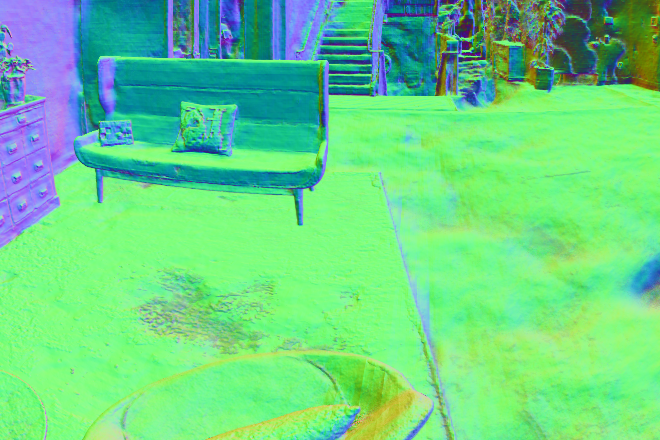}
		\end{subfigure}
	\end{minipage}

	\vspace{4pt}

	\begin{minipage}{0.14\textwidth}
		\centering
		\begin{subfigure}{2.34cm}
			\caption*{Test Image}
			\includegraphics[width=\linewidth]{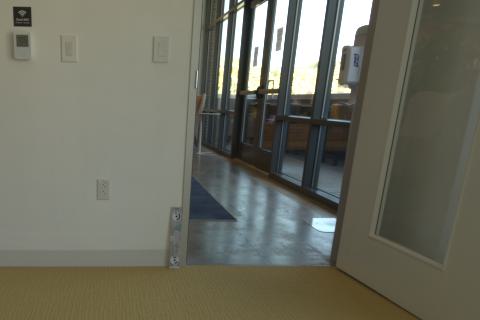}
		\end{subfigure}
	\end{minipage}
	\begin{minipage}{0.855\textwidth}
		\rotatebox{90}{\hspace{10pt} \centering\footnotesize Ours}
		\begin{subfigure}{2.34cm}
			\includegraphics[width=\linewidth]{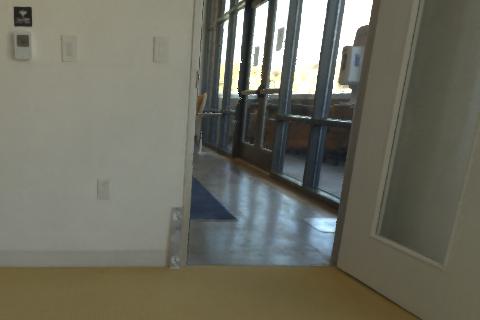}
		\end{subfigure}
		\hfill
		\begin{subfigure}{2.34cm}
			\includegraphics[width=\linewidth]{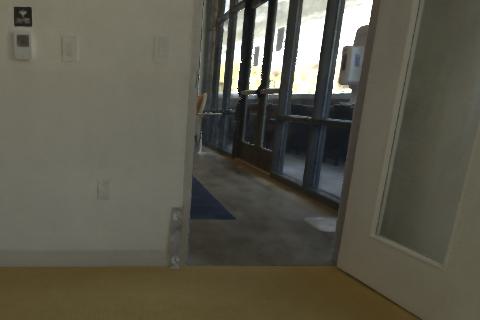}
		\end{subfigure}
		\hfill
		\begin{subfigure}{2.34cm}
			\includegraphics[width=\linewidth]{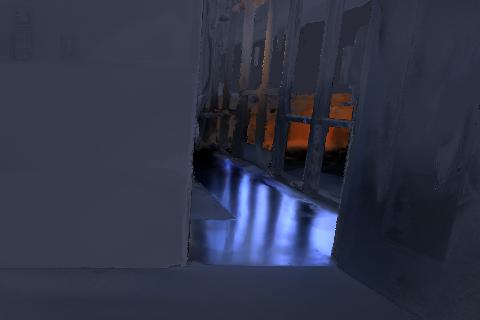}
		\end{subfigure}
		\hfill
		\begin{subfigure}{2.34cm}
			\includegraphics[width=\linewidth]{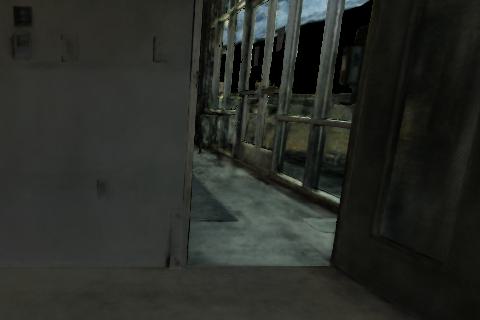}
		\end{subfigure}
		\hfill
		\begin{subfigure}{2.34cm}
			\includegraphics[width=\linewidth]{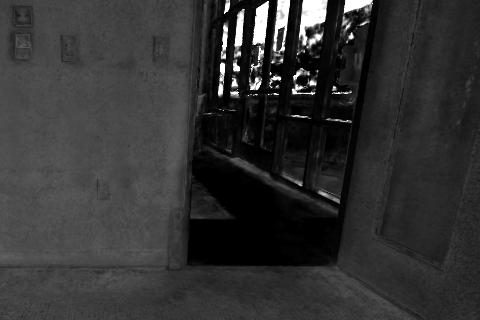}
		\end{subfigure}
		\hfill
		\begin{subfigure}{2.34cm}
			\includegraphics[width=\linewidth]{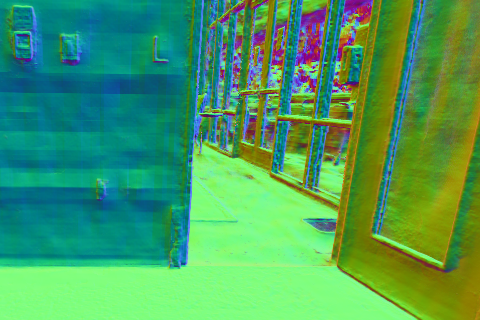}
		\end{subfigure} \\
		\rotatebox{90}{\hspace{3pt} \centering\footnotesize Ref-NeRF}
		\begin{subfigure}{2.34cm}
			\includegraphics[width=\linewidth]{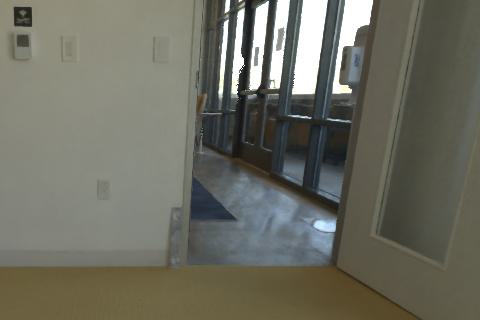}
		\end{subfigure}
		\hfill
		\begin{subfigure}{2.34cm}
			\includegraphics[width=\linewidth]{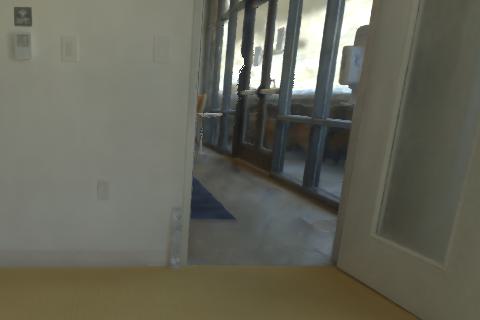}
		\end{subfigure}
		\hfill
		\begin{subfigure}{2.34cm}
			\includegraphics[width=\linewidth]{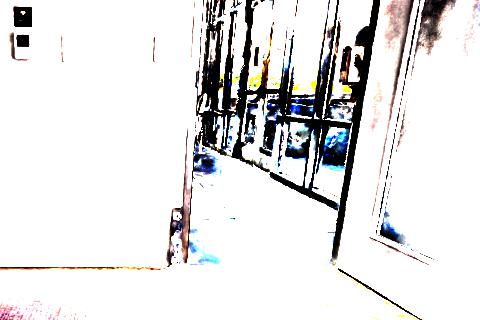}
		\end{subfigure}
		\hfill
		\begin{subfigure}{2.34cm}
			\includegraphics[width=\linewidth]{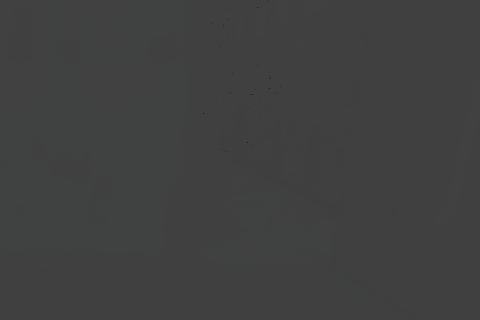}
		\end{subfigure}
		\hfill
		\begin{subfigure}{2.34cm}
			\includegraphics[width=\linewidth]{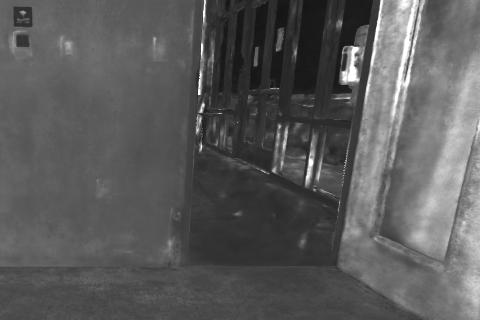}
		\end{subfigure}
		\hfill
		\begin{subfigure}{2.34cm}
			\includegraphics[width=\linewidth]{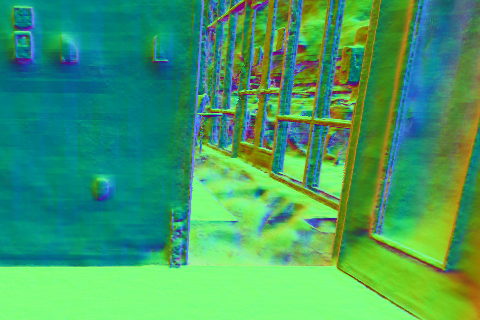}
		\end{subfigure}
	\end{minipage}
    \vspace{-3mm}
    \caption{\label{fig:decompositions}%
        Intermediate components of our approach compared to Ref-NeRF \cite{VerbiHMZBS2022}.
        Our approach produces a more meaningful decomposition under room-scale lighting settings.
    }
    \vspace{-5mm}
\end{figure*}

\paragraph{Datasets}
We evaluate our method on several datasets with a focus on indoor scenes characterized by near-field lighting conditions.
First, we evaluate on the Eyeful Tower dataset \cite{XuALGBKRPKBLZR2023}, which provides high-quality HDR captures of 11 indoor scenes.
Each scene is coupled with calibrated camera parameters and a mesh reconstructed via Agisoft Metashape \cite{AgisoftMetashape}.
We select 9 scenes that feature notable reflective properties.
We downsample the images of each scene to a resolution of 854$\times$1280\,pixels.
We curated around 50--70 views per scene that contain glossy surfaces for evaluation, leaving the remaining views for training.
We also evaluate our approach on public indoor datasets NISR \cite{WuXZBHTX2022} and Inria \cite{PhiliMGD2021} (NISR+Inria). 
Moreover, to assess the performance under far-field lighting, we evaluate the real shiny dataset in Ref-NeRF \cite{VerbiHMZBS2022}.
We report the average PSNR, SSIM, and LPIPS \cite{LPIPS} metrics for evaluating rendering quality.

\subsection{Comparisons}

We compare our method with several baselines: NeRF \cite{MildeSTBRN2020}, Ref-NeRF \cite{VerbiHMZBS2022}, and MS-NeRF \cite{YinQCR2023}, which specializes in mirror-like reflections by decomposing NeRF into multiple spaces.
For a fair comparison, we re-implement these baselines, such that we share the same NeRF backbone and rendering configurations, with the only difference being the way different methods decompose and parameterize the output color.
We report the numerical results across three datasets in \cref{tab:comparison}.
Our method demonstrates superior performance on the Eyeful Tower dataset, indicating the effectiveness of our method.
On the NISR+Inria datasets, our method marginally outperforms the baselines, likely due to the dataset containing few reflection surfaces.
Notably, while our method is tailored for near-field lighting conditions, it also shows promising results on the Shiny dataset, which comprises far-field lighting scenarios.
This is because our Gaussian directional encoding can simulate a spatially invariant encoding by positioning Gaussians at a significant distance.

Qualitative results on the Eyeful Tower and NISR+Inria datasets are provided in \cref{fig:comparisons}.
We can see that while other baselines occasionally synthesize plausible reflections, they resort to approximations that fake the reflections by placing emitters underneath the surface.
As a result, they either produce incorrect geometry, or fail to model the reflections.
Our method, in contrast, successfully models specular highlights on the surface.
We provide additional video results in the supplementary material.

We also visualize and compare the decomposition produced by our method and Ref-NeRF in \cref{fig:decompositions}.
We can see that Ref-NeRF fails to obtain a meaningful decomposition under near-field lighting, and produces holes in the geometry, whereas our method consistently achieves a realistic separation of specular and diffuse components.

\subsection{Ablation Studies}

\begin{figure}
    \centering
    \includegraphics[width=\linewidth]{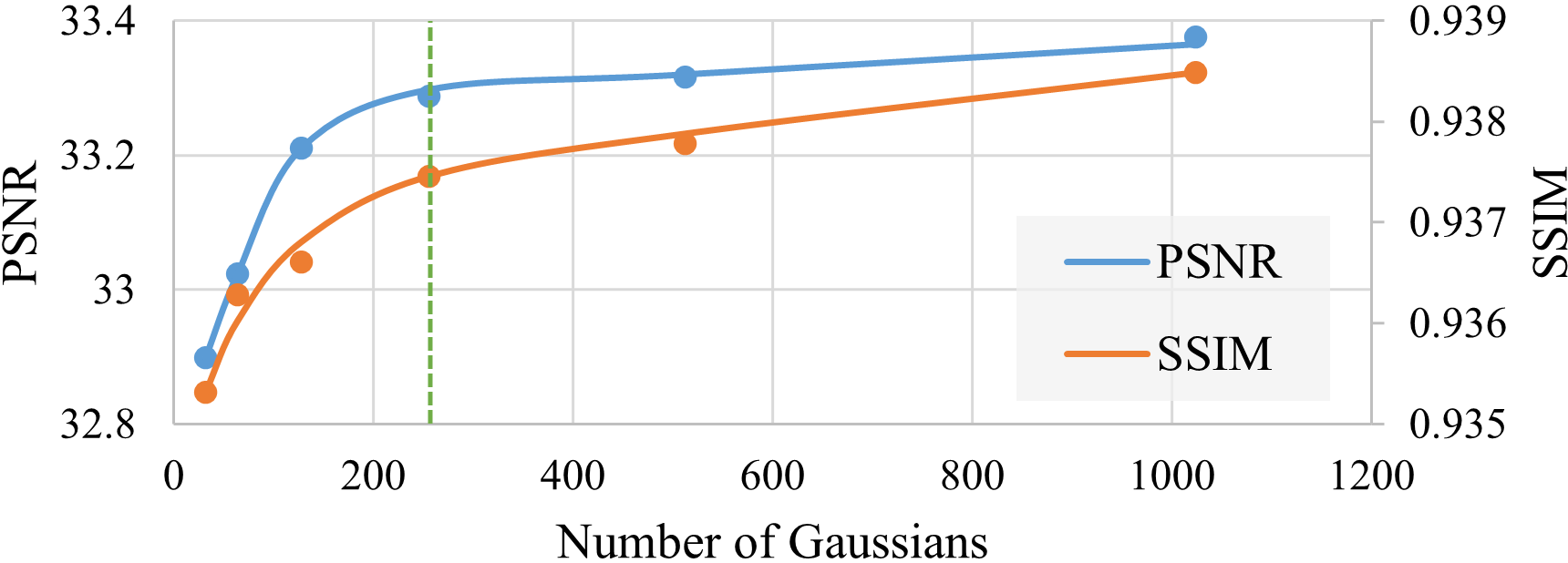}
    \vspace{-6mm}
    \caption{\label{fig:ab_numgau}%
        We evaluate the novel-view synthesis quality with respect to the number of Gaussians across five scenes.
        The green dashed line is the setting we use in our experiments.
        \vspace{-4mm}
    }
\end{figure}

\paragraph{Number of Gaussians}
One important hyperparameter in the Gaussian directional \encoding is the number of Gaussians, as it directly influences the model's capacity to represent specular colors.
We conduct experiments to evaluate the impact of varying the number of Gaussians on five scenes from the Eyeful Tower dataset, and show the relationship between the number of Gaussians and the rendering quality in \cref{fig:ab_numgau}.
The rendering quality improves when using more Gaussians, but the improvement saturates as the number increases beyond 400.
Note that using a larger number of Gaussians also entails greater computation costs and GPU memory requirements for every rendered pixel.
Therefore, we use 256 Gaussians for all experiments, to strike a balance between rendering quality and computational efficiency.

\begin{figure}
  \centering
  
  \rotatebox{90}{\footnotesize \phantom{g}}
  \begin{subfigure}{0.31\linewidth}
  	\caption*{Full}
  \end{subfigure}
  \hspace{-3pt}
  \begin{subfigure}{0.31\linewidth}
  	\caption*{w/o init}
  \end{subfigure}
  \hspace{-3pt}
  \begin{subfigure}{0.31\linewidth}
  	\caption*{w/o opt.}
  \end{subfigure}
  
  \rotatebox{90}{\hspace{4pt} \footnotesize Specular}
  \begin{subfigure}{0.31\linewidth}
    \centering
    \includegraphics[width=\linewidth]{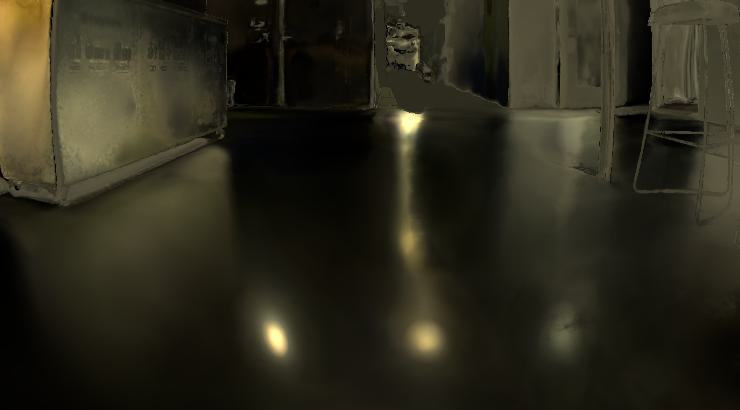}
  \end{subfigure}
  \hspace{-3pt}
  \begin{subfigure}{0.31\linewidth}
    \includegraphics[width=\linewidth]{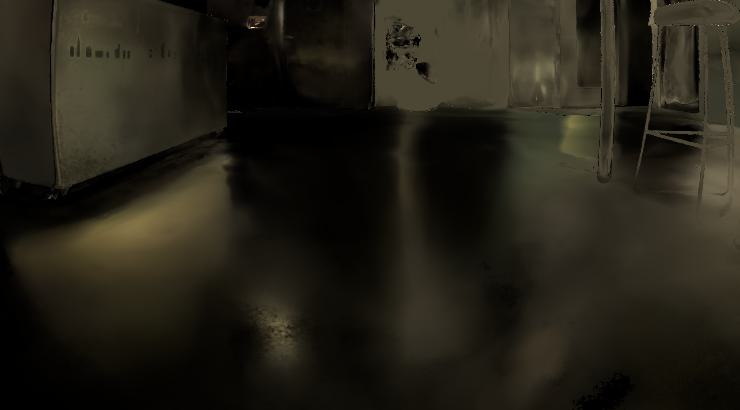}
  \end{subfigure}
  \hspace{-3pt}
  \begin{subfigure}{0.31\linewidth}
    \includegraphics[width=\linewidth]{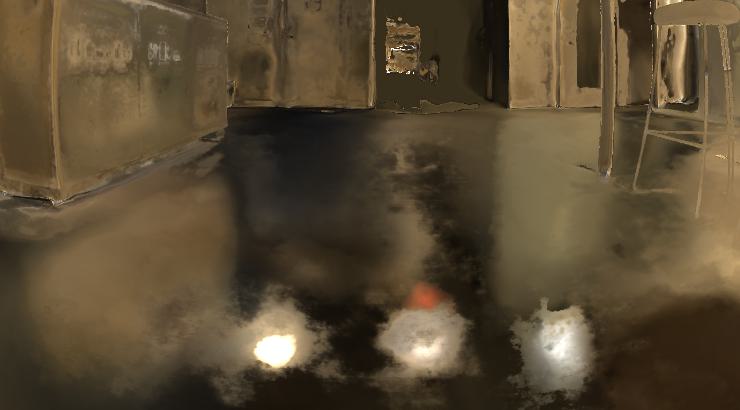}
  \end{subfigure}

  \rotatebox{90}{\hspace{11pt} \footnotesize Tint\phantom{g}}
  \begin{subfigure}{0.31\linewidth}
    \includegraphics[width=\linewidth]{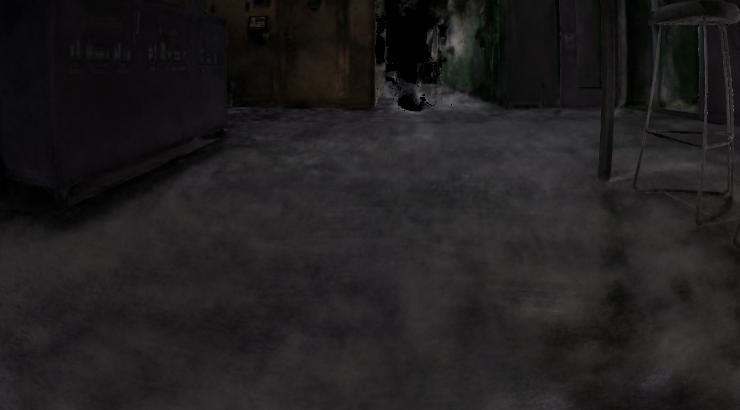}
  \end{subfigure}
  \hspace{-3pt}
  \begin{subfigure}{0.31\linewidth}
    \includegraphics[width=\linewidth]{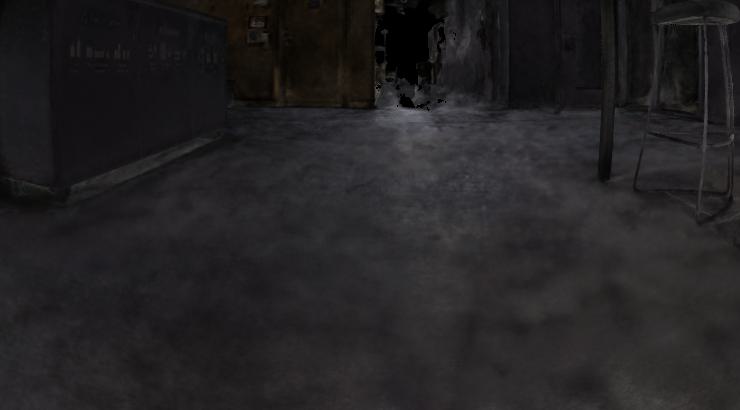}
  \end{subfigure}
  \hspace{-3pt}
  \begin{subfigure}{0.31\linewidth}
    \includegraphics[width=\linewidth]{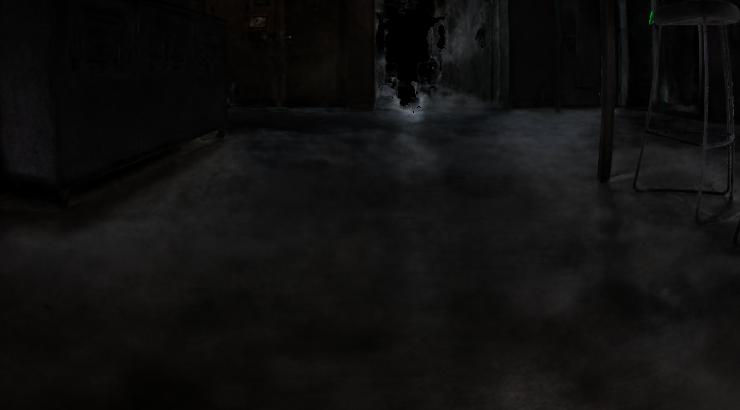}
  \end{subfigure}

  \rotatebox{90}{\hspace{8pt} \footnotesize Final\phantom{g}}
  \begin{subfigure}{0.31\linewidth}
    \includegraphics[width=\linewidth]{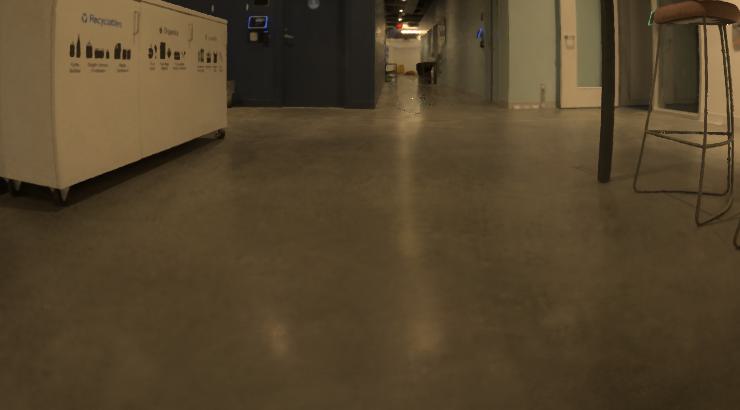}
  \end{subfigure}
  \hspace{-3pt}
  \begin{subfigure}{0.31\linewidth}
    \includegraphics[width=\linewidth]{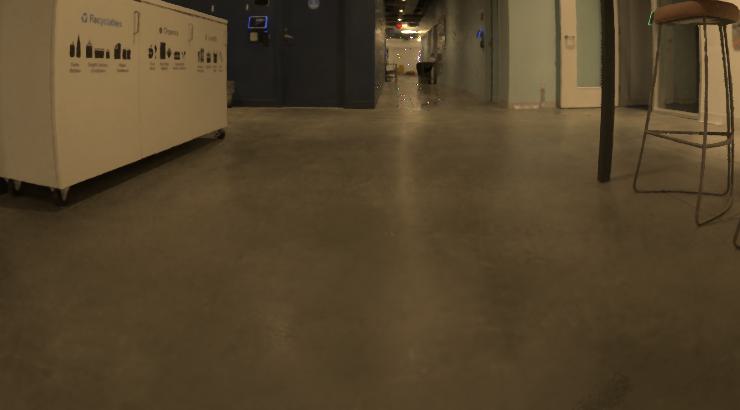}
  \end{subfigure}
  \hspace{-3pt}
  \begin{subfigure}{0.31\linewidth}
    \includegraphics[width=\linewidth]{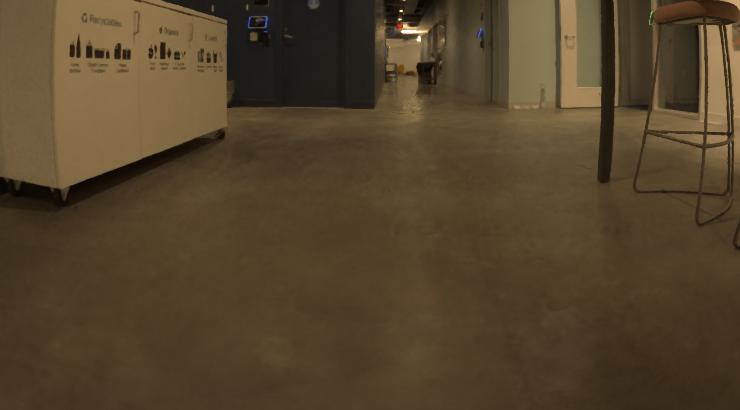}
  \end{subfigure}

  \rotatebox{90}{\hspace{2pt} \footnotesize Error map}
  \begin{subfigure}{0.31\linewidth}
    \includegraphics[width=\linewidth]{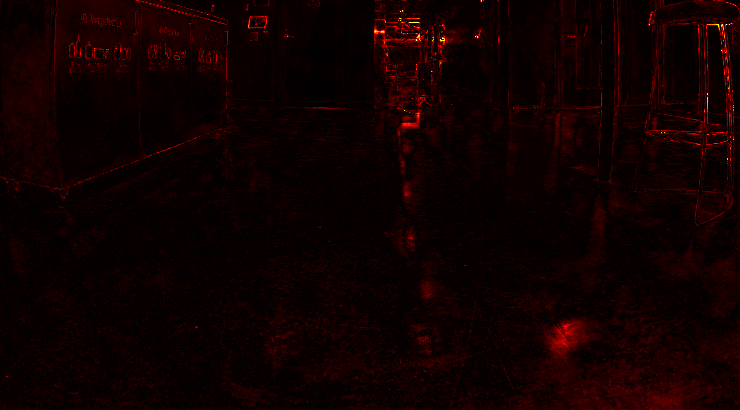}
  \end{subfigure}
  \hspace{-3pt}
  \begin{subfigure}{0.31\linewidth}
    \includegraphics[width=\linewidth]{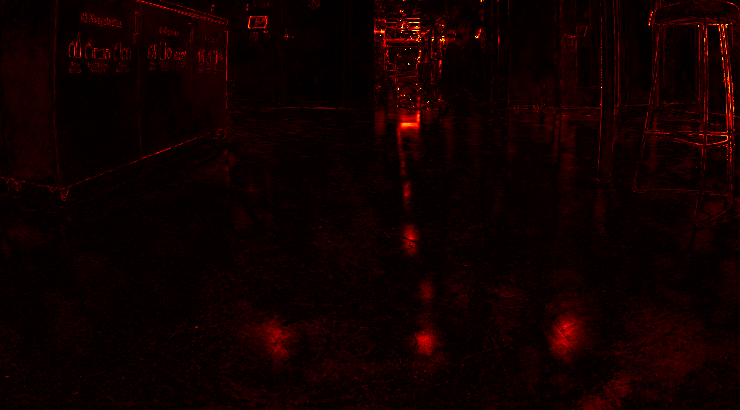}
  \end{subfigure}
  \hspace{-3pt}
  \begin{subfigure}{0.31\linewidth}
    \includegraphics[width=\linewidth]{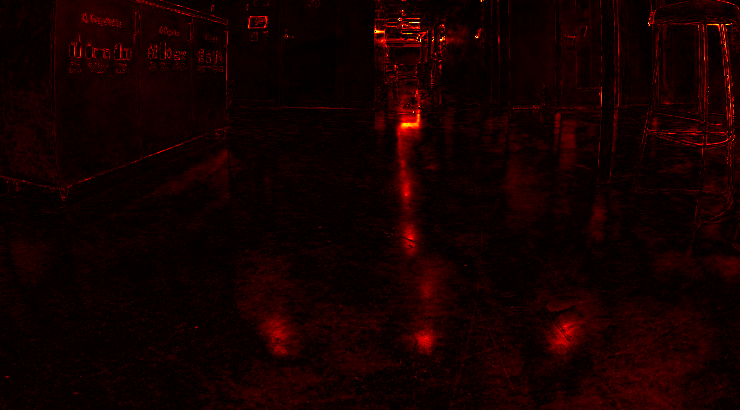}
  \end{subfigure}
  
  \vspace{-2mm}
    \caption{%
    	Example results under different Gaussian optimization settings.
    	Without initializing the Gaussian parameters (`w/o init') or optimizing Gaussians jointly with the NeRF (`w/o opt.'), the Gaussian embedding struggles to model specularities accurately.
     }
    \label{fig:ab_gauopt}
    \vspace{-5mm}
\end{figure}

\paragraph{Optimizing Gaussians}
To optimize the Gaussian directional \encoding effectively, we first initialize them by training an incident light field, and then jointly finetune the Gaussian \encoding together with the NeRF model.
We demonstrate the significance of initialization (`w/o init') and fine-tuning (`w/o opt.') by omitting each process individually.
We show quantitative results in \cref{tab:ablation} and a qualitative example in \cref{fig:ab_gauopt}.
Without initialization, the model can still reconstruct reflections to some extent, resulting in a slightly better average LPIPS score, yet it fails to model some specular details, such as the light blobs.
Neglecting the joint optimization of Gaussians leads to complete failure in modeling specular reflections.
As illustrated in \cref{fig:ab_gauopt}, with the inaccurate specular modeling, the tints suppress the specular reflections, which ultimately leads to the inability to represent reflections in the final rendered image.

\begin{table}
\caption{Ablations of our method on the Eyeful Tower dataset \cite{XuALGBKRPKBLZR2023}.
	The ``e.s.'' indicates early stopping the $\mathcal{L}_\text{mono}$ after 4K iterations.}
\vspace{-2mm}
\label{tab:ablation}
\setlength{\tabcolsep}{2pt}
\small\centering
\begin{tabular}{lcccc|@{\hspace{6pt}}ccc}
  \toprule
  &
  \multicolumn{2}{c}{Gaussians} &
  \multirow[c]{2}{*}{\makecell{Mono.\\Prior}} &
  \multirow[c]{2}{*}{\makecell{E. S. \\ $\mathcal{L}_\text{mono}$}} &
  &
  & \\
  Method & 
  Init. & 
  Opt. & 
  & &
  PSNR$\uparrow$ &
  SSIM$\uparrow$ &
  LPIPS$\downarrow$ \\ \midrule
  
Full                          & \yesmark & \yesmark & \yesmark & \yesmark & \bf 32.58 & \bf 0.9328 &     0.1445 \\
w/o init                      & \nomark  & \yesmark & \yesmark & \yesmark &     32.52 &     0.9304 & \bf 0.1429 \\
w/o opt.                      & \yesmark & \nomark  & \yesmark & \yesmark &     32.06 &     0.9265 &     0.1581 \\
w/o $\mathcal{L}_\text{mono}$ & \yesmark & \yesmark & \nomark  & ---      &     32.31 &     0.9288 &     0.1503 \\
w/o e.s.                      & \yesmark & \yesmark & \yesmark & \nomark  &     32.46 &     0.9292 &     0.1502\\
\bottomrule
\end{tabular}
\vspace{-4mm}
\end{table}

\section{Discussion and Conclusion}

\begin{figure}
    \centering
    \includegraphics[width=\linewidth]{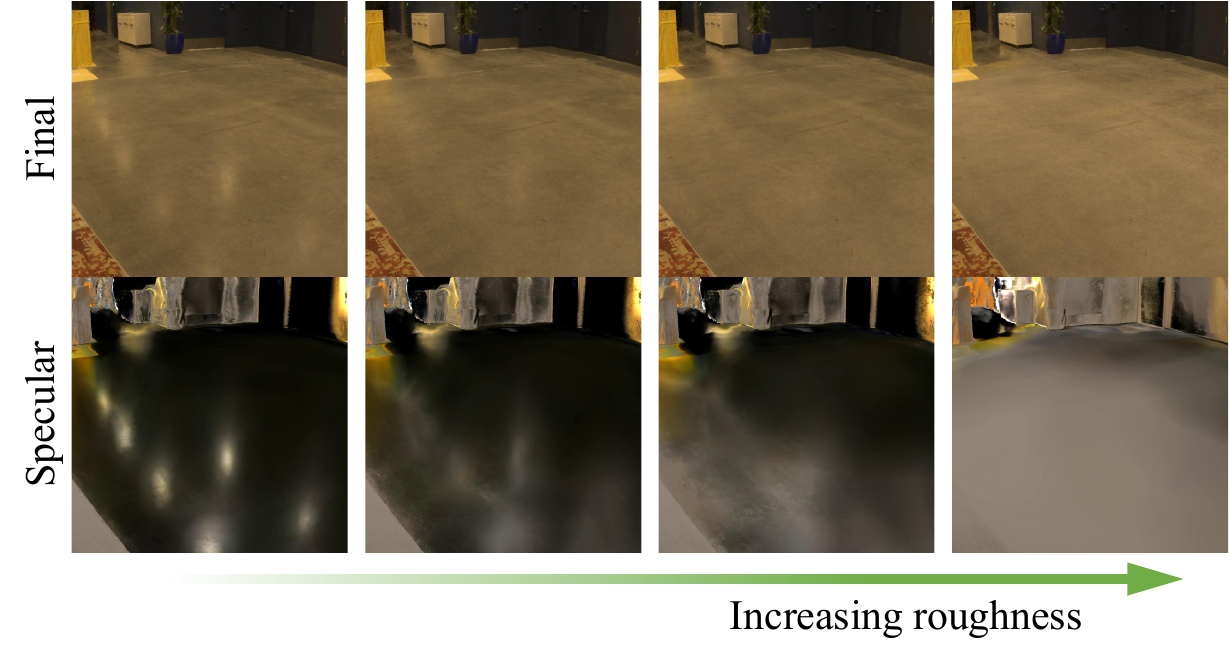}
    \vspace{-8mm}
    \caption{We can control the roughness of the scene by adding an offset to the input roughness.}
    \vspace{-2mm}
    \label{fig:roughness}
\end{figure}

\paragraph{Applications.}
Our primary goal was to improve the quality of novel-view synthesis with specular reflective surfaces.
We achieve this via our proposed Gaussian directional encoding that enables a meaningful decomposition of specular and diffuse components in a scene.
Moreover, this also enables applications other than novel-view synthesis, such as reflection removal, and surface roughness editing.
For instance, \cref{fig:decompositions} shows that we can easily remove reflections using the diffuse component.
Furthermore, \cref{fig:roughness} demonstrates an example of editing roughness.
By adding an offset to the predicted roughness during rendering, we can effectively manipulate the glossiness of the real surface.

\begin{figure}
  \centering
  \begin{subfigure}{0.49\linewidth}
    \caption*{Ours}
    \includegraphics[width=\linewidth]{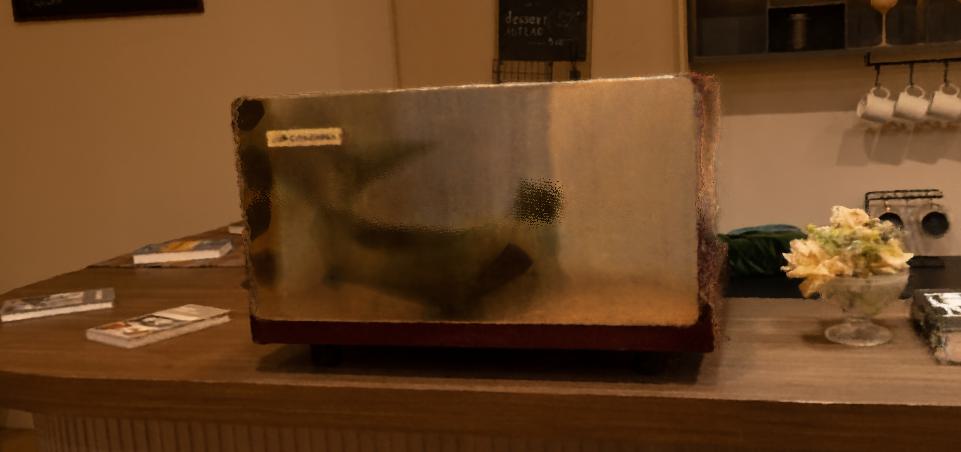}
  \end{subfigure}
  \hspace{-3pt}
  \begin{subfigure}{0.49\linewidth}
    \caption*{GT}
    \includegraphics[width=\linewidth]{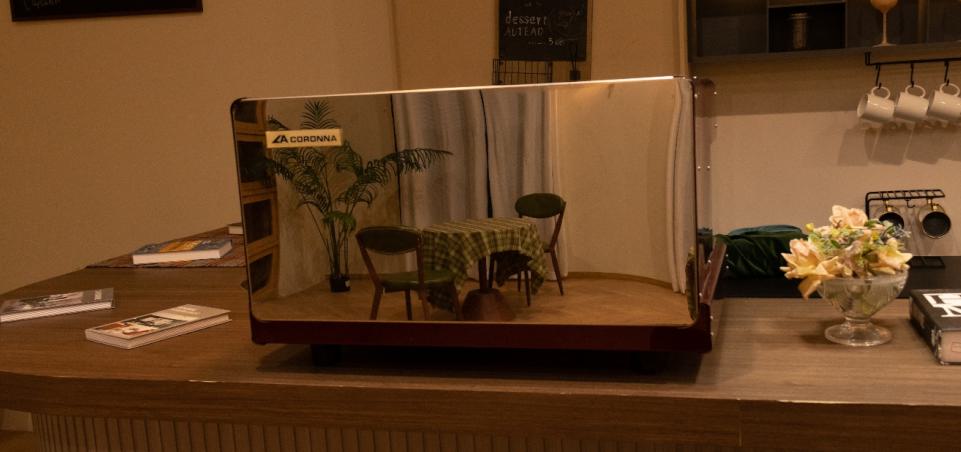}
  \end{subfigure}
  \vspace{-2mm}
  \caption{Our method cannot reconstruct mirror-like perfect reflections due to the limited capacity of the 3D Gaussian encoding.}
  \vspace{-4mm}
  \label{fig:limitation}
\end{figure}

\paragraph{Limitations}
While our method improves on existing baselines, it has some limitations.
As we parameterize the specular color via only several hundreds of Gaussians, the encoding is limited to relatively low frequency compared with perfect mirror-like reflections.
We show such a failure case in \cref{fig:limitation}.
We can see that our method is only able to learn a blurry version of the reflection.
This could be alleviated by using many more Gaussians, as demonstrated in 3D Gaussian splatting \cite{KerblKLD2023}.
However, the computational cost of traversing all Gaussians for every pixel quickly becomes prohibitive in our implementation.
More efficient traversal, such as by rasterization, could be interesting future work.

\paragraph{Conclusion}
In this paper, we proposed a pipeline to improve the existing approach in modeling and reconstructing view-dependent effects in a NeRF representation.
Central to our approach is a new Gaussian directional \encoding to enhance the capability of neural radiance fields to model specular reflections under near-field lighting.
We also utilize monocular normal supervision to help resolve shape--radiance ambiguity.
Experiments have demonstrated the effectiveness of each of our contributions.
We believe this work proposes a practical and effective solution for reconstructing NeRFs in room-scale scenes, specifically addressing the challenges of accurately capturing specular reflections.

\paragraph{Acknowledgments}
The authors from HKUST were partly supported by the Hong Kong Research Grants Council (RGC).
We would like to thank Linning Xu and Zhao Dong for helpful discussions.\par

{
    \small
    \bibliographystyle{ieeenat_fullname}
    \bibliography{main,SpecNeRF-CR}

\begin{thebibliography}{68}
\providecommand{\natexlab}[1]{#1}
\providecommand{\url}[1]{\texttt{#1}}
\expandafter\ifx\csname urlstyle\endcsname\relax
  \providecommand{\doi}[1]{doi: #1}\else
  \providecommand{\doi}{doi: \begingroup \urlstyle{rm}\Url}\fi

\bibitem[{Agisoft LLC}(2022)]{AgisoftMetashape}
{Agisoft LLC}.
\newblock {Agisoft Metashape Professional}.
\newblock {Computer software}, 2022.

\bibitem[Azinović et~al.(2019)Azinović, Li, Kaplanyan, and Nießner]{AzinoLKN2019}
Dejan Azinović, Tzu-Mao Li, Anton Kaplanyan, and Matthias Nießner.
\newblock Inverse path tracing for joint material and lighting estimation.
\newblock In \emph{CVPR}, 2019.

\bibitem[Barron et~al.(2022)Barron, Mildenhall, Verbin, Srinivasan, and Hedman]{mipnerf360}
Jonathan~T. Barron, Ben Mildenhall, Dor Verbin, Pratul~P. Srinivasan, and Peter Hedman.
\newblock {Mip-NeRF 360}: Unbounded anti-aliased neural radiance fields.
\newblock In \emph{CVPR}, 2022.

\bibitem[Bhat et~al.(2023)Bhat, Birkl, Wofk, Wonka, and Müller]{BhatBWWM2023}
Shariq~Farooq Bhat, Reiner Birkl, Diana Wofk, Peter Wonka, and Matthias Müller.
\newblock {ZoeDepth}: Zero-shot transfer by combining relative and metric depth.
\newblock \href{https://arxiv.org/abs/2302.12288}{arXiv:2302.12288}, 2023.

\bibitem[Bi et~al.(2020)Bi, Xu, Srinivasan, Mildenhall, Sunkavalli, Hašan, Hold-Geoffroy, Kriegman, and Ramamoorthi]{BiXSMSHHKR2020}
Sai Bi, Zexiang Xu, Pratul Srinivasan, Ben Mildenhall, Kalyan Sunkavalli, Miloš Hašan, Yannick Hold-Geoffroy, David Kriegman, and Ravi Ramamoorthi.
\newblock Neural reflectance fields for appearance acquisition.
\newblock \href{https://arxiv.org/abs/2008.03824}{arXiv:2008.03824}, 2020.

\bibitem[Boss et~al.(2021{\natexlab{a}})Boss, Braun, Jampani, Barron, Liu, and Lensch]{BossBJBLL2021}
Mark Boss, Raphael Braun, Varun Jampani, Jonathan~T. Barron, Ce Liu, and Hendrik~P.A. Lensch.
\newblock {NeRD}: Neural reflectance decomposition from image collections.
\newblock In \emph{ICCV}, 2021{\natexlab{a}}.

\bibitem[Boss et~al.(2021{\natexlab{b}})Boss, Jampani, Braun, Liu, Barron, and Lensch]{BossJBLBL2021}
Mark Boss, Varun Jampani, Raphael Braun, Ce Liu, Jonathan~T. Barron, and Hendrik~P.A. Lensch.
\newblock Neural-{PIL}: Neural pre-integrated lighting for reflectance decomposition.
\newblock In \emph{NeurIPS}, 2021{\natexlab{b}}.

\bibitem[Bradski(2000)]{opencv}
G. Bradski.
\newblock {The OpenCV Library}.
\newblock \emph{Dr. Dobb's Journal of Software Tools}, 2000.

\bibitem[Chen et~al.(2022)Chen, Xu, Geiger, Yu, and Su]{ChenXGYS2022}
Anpei Chen, Zexiang Xu, Andreas Geiger, Jingyi Yu, and Hao Su.
\newblock {TensoRF}: Tensorial radiance fields.
\newblock In \emph{ECCV}, 2022.

\bibitem[Cook and Torrance(1982)]{cooktorrance}
Robert~L Cook and Kenneth~E Torrance.
\newblock A reflectance model for computer graphics.
\newblock \emph{ACM Trans. Graph.}, 1\penalty0 (1):\penalty0 7--24, 1982.

\bibitem[Deng et~al.(2024)Deng, Li, Liu, and Yang]{DengLLY2024}
Youming Deng, Xueting Li, Sifei Liu, and Ming-Hsuan Yang.
\newblock {DIP}: Differentiable interreflection-aware physics-based inverse rendering.
\newblock In \emph{3DV}, 2024.

\bibitem[Eftekhar et~al.(2021)Eftekhar, Sax, Malik, and Zamir]{EftekSMZ2021}
Ainaz Eftekhar, Alexander Sax, Jitendra Malik, and Amir Zamir.
\newblock Omnidata: A scalable pipeline for making multi-task mid-level vision datasets from {3D} scans.
\newblock In \emph{ICCV}, 2021.

\bibitem[Gardner et~al.(2019)Gardner, Hold-Geoffroy, Sunkavalli, Gagne, and Lalonde]{indoorlight}
Marc-Andre Gardner, Yannick Hold-Geoffroy, Kalyan Sunkavalli, Christian Gagne, and Jean-Francois Lalonde.
\newblock Deep parametric indoor lighting estimation.
\newblock In \emph{ICCV}, 2019.

\bibitem[Ge et~al.(2023)Ge, Hu, Zhao, Liu, and Chen]{GeHZLC2023}
Wenhang Ge, Tao Hu, Haoyu Zhao, Shu Liu, and Ying-Cong Chen.
\newblock Ref-{NeuS}: Ambiguity-reduced neural implicit surface learning for multi-view reconstruction with reflection.
\newblock In \emph{ICCV}, 2023.

\bibitem[Guo et~al.(2022)Guo, Kang, Bao, He, and Zhang]{GuoKBHZ2022}
Yuan-Chen Guo, Di Kang, Linchao Bao, Yu He, and Song-Hai Zhang.
\newblock {NeRFReN}: Neural radiance fields with reflections.
\newblock In \emph{CVPR}, 2022.

\bibitem[Holland et~al.(2023)Holland, Bliersbach, Müller, Stotko, and Klein]{HollaBMSK2023}
Leif~Van Holland, Ruben Bliersbach, Jan~U. Müller, Patrick Stotko, and Reinhard Klein.
\newblock {TraM-NeRF}: Tracing mirror and near-perfect specular reflections through neural radiance fields.
\newblock \href{https://arxiv.org/abs/2310.10650}{arXiv:2310.10650}, 2023.

\bibitem[Jeong et~al.(2024)Jeong, Shin, and Park]{nerfactory}
Yoonwoo Jeong, Seungjoo Shin, and Kibaek Park.
\newblock {NeRF-Factory}: An awesome {PyTorch NeRF} collection, 2024.

\bibitem[Jin et~al.(2023)Jin, Liu, Xu, Zhang, Han, Bi, Zhou, Xu, and Su]{INV_TensoIR}
Haian Jin, Isabella Liu, Peijia Xu, Xiaoshuai Zhang, Songfang Han, Sai Bi, Xiaowei Zhou, Zexiang Xu, and Hao Su.
\newblock {TensoIR}: Tensorial inverse rendering.
\newblock In \emph{CVPR}, 2023.

\bibitem[Kajiya(1986)]{Kajiy1986}
James~T. Kajiya.
\newblock The rendering equation.
\newblock \emph{Computer Graphics (Proceedings of SIGGRAPH)}, 20\penalty0 (4):\penalty0 143--150, 1986.

\bibitem[Kerbl et~al.(2023)Kerbl, Kopanas, Leimkühler, and Drettakis]{KerblKLD2023}
Bernhard Kerbl, Georgios Kopanas, Thomas Leimkühler, and George Drettakis.
\newblock {3D Gaussian} splatting for real-time radiance field rendering.
\newblock \emph{ACM Trans. Graph.}, 42\penalty0 (4):\penalty0 139:1--14, 2023.

\bibitem[Kingma and Ba(2015)]{KingmB2015}
Diederik~P. Kingma and Jimmy Ba.
\newblock Adam: A method for stochastic optimization.
\newblock In \emph{ICLR}, 2015.

\bibitem[Kopanas et~al.(2022)Kopanas, Leimkühler, Rainer, Jambon, and Drettakis]{KopanLRJD2022}
Georgios Kopanas, Thomas Leimkühler, Gilles Rainer, Clément Jambon, and George Drettakis.
\newblock Neural point catacaustics for novel-view synthesis of reflections.
\newblock \emph{ACM Trans. Graph.}, 41\penalty0 (6):\penalty0 201:1--15, 2022.

\bibitem[Li and Li(2022)]{LiL2022}
Junxuan Li and Hongdong Li.
\newblock Neural reflectance for shape recovery with shadow handling.
\newblock In \emph{CVPR}, 2022.

\bibitem[Li et~al.(2022{\natexlab{a}})Li, Guo, Fei, Li, and Guo]{LiGFLG2022}
Quewei Li, Jie Guo, Yang Fei, Feichao Li, and Yanwen Guo.
\newblock {NeuLighting}: Neural lighting for free viewpoint outdoor scene relighting with unconstrained photo collections.
\newblock In \emph{SIGGRAPH Asia}, pages 13:1--9, 2022{\natexlab{a}}.

\bibitem[Li et~al.(2018)Li, Aittala, Durand, and Lehtinen]{LiADL2018}
Tzu-Mao Li, Miika Aittala, Frédo Durand, and Jaakko Lehtinen.
\newblock Differentiable {Monte Carlo} ray tracing through edge sampling.
\newblock \emph{ACM Trans. Graph.}, 37\penalty0 (6):\penalty0 222:1--11, 2018.

\bibitem[Li et~al.(2020)Li, Shafiei, Ramamoorthi, Sunkavalli, and Chandraker]{svlight}
Zhengqin Li, Mohammad Shafiei, Ravi Ramamoorthi, Kalyan Sunkavalli, and Manmohan Chandraker.
\newblock Inverse rendering for complex indoor scenes: Shape, spatially-varying lighting and {SVBRDF} from a single image.
\newblock In \emph{CVPR}, 2020.

\bibitem[Li et~al.(2022{\natexlab{b}})Li, Shi, Bi, Zhu, Sunkavalli, Ha\v{s}an, Xu, Ramamoorthi, and Chandraker]{lightediting}
Zhengqin Li, Jia Shi, Sai Bi, Rui Zhu, Kalyan Sunkavalli, Milo\v{s} Ha\v{s}an, Zexiang Xu, Ravi Ramamoorthi, and Manmohan Chandraker.
\newblock Physically-based editing of indoor scene lighting from a single image.
\newblock In \emph{ECCV}, 2022{\natexlab{b}}.

\bibitem[Li et~al.(2023)Li, M\"uller, Evans, Taylor, Unberath, Liu, and Lin]{neuralangelo}
Zhaoshuo Li, Thomas M\"uller, Alex Evans, Russell~H Taylor, Mathias Unberath, Ming-Yu Liu, and Chen-Hsuan Lin.
\newblock Neuralangelo: High-fidelity neural surface reconstruction.
\newblock In \emph{CVPR}, 2023.

\bibitem[Liang et~al.(2022)Liang, Zhang, Li, Yang, Guan, and Vijaykumar]{LiangZLYGV2022}
Ruofan Liang, Jiahao Zhang, Haoda Li, Chen Yang, Yushi Guan, and Nandita Vijaykumar.
\newblock {SPIDR}: {SDF}-based neural point fields for illumination and deformation.
\newblock \href{https://arxiv.org/abs/2210.08398}{arXiv:2210.08398}, 2022.

\bibitem[Liu et~al.(2023{\natexlab{a}})Liu, Tai, and Tang]{LiuTT2023}
Xinhang Liu, Yu-Wing Tai, and Chi-Keung Tang.
\newblock Clean-{NeRF}: Reformulating {NeRF} to account for view-dependent observations.
\newblock \href{https://arxiv.org/abs/2303.14707}{arXiv:2303.14707}, 2023{\natexlab{a}}.

\bibitem[Liu et~al.(2023{\natexlab{b}})Liu, Wang, Lin, Long, Wang, Liu, Komura, and Wang]{LiuWLLWLKW2023}
Yuan Liu, Peng Wang, Cheng Lin, Xiaoxiao Long, Jiepeng Wang, Lingjie Liu, Taku Komura, and Wenping Wang.
\newblock {NeRO}: Neural geometry and {BRDF} reconstruction of reflective objects from multiview images.
\newblock \emph{ACM Trans. Graph.}, pages 114:1--22, 2023{\natexlab{b}}.

\bibitem[Lyu et~al.(2022)Lyu, Tewari, Leimkuehler, Habermann, and Theobalt]{INV_NRTF}
Linjie Lyu, Ayush Tewari, Thomas Leimkuehler, Marc Habermann, and Christian Theobalt.
\newblock Neural radiance transfer fields for relightable novel-view synthesis with global illumination.
\newblock In \emph{ECCV}, 2022.

\bibitem[Mai et~al.(2023)Mai, Verbin, Kuester, and Fridovich-Keil]{MaiVKF2023}
Alexander Mai, Dor Verbin, Falko Kuester, and Sara Fridovich-Keil.
\newblock Neural microfacet fields for inverse rendering.
\newblock In \emph{ICCV}, 2023.

\bibitem[Mildenhall et~al.(2020)Mildenhall, Srinivasan, Tancik, Barron, Ramamoorthi, and Ng]{MildeSTBRN2020}
Ben Mildenhall, Pratul~P. Srinivasan, Matthew Tancik, Jonathan~T. Barron, Ravi Ramamoorthi, and Ren Ng.
\newblock {NeRF}: Representing scenes as neural radiance fields for view synthesis.
\newblock In \emph{ECCV}, 2020.

\bibitem[Miller and Hoffman(1984)]{IBL_sig84}
Gene~S Miller and CR Hoffman.
\newblock Illumination and reflection maps.
\newblock In \emph{ACM SIGGRAPH}, 1984.

\bibitem[Müller et~al.(2022)Müller, Evans, Schied, and Keller]{MuelleESK2022}
Thomas Müller, Alex Evans, Christoph Schied, and Alexander Keller.
\newblock Instant neural graphics primitives with a multiresolution hash encoding.
\newblock \emph{ACM Trans. Graph.}, 41\penalty0 (4):\penalty0 102:1--15, 2022.

\bibitem[Paszke et~al.(2019)Paszke, Gross, Massa, Lerer, Bradbury, Chanan, Killeen, Lin, Gimelshein, Antiga, Desmaison, Kopf, Yang, DeVito, Raison, Tejani, Chilamkurthy, Steiner, Fang, Bai, and Chintala]{pytorch}
Adam Paszke, Sam Gross, Francisco Massa, Adam Lerer, James Bradbury, Gregory Chanan, Trevor Killeen, Zeming Lin, Natalia Gimelshein, Luca Antiga, Alban Desmaison, Andreas Kopf, Edward Yang, Zachary DeVito, Martin Raison, Alykhan Tejani, Sasank Chilamkurthy, Benoit Steiner, Lu Fang, Junjie Bai, and Soumith Chintala.
\newblock {PyTorch}: An imperative style, high-performance deep learning library.
\newblock In \emph{NeurIPS}, 2019.

\bibitem[Philip et~al.(2021)Philip, Morgenthaler, Gharbi, and Drettakis]{PhiliMGD2021}
Julien Philip, Sébastien Morgenthaler, Michaël Gharbi, and George Drettakis.
\newblock Free-viewpoint indoor neural relighting from multi-view stereo.
\newblock \emph{ACM Trans. Graph.}, 40\penalty0 (5):\penalty0 194:1--18, 2021.

\bibitem[Ranftl et~al.(2022)Ranftl, Lasinger, Hafner, Schindler, and Koltun]{RanftLHSK2022}
René Ranftl, Katrin Lasinger, David Hafner, Konrad Schindler, and Vladlen Koltun.
\newblock Towards robust monocular depth estimation: Mixing datasets for zero-shot cross-dataset transfer.
\newblock \emph{TPAMI}, 44\penalty0 (3):\penalty0 1623--1637, 2022.

\bibitem[Rudnev et~al.(2022)Rudnev, Elgharib, Smith, Liu, Golyanik, and Theobalt]{INV_Outdoor_NeRFRelit}
Viktor Rudnev, Mohamed Elgharib, William Smith, Lingjie Liu, Vladislav Golyanik, and Christian Theobalt.
\newblock {NeRF} for outdoor scene relighting.
\newblock In \emph{ECCV}, 2022.

\bibitem[Srinivasan et~al.(2020)Srinivasan, Mildenhall, Tancik, Barron, Tucker, and Snavely]{lighthouse}
Pratul~P. Srinivasan, Ben Mildenhall, Matthew Tancik, Jonathan~T. Barron, Richard Tucker, and Noah Snavely.
\newblock Lighthouse: Predicting lighting volumes for spatially-coherent illumination.
\newblock In \emph{CVPR}, 2020.

\bibitem[Srinivasan et~al.(2021)Srinivasan, Deng, Zhang, Tancik, Mildenhall, and Barron]{SriniDZTMB2021}
Pratul~P. Srinivasan, Boyang Deng, Xiuming Zhang, Matthew Tancik, Ben Mildenhall, and Jonathan~T. Barron.
\newblock {NeRV}: Neural reflectance and visibility fields for relighting and view synthesis.
\newblock In \emph{CVPR}, 2021.

\bibitem[Tancik et~al.(2020)Tancik, Srinivasan, Mildenhall, Fridovich-Keil, Raghavan, Singhal, Ramamoorthi, Barron, and Ng]{TanciSMFRSRBN2020}
Matthew Tancik, Pratul~P. Srinivasan, Ben Mildenhall, Sara Fridovich-Keil, Nithin Raghavan, Utkarsh Singhal, Ravi Ramamoorthi, Jonathan~T. Barron, and Ren Ng.
\newblock Fourier features let networks learn high frequency functions in low dimensional domains.
\newblock In \emph{NeurIPS}, 2020.

\bibitem[Tancik et~al.(2023)Tancik, Weber, Ng, Li, Yi, Wang, Kristoffersen, Austin, Salahi, Ahuja, Mcallister, Kerr, and Kanazawa]{TanciWNLYWKASAMKK2023}
Matthew Tancik, Ethan Weber, Evonne Ng, Ruilong Li, Brent Yi, Terrance Wang, Alexander Kristoffersen, Jake Austin, Kamyar Salahi, Abhik Ahuja, David Mcallister, Justin Kerr, and Angjoo Kanazawa.
\newblock Nerfstudio: A modular framework for neural radiance field development.
\newblock In \emph{SIGGRAPH}, pages 72:1--12, 2023.

\bibitem[Tewari et~al.(2022)Tewari, Thies, Mildenhall, Srinivasan, Tretschk, Wang, Lassner, Sitzmann, Martin-Brualla, Lombardi, Simon, Theobalt, Niessner, Barron, Wetzstein, Zollhöfer, and Golyanik]{TewarTMSTWLSMLSTNBWZG2022}
Ayush Tewari, Justus Thies, Ben Mildenhall, Pratul Srinivasan, Edgar Tretschk, Yifan Wang, Christoph Lassner, Vincent Sitzmann, Ricardo Martin-Brualla, Stephen Lombardi, Tomas Simon, Christian Theobalt, Matthias Niessner, Jonathan~T. Barron, Gordon Wetzstein, Michael Zollhöfer, and Vladislav Golyanik.
\newblock Advances in neural rendering.
\newblock \emph{Comput. Graph. Forum}, 41\penalty0 (2):\penalty0 703--735, 2022.

\bibitem[Tiwary et~al.(2023)Tiwary, Dave, Behari, Klinghoffer, Veeraraghavan, and Raskar]{TiwarDBKVR2023}
Kushagra Tiwary, Akshat Dave, Nikhil Behari, Tzofi Klinghoffer, Ashok Veeraraghavan, and Ramesh Raskar.
\newblock {ORCa}: Glossy objects as radiance field cameras.
\newblock In \emph{CVPR}, 2023.

\bibitem[Verbin et~al.(2022)Verbin, Hedman, Mildenhall, Zickler, Barron, and Srinivasan]{VerbiHMZBS2022}
Dor Verbin, Peter Hedman, Ben Mildenhall, Todd Zickler, Jonathan~T. Barron, and Pratul~P. Srinivasan.
\newblock {Ref-NeRF}: Structured view-dependent appearance for neural radiance fields.
\newblock In \emph{CVPR}, 2022.

\bibitem[Wang et~al.(2023)Wang, Shen, Gao, Huang, Munkberg, Hasselgren, Gojcic, Chen, and Fidler]{WangSGHMHGCF2023}
Zian Wang, Tianchang Shen, Jun Gao, Shengyu Huang, Jacob Munkberg, Jon Hasselgren, Zan Gojcic, Wenzheng Chen, and Sanja Fidler.
\newblock Neural fields meet explicit geometric representations for inverse rendering of urban scenes.
\newblock In \emph{CVPR}, 2023.

\bibitem[Wu et~al.(2023)Wu, Zhu, Yaldiz, Zhu, Cai, Matai, Porikli, Li, Chandraker, and Ramamoorthi]{RT_FIPT}
Liwen Wu, Rui Zhu, Mustafa~B. Yaldiz, Yinhao Zhu, Hong Cai, Janarbek Matai, Fatih Porikli, Tzu-Mao Li, Manmohan Chandraker, and Ravi Ramamoorthi.
\newblock Factorized inverse path tracing for efficient and accurate material-lighting estimation.
\newblock In \emph{ICCV}, 2023.

\bibitem[Wu et~al.(2022)Wu, Xu, Zhu, Bao, Huang, Tompkin, and Xu]{WuXZBHTX2022}
Xiuchao Wu, Jiamin Xu, Zihan Zhu, Hujun Bao, Qixing Huang, James Tompkin, and Weiwei Xu.
\newblock Scalable neural indoor scene rendering.
\newblock \emph{ACM Trans. Graph.}, 41\penalty0 (4):\penalty0 98:1--16, 2022.

\bibitem[Xie et~al.(2022)Xie, Takikawa, Saito, Litany, Yan, Khan, Tombari, Tompkin, Sitzmann, and Sridhar]{XieTSLYKTTSS2022}
Yiheng Xie, Towaki Takikawa, Shunsuke Saito, Or Litany, Shiqin Yan, Numair Khan, Federico Tombari, James Tompkin, Vincent Sitzmann, and Srinath Sridhar.
\newblock Neural fields in visual computing and beyond.
\newblock \emph{Comput. Graph. Forum}, 2022.

\bibitem[Xu et~al.(2021)Xu, Wu, Zhu, Huang, Yang, Bao, and Xu]{XuWZHYBX2021}
Jiamin Xu, Xiuchao Wu, Zihan Zhu, Qixing Huang, Yin Yang, Hujun Bao, and Weiwei Xu.
\newblock Scalable image-based indoor scene rendering with reflections.
\newblock \emph{ACM Trans. Graph.}, 40\penalty0 (4):\penalty0 60:1--14, 2021.

\bibitem[Xu et~al.(2023)Xu, Agrawal, Laney, Garcia, Bansal, Kim, Rota~Bulò, Porzi, Kontschieder, Božič, Lin, Zollhöfer, and Richardt]{XuALGBKRPKBLZR2023}
Linning Xu, Vasu Agrawal, William Laney, Tony Garcia, Aayush Bansal, Changil Kim, Samuel Rota~Bulò, Lorenzo Porzi, Peter Kontschieder, Aljaž Božič, Dahua Lin, Michael Zollhöfer, and Christian Richardt.
\newblock {VR-NeRF}: High-fidelity virtualized walkable spaces.
\newblock In \emph{SIGGRAPH Asia}, 2023.

\bibitem[Yan et~al.(2023)Yan, Li, and Lee]{YanLL2023}
Zhiwen Yan, Chen Li, and Gim~Hee Lee.
\newblock {NeRF-DS}: Neural radiance fields for dynamic specular objects.
\newblock In \emph{CVPR}, 2023.

\bibitem[Yang et~al.(2023)Yang, Pavone, and Wang]{FreeNeRF}
Jiawei Yang, Marco Pavone, and Yue Wang.
\newblock {FreeNeRF}: Improving few-shot neural rendering with free frequency regularization.
\newblock In \emph{CVPR}, 2023.

\bibitem[Yao et~al.(2022)Yao, Zhang, Liu, Qu, Fang, McKinnon, Tsin, and Quan]{YaoZLQFMTQ2022}
Yao Yao, Jingyang Zhang, Jingbo Liu, Yihang Qu, Tian Fang, David McKinnon, Yanghai Tsin, and Long Quan.
\newblock {NeILF}: Neural incident light field for physically-based material estimation.
\newblock In \emph{ECCV}, 2022.

\bibitem[Yin et~al.(2023)Yin, Qiu, Cheng, and Ren]{YinQCR2023}
Ze-Xin Yin, Jiaxiong Qiu, Ming-Ming Cheng, and Bo Ren.
\newblock Multi-space neural radiance fields.
\newblock In \emph{CVPR}, 2023.

\bibitem[Yu et~al.(2023)Yu, Yang, Cui, Dong, Chen, and Shi]{INV_Outdoor_MILO}
Bohan Yu, Siqi Yang, Xuanning Cui, Siyan Dong, Baoquan Chen, and Boxin Shi.
\newblock {MILO}: Multi-bounce inverse rendering for indoor scene with light-emitting objects.
\newblock \emph{IEEE TPAMI}, 45\penalty0 (8):\penalty0 10129--10142, 2023.

\bibitem[Yu et~al.(2022)Yu, Peng, Niemeyer, Sattler, and Geiger]{YuPNSG2022}
Zehao Yu, Songyou Peng, Michael Niemeyer, Torsten Sattler, and Andreas Geiger.
\newblock {MonoSDF}: Exploring monocular geometric cues for neural implicit surface reconstruction.
\newblock In \emph{NeurIPS}, 2022.

\bibitem[Zeng et~al.(2023{\natexlab{a}})Zeng, Chen, Dong, Peers, Wu, and Tong]{ZengCDPWT2023}
Chong Zeng, Guojun Chen, Yue Dong, Pieter Peers, Hongzhi Wu, and Xin Tong.
\newblock Relighting neural radiance fields with shadow and highlight hints.
\newblock In \emph{SIGGRAPH}, 2023{\natexlab{a}}.

\bibitem[Zeng et~al.(2023{\natexlab{b}})Zeng, Bao, Chen, Dong, Zhang, Bao, and Cui]{ZengBCDZBC2023}
Junyi Zeng, Chong Bao, Rui Chen, Zilong Dong, Guofeng Zhang, Hujun Bao, and Zhaopeng Cui.
\newblock Mirror-{NeRF}: Learning neural radiance fields for mirrors with {Whitted}-style ray tracing.
\newblock In \emph{ACM Multimedia}, 2023{\natexlab{b}}.

\bibitem[Zhang et~al.(2023{\natexlab{a}})Zhang, Yao, Li, Liu, Fang, McKinnon, Tsin, and Quan]{ZhangYLLFMTQ2023}
Jingyang Zhang, Yao Yao, Shiwei Li, Jingbo Liu, Tian Fang, David McKinnon, Yanghai Tsin, and Long Quan.
\newblock {NeILF}++: Inter-reflectable light fields for geometry and material estimation.
\newblock In \emph{ICCV}, 2023{\natexlab{a}}.

\bibitem[Zhang et~al.(2021{\natexlab{a}})Zhang, Luan, Wang, Bala, and Snavely]{ZhangLWBS2021}
Kai Zhang, Fujun Luan, Qianqian Wang, Kavita Bala, and Noah Snavely.
\newblock {PhySG}: Inverse rendering with spherical {Gaussians} for physics-based material editing and relighting.
\newblock In \emph{CVPR}, 2021{\natexlab{a}}.

\bibitem[Zhang et~al.(2018)Zhang, Isola, Efros, Shechtman, and Wang]{LPIPS}
Richard Zhang, Phillip Isola, Alexei~A Efros, Eli Shechtman, and Oliver Wang.
\newblock The unreasonable effectiveness of deep features as a perceptual metric.
\newblock In \emph{CVPR}, 2018.

\bibitem[Zhang et~al.(2021{\natexlab{b}})Zhang, Srinivasan, Deng, Debevec, Freeman, and Barron]{ZhangSDDFB2021}
Xiuming Zhang, Pratul~P. Srinivasan, Boyang Deng, Paul Debevec, William~T. Freeman, and Jonathan~T. Barron.
\newblock {NeRFactor}: Neural factorization of shape and reflectance under an unknown illumination.
\newblock \emph{ACM Trans. Graph.}, 40\penalty0 (6):\penalty0 237:1--18, 2021{\natexlab{b}}.

\bibitem[Zhang et~al.(2022)Zhang, Sun, He, Fu, Jia, and Zhou]{INV_invrender}
Yuanqing Zhang, Jiaming Sun, Xingyi He, Huan Fu, Rongfei Jia, and Xiaowei Zhou.
\newblock Modeling indirect illumination for inverse rendering.
\newblock In \emph{CVPR}, 2022.

\bibitem[Zhang et~al.(2023{\natexlab{b}})Zhang, Xu, Yu, Ye, Wang, Jing, Yu, and Yang]{ZhangXYYWJYY2023}
Youjia Zhang, Teng Xu, Junqing Yu, Yuteng Ye, Junle Wang, Yanqing Jing, Jingyi Yu, and Wei Yang.
\newblock {NeMF}: Inverse volume rendering with neural microflake field.
\newblock In \emph{ICCV}, 2023{\natexlab{b}}.

\bibitem[Zhuang et~al.(2024)Zhuang, Zhang, Wang, Zhu, Feng, Li, Shan, and Cao]{ZhuanZWZFLSC2024}
Yiyu Zhuang, Qi Zhang, Xuan Wang, Hao Zhu, Ying Feng, Xiaoyu Li, Ying Shan, and Xun Cao.
\newblock {NeAI}: A pre-convoluted representation for plug-and-play neural ambient illumination.
\newblock In \emph{AAAI}, 2024.

\end{thebibliography}
}

\clearpage
\setcounter{page}{1}
\maketitlesupplementary
\appendix

\section{Supplementary Video}
For more information regarding the method, please visit our project website at \url{https://limacv.github.io/SpecNeRF_web/}. We also provide a supplementary video for visual comparisons under a moving camera trajectory, which can be accessed at \url{https://youtu.be/3nUooe3pVA0}.
We highly encourage readers to watch our video, where our method produces results with better specular reflection reconstruction.

\section{Gaussian Directional Encoding Proofs}
Recall that we define each Gaussian as:
\begin{equation}
    \mathcal{G}(\pos) = \exp \!\left( -\norm{
    	\mathcal{Q}(\pos - \mean; \quat) \odot \scale^{-1}
    }^2_2 \right) \text{,}
    \label{eq:supple_gau}
\end{equation}

where $\mean$ is the position and $\scale$ the scale of the Gaussian.
$\mathcal{Q}(\pos; \quat)$ applies the quaternion rotation $\quat$ to a 3D vector $\pos$.
For ease of notation, we omit the subscript $i$ (compared to the main paper)
as the same equation is applied to every Gaussian.
In practice, we optimize the inverse scale $\invscale = \scale^{-1}$ instead of directly using $\scale$, to improve numerical stability.

We further define the basis function $\mathcal{P}(\origin, \dir)$ over a given ray $\origin {+} t \dir$ with the Gaussian parameters $(\mean, \scale, \quat)$ as:
\begin{equation}
    \mathcal{P}(\origin, \dir) = \max_{t \geq 0}\mathcal{G}(\origin + t \dir) \text{.}
\end{equation}
We start by applying the following variable substitution that converts the ray origin $\origin$ and direction $\dir$ from world-space into the space of the Gaussian (origin $\overline{\origin}$ and direction $\overline{\dir}$):
\begin{align}
    \overline{\origin} &= \mathcal{Q}(\origin - \mean; \quat) \odot \invscale \text{,} \\
    \overline{\dir} &= \mathcal{Q}(\dir; \quat) \odot \invscale \text{.}
\end{align}
It follows that
\begin{align}
    \mathcal{G}(\origin + t \dir) &= \exp \!\left( -\norm{
    	\mathcal{Q}(\origin + t \dir - \mean; \quat) \odot \invscale
    }^2_2 \right) \\
    &= \exp \!\left(- \norm{\overline{\origin} + t  \overline{\dir}}_2^2 \right) \\
    &= \exp \!\left(- \overline{\origin}^\top \overline{\origin} - 2 \overline{\origin}^\top \overline{\dir} t - \overline{\dir}^\top \overline{\dir} t^2\right).
    \label{eq:quad}
\end{align}
Since the exponential function is monotonic, $\mathcal{G}$ is maximized when the quadratic function (in $t$)
\begin{equation}
f(t) = - \overline{\origin}^\top \overline{\origin} - 2 \overline{\origin}^\top \overline{\dir} t - \overline{\dir}^\top \overline{\dir} t^2
\end{equation}
reaches its maximum.
Since the quadratic coefficient, $-\overline{\dir}^\top \overline{\dir}$, is negative for any non-zero vector $\overline{\dir}$, $f(t)$ reaches its maximum when
$f'(t) = 0$, i.e. for
$t = t_0 = -\frac{ \overline{\origin}^\top \overline{\dir}}{\overline{\dir}^\top \overline{\dir}}$.
Furthermore, $\mathcal{G}(\origin + t \dir)$ monotonically decreases for $t \geq t_0$.
Given that $t \geq 0$, when $t_0 \leq 0$, the $\mathcal{G}(\origin + t \dir)$ reaches maximum always at $t = 0$.
To sum up, the maximum value of $\mathcal{G}(\origin + t \dir)$ falls into the following two cases:
\begin{equation}
    \max_{t \geq 0}\mathcal{G}(\origin + t \dir) = \begin{cases}
	    \exp\!\left(
        - \norm{\overline{\origin} + t_0 \overline{\dir}}_2^2
        \right)
	    & t_0 > 0 \\
	    \exp\!\left(
        - \norm{\overline{\origin}}_2^2
        \right)
	    & \text{otherwise,}
    \end{cases}
\end{equation}
By substituting $-\frac{ \overline{\origin}^\top \overline{\dir}}{\overline{\dir}^\top \overline{\dir}}$ for $t_0$, this equation is the same as \cref{eq:gau_embed} in the main paper.

\section{Implementation details}
\label{sec:supplementary}

\begin{figure*}
    \centering
    \includegraphics[width=\textwidth]{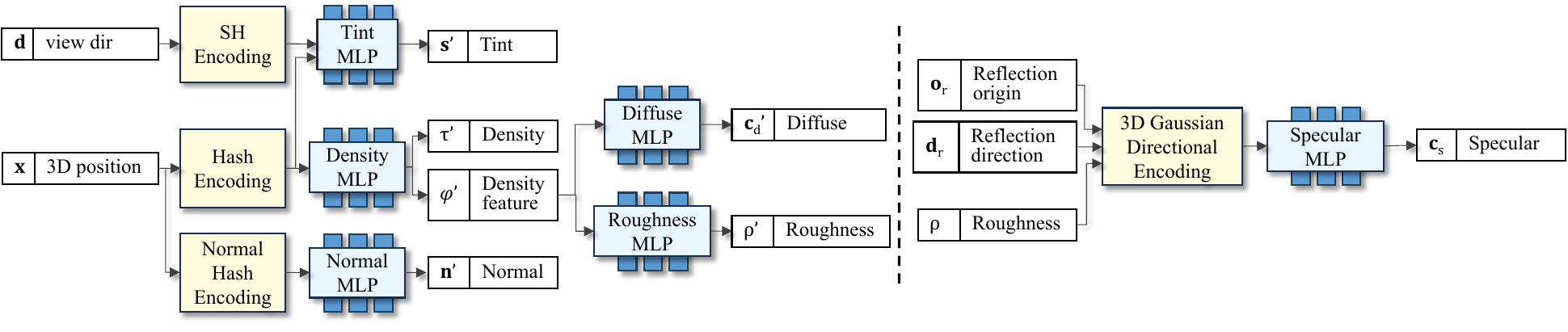}
    \caption{We zoom in the MLPs and some important modules as in \cref{fig:overview}. The detailed module configurations are shown in \cref{tab:supple_model_detail}.}
    \label{fig:supple_model_detail}
\end{figure*}

\Cref{fig:supple_model_detail} zooms into our model's network architecture and clarifies the role of each used MLP.

\begin{table}
	\caption{The value for each parameter.
		The module names are consistent with those shown in \cref{fig:supple_model_detail}.
		For all MLPs, we use ReLU activations in hidden layers.}
    \centering
    \small
    \begin{tabular}{c|l|c}
    \toprule
        \textbf{Module} & \textbf{Configuration} & \textbf{Value} \\
    \midrule
        \multirow{1}{*}{SH Encoding} & Order & $3$ \\
    \midrule
        \multirow{3}{*}{\makecell{Tint MLP}} & \# of hidden layer & $2$ \\
         & \# of neuron per layer & $64$ \\
         & Output activation & Sigmoid \\
    \midrule
        \multirow{5}{*}{\makecell{Hash\\Encoding}} & \# of levels & $16$ \\
         & Hash table size & $2^{22}$ \\
         & \# of feature dim. per entry & $2$ \\
         & Coarse resolution & $128$ \\
         & Scale factor per level & $1.4$ \\
    \midrule
        \multirow{4}{*}{\makecell{Density\\MLP}} & \# of hidden layer & $1$ \\
         & \# of neuron per layer & $64$ \\
         & Output activation & Exp \\
         & Density feature dim. & $16$ \\
    \midrule
        \multirow{3}{*}{\makecell{Diffuse\\MLP}} & \# of hidden layer & $2$ \\
         & \# of neuron per layer & $64$ \\
         & Output activation & Sigmoid \\
    \midrule
        \multirow{3}{*}{\makecell{Roughness\\MLP}} & \# of hidden layer & $2$ \\
         & \# of neuron per layer & $64$ \\
         & Output activation & Softplus \\
    \midrule
        \multirow{5}{*}{\makecell{Normal\\Hash\\Encoding}} & \# of levels & $4$ \\
         & Hash table size & $2^{19}$ \\
         & \# of feature dim. per entry & $4$ \\
         & Coarse resolution & $16$ \\
         & Scale factor per level & $1.5$ \\
    \midrule
        \multirow{3}{*}{\makecell{Normal\\MLP}} & \# of hidden layer & $1$ \\
         & \# of neuron per layer & $64$ \\
         & Output activation & None \\
    \midrule
        \multirow{3}{*}{\makecell{Specular\\MLP}} & \# of hidden layer & $2$ \\
         & \# of neuron per layer & $64$ \\
         & Output activation & Sigmoid \\

    \bottomrule
    \end{tabular}
    \label{tab:supple_model_detail}
\end{table}

\subsection{Model Structure}
We list the model structure parameters in \cref{tab:supple_model_detail}.
We use separate MLP heads to predict each property at each sample location.
Note that we use a lower resolution configuration for normal hash encoding, because we find that constraining the smoothness of the normal stabilizes the optimization process and leads to better specular reflection reconstruction.

\begin{figure}
  \centering
  
  \rotatebox{90}{\footnotesize \phantom{g}}
  \begin{subfigure}{0.31\linewidth}\hspace{4pt}
  	\caption*{Ground Truth}
  \end{subfigure}
  \hspace{-3pt}
  \begin{subfigure}{0.31\linewidth}
  	\caption*{w/o normal corr.\phantom{G}}
  \end{subfigure}
  \hspace{-3pt}
  \begin{subfigure}{0.31\linewidth}
  	\caption*{w/ normal corr.\phantom{G}}
  \end{subfigure}
  
  \rotatebox{90}{\hspace{10mm} \footnotesize Image}
  \begin{subfigure}{0.31\linewidth}
    \centering
    \includegraphics[width=\linewidth]{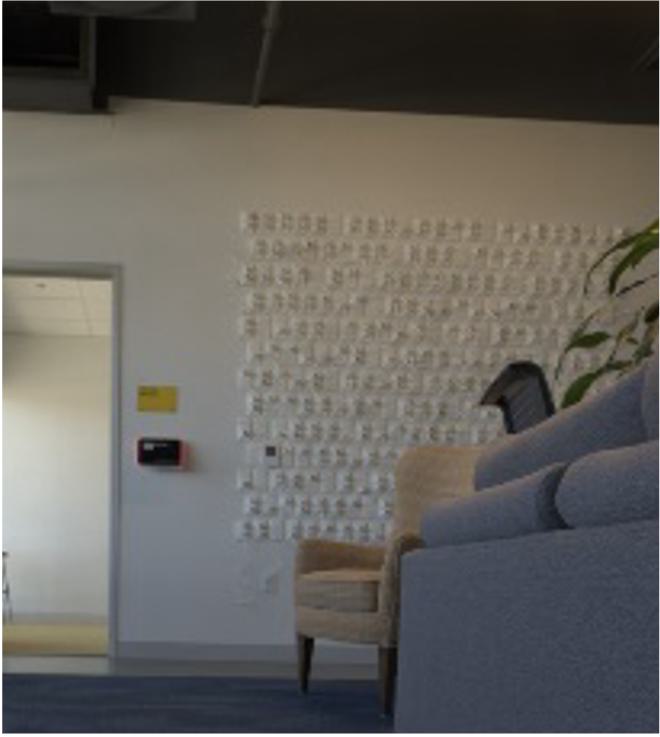}
  \end{subfigure}
  \hspace{-3pt}
  \begin{subfigure}{0.31\linewidth}
    \includegraphics[width=\linewidth]{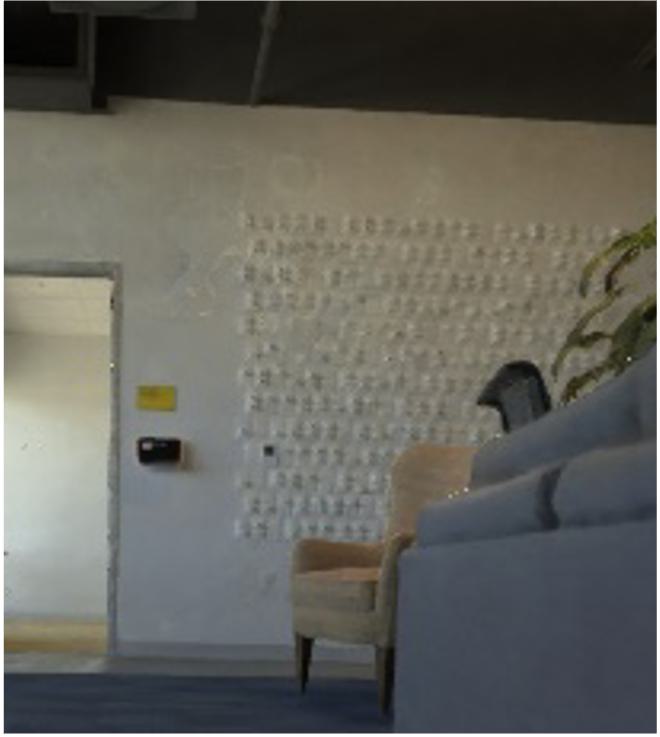}
  \end{subfigure}
  \hspace{-3pt}
  \begin{subfigure}{0.31\linewidth}
    \includegraphics[width=\linewidth]{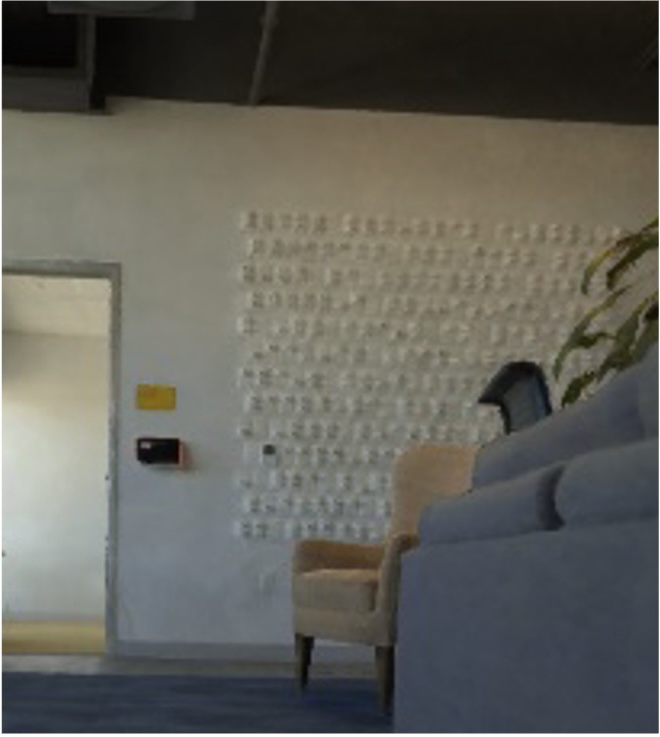}
  \end{subfigure}

  \begin{subfigure}{0.31\linewidth}
    \includegraphics[width=\linewidth]{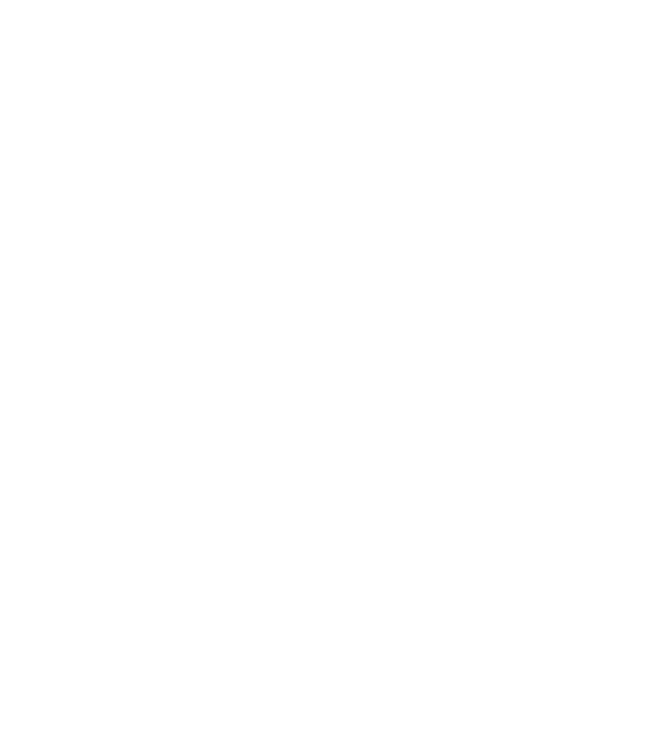}
  \end{subfigure}
  \rotatebox{90}{\hspace{11pt} \footnotesize Predicted normal\phantom{g}}
  \hspace{-3pt}
  \begin{subfigure}{0.31\linewidth}
    \includegraphics[width=\linewidth]{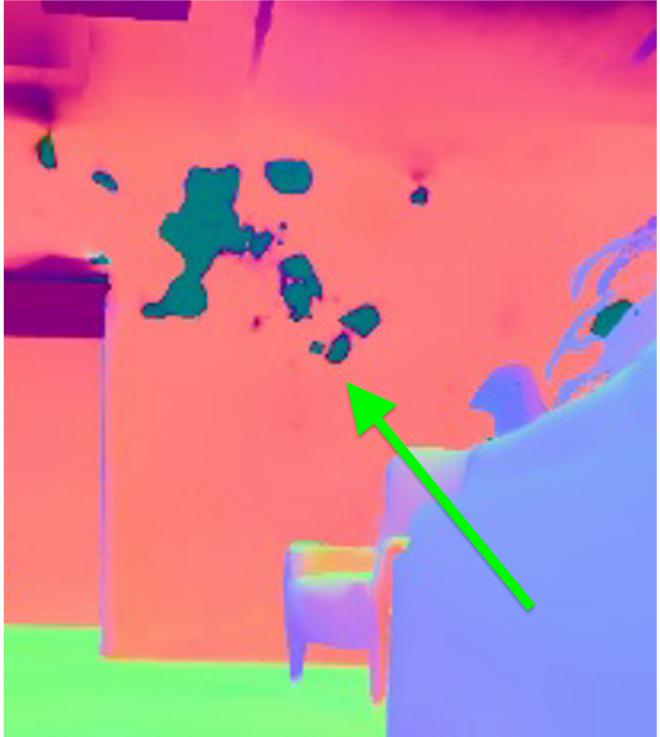}
  \end{subfigure}
  \hspace{-3pt}
  \begin{subfigure}{0.31\linewidth}
    \includegraphics[width=\linewidth]{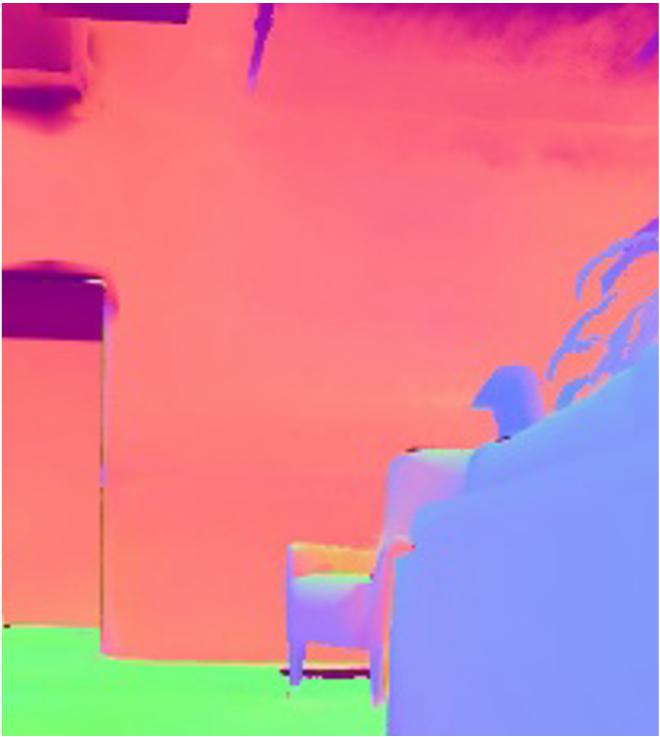}
  \end{subfigure}

  \begin{subfigure}{0.31\linewidth}
    \includegraphics[width=\linewidth]{fig_and_table/supple/supple_normalflip_placeholder.jpg}
  \end{subfigure}
  \rotatebox{90}{\hspace{8pt} \footnotesize Density gradient \phantom{g}}
  \hspace{-3pt}
  \begin{subfigure}{0.31\linewidth}
    \includegraphics[width=\linewidth]{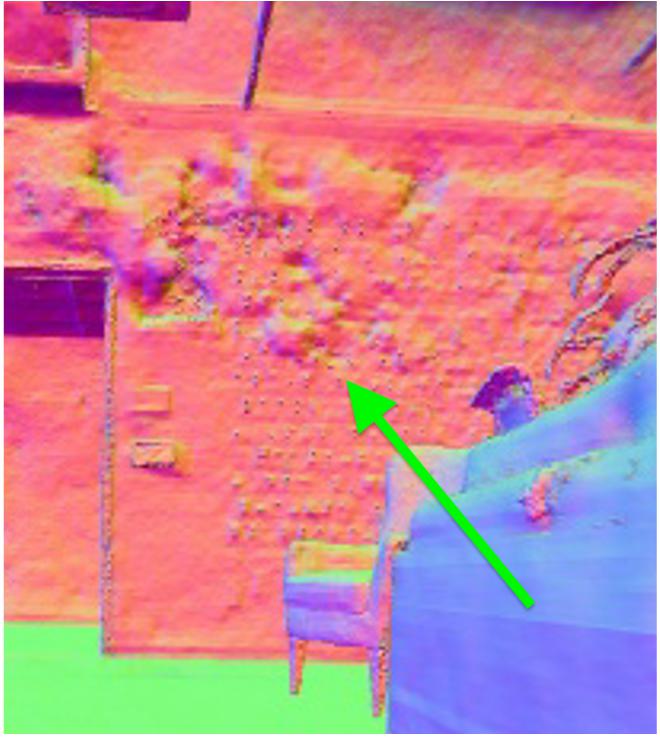}
  \end{subfigure}
  \hspace{-3pt}
  \begin{subfigure}{0.31\linewidth}
    \includegraphics[width=\linewidth]{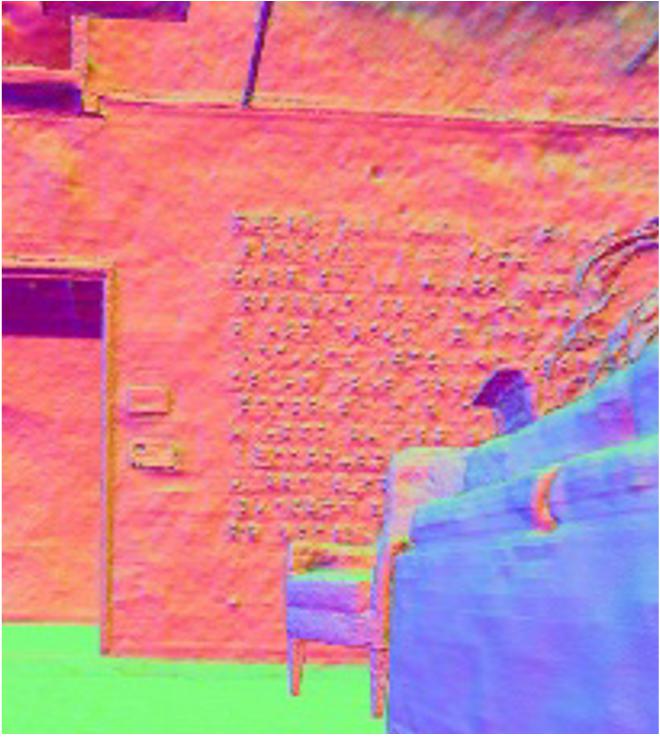}
  \end{subfigure}
  
  \vspace{-2mm}
    \caption{%
    	An example of the normal flip issue.
    	As indicated by the green arrow, the predicted normal is occasionally flipped due to small perturbation during training, which leads to artifacts in rendering images and the density gradient.
    	Our normal correction (normal corr.) prevents the flip issue by optionally reversing the normal direction based on the view direction.
     }
    \label{fig:supple_normalflip}
\end{figure}

\paragraph{Normal parameterization}
To predict normals, we first output a 3-element vector $\normal_\text{raw}'$ using the normal MLP without any output activation and normalize it to get the predicted normal $\normal' = \normal_\text{raw}' / \norm{\normal_\text{raw}'}_2$.
However, in practice, we find that this will occasionally lead to a normal flipping issue when $\normal_\text{raw}'$ is numerically small and $\normal'$ will flip its direction with only a very small deviation of $\normal_\text{raw}'$ during training.
\Cref{fig:supple_normalflip} visualizes this issue.
The flip of the predicted normal will further lead to suboptimal normals derived from the density gradient due to the normal prediction loss $\mathcal{L}_\text{norm}$.
To alleviate this normal flip issue, we correct the direction of the predicted normal by forcing the angle between the final normal $\normal'$ and the view direction to be smaller than 90°:
\begin{equation}
    \normal' = -\sign(\dir \cdot \normal_\text{raw}') \frac{\normal_\text{raw}'}{\norm{\normal_\text{raw}'}_2} \text{,}
\end{equation}
where $\dir$ is the ray direction.
We can see from \cref{fig:supple_normalflip} that this normal correction operation helps us prevent the normal flip, and yields a better normal prediction.

\begin{figure*}
  \centering
  
  \rotatebox{90}{\footnotesize \phantom{g}}
  \begin{subfigure}{0.15\linewidth}
  	\caption*{Blur kernel size = 1}
  \end{subfigure}
  \begin{subfigure}{0.15\linewidth}
  	\caption*{9}
  \end{subfigure}
  \begin{subfigure}{0.15\linewidth}
  	\caption*{17}
  \end{subfigure}
  \begin{subfigure}{0.15\linewidth}
  	\caption*{33}
  \end{subfigure}
  \begin{subfigure}{0.15\linewidth}
  	\caption*{65}
  \end{subfigure}
  \begin{subfigure}{0.15\linewidth}
  	\caption*{129}
  \end{subfigure}
  
  \rotatebox{90}{\hspace{1cm} \footnotesize Blurred Input}
  \begin{subfigure}{0.15\linewidth}
    \centering
    \includegraphics[width=\linewidth]{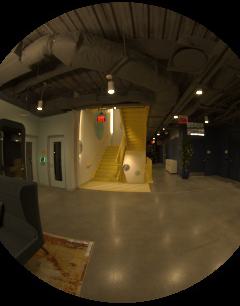}
  \end{subfigure}
  \begin{subfigure}{0.15\linewidth}
    \centering
    \includegraphics[width=\linewidth]{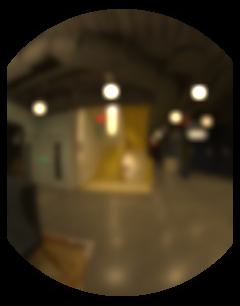}
  \end{subfigure}
  \begin{subfigure}{0.15\linewidth}
    \centering
    \includegraphics[width=\linewidth]{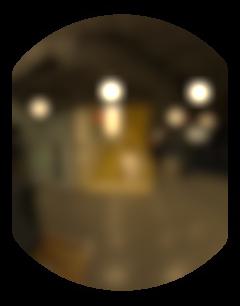}
  \end{subfigure}
  \begin{subfigure}{0.15\linewidth}
    \centering
    \includegraphics[width=\linewidth]{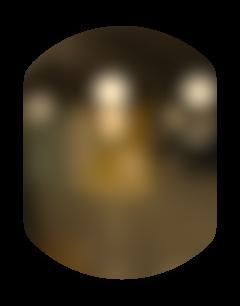}
  \end{subfigure}
  \begin{subfigure}{0.15\linewidth}
    \centering
    \includegraphics[width=\linewidth]{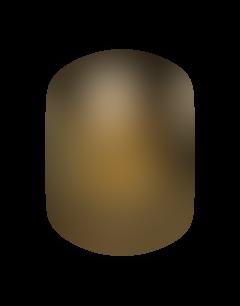}
  \end{subfigure}
  \begin{subfigure}{0.15\linewidth}
    \centering
    \includegraphics[width=\linewidth]{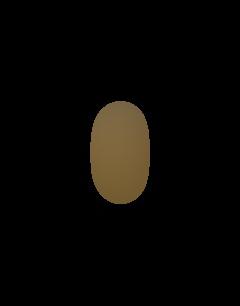}
  \end{subfigure}
  
    \rotatebox{90}{\hspace{1cm} \footnotesize Prediction\phantom{g}}
  \begin{subfigure}{0.15\linewidth}
    \centering
    \includegraphics[width=\linewidth]{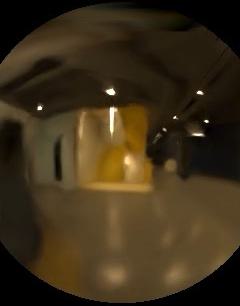}
  \end{subfigure}
  \begin{subfigure}{0.15\linewidth}
    \centering
    \includegraphics[width=\linewidth]{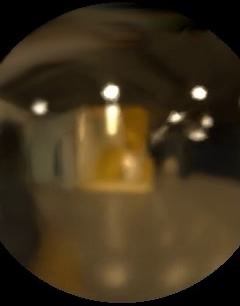}
  \end{subfigure}
  \begin{subfigure}{0.15\linewidth}
    \centering
    \includegraphics[width=\linewidth]{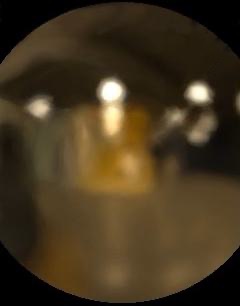}
  \end{subfigure}
  \begin{subfigure}{0.15\linewidth}
    \centering
    \includegraphics[width=\linewidth]{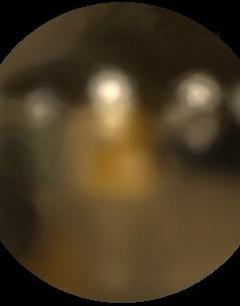}
  \end{subfigure}
  \begin{subfigure}{0.15\linewidth}
    \centering
    \includegraphics[width=\linewidth]{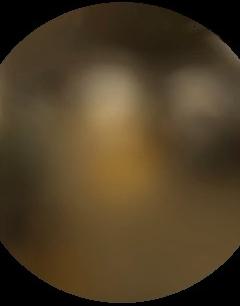}
  \end{subfigure}
  \begin{subfigure}{0.15\linewidth}
    \centering
    \includegraphics[width=\linewidth]{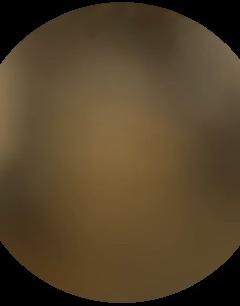}
  \end{subfigure}
  
  \vspace{-2mm}
    \caption{%
    	One example of the Gaussian initialization input (top) and predictions (bottom) for different scales.
     }
    \label{fig:supple_gauinit}
    \vspace{-4mm}
\end{figure*}

\subsection{Training and Rendering Configuration}
We use the Adam optimizer \cite{KingmB2015} to train our NeRF model using
the default parameter configurations in PyTorch \cite{pytorch} for the optimizer, except that we set the learning rate to 0.005.
When rendering a pixel, we first shoot rays from the camera origin to the pixel locations, and then sample points along each ray.
Similar to Nerfstudio \cite{TanciWNLYWKASAMKK2023}, we use two levels of proposal sampling, guided by two density fields.
Specifically, in the first round, we sample 256 points using exponential distance.
We set the far distance to a constant value of 800\,meters, and we determine the near distance for each scene using the minimum distance between all the structure-from-motion points and the viewing cameras.
Then, in each iteration of the proposal sampling process, we feed the samples into the proposal network sampler and generate new samples based on the integration weights of the input samples.
We sample 96 samples in the first iteration of the proposal process, followed by 48 samples in the second.
The model structures of the proposal networks follow those in the ``nerfactor'' model in Nerfstudio \cite{TanciWNLYWKASAMKK2023}.

\subsection{Gaussian Parameter Optimization}
\begin{figure}
	\centering
	\includegraphics[width=\linewidth]{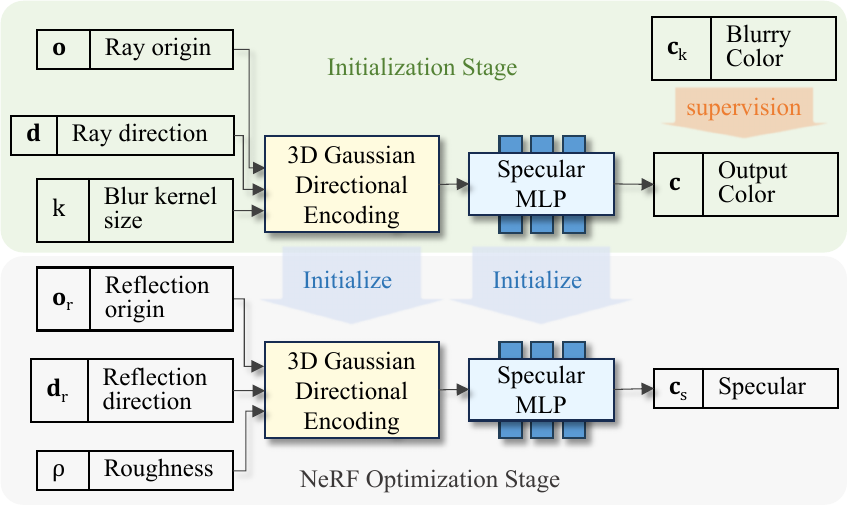}
	\caption{Illustration of the initialization stage.
		We optimize the Gaussian parameters and the Specular MLP using the Gaussian-blurred input images, and then use them
		as initialization for the NeRF optimization stage.}
	\label{fig:supple_gauinit_overview}
\end{figure}

To obtain optimal parameters for the Gaussian directional encoding, we use an initialization stage to seed the Gaussian parameters and specular MLP weights.
The process is illustrated in \cref{fig:supple_gauinit_overview}.
We optimize a preconvolved incident light field composed of our 3D Gaussian directional encoding and the Specular MLP.
We first apply a range of Gaussian blurs to all input images using a series of standard deviations,
generating pyramids of blurry input images.
In our experiments, we first scale the input images to have 360 pixels along the longest axis.
Then, we apply OpenCV's \texttt{GaussianBlur} \cite{opencv} with kernel sizes (1, 3, 5, 9, 17, 33, 65, 129).
Regions that involve the image border during blurring are marked as invalid, resulting in a wider invalid border with larger kernel size.
\Cref{fig:supple_gauinit} showcases one example view with some of the blur kernels used.

\begin{figure}
  \centering
  
  \begin{subfigure}{0.32\linewidth}
  	\caption*{SfM mesh}
  \end{subfigure}
  \begin{subfigure}{0.32\linewidth}
  	\caption*{Gaussian blobs}
  \end{subfigure}
  \begin{subfigure}{0.32\linewidth}
  	\caption*{Combined view}
  \end{subfigure}
  
  \begin{subfigure}{0.325\linewidth}
    \centering
    \includegraphics[width=\linewidth]{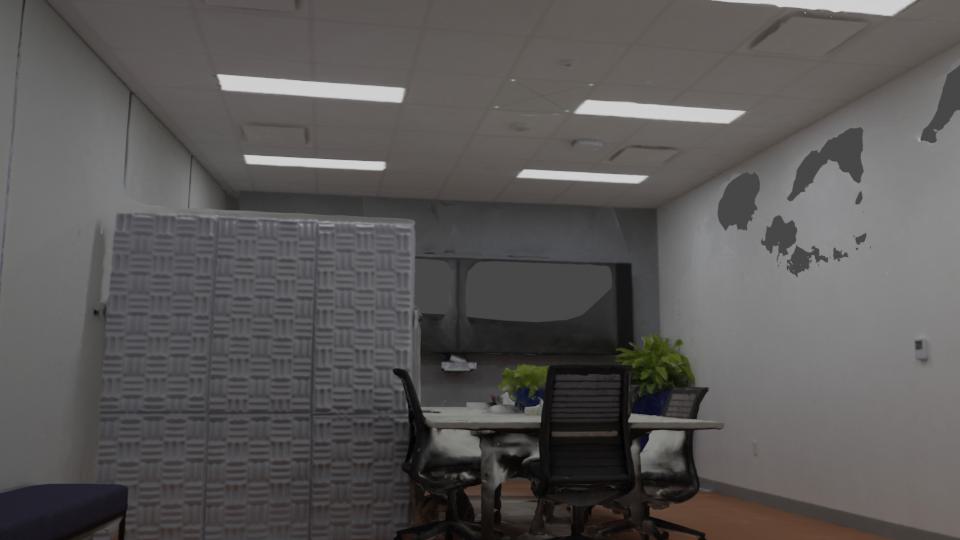}
  \end{subfigure}
  \begin{subfigure}{0.325\linewidth}
    \centering
    \includegraphics[width=\linewidth]{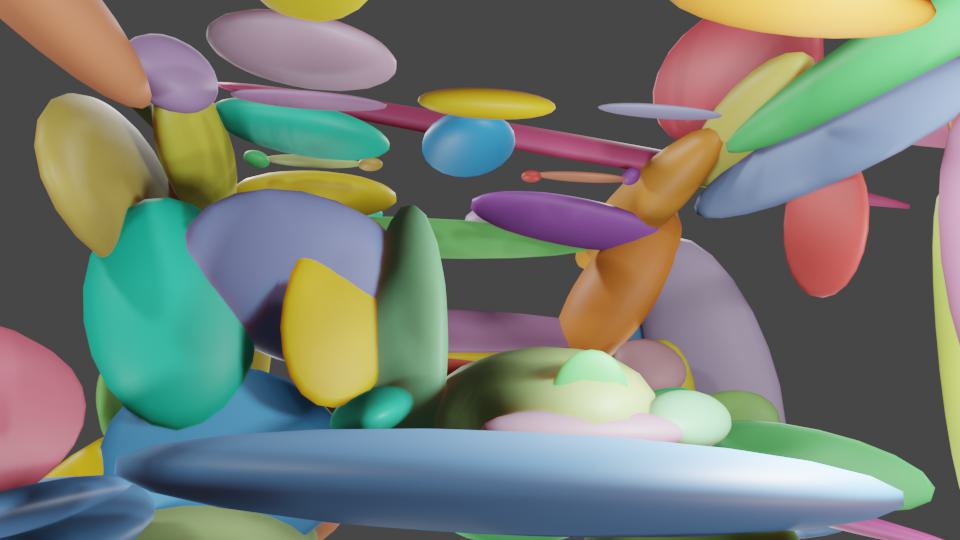}
  \end{subfigure}
  \begin{subfigure}{0.325\linewidth}
    \centering
    \includegraphics[width=\linewidth]{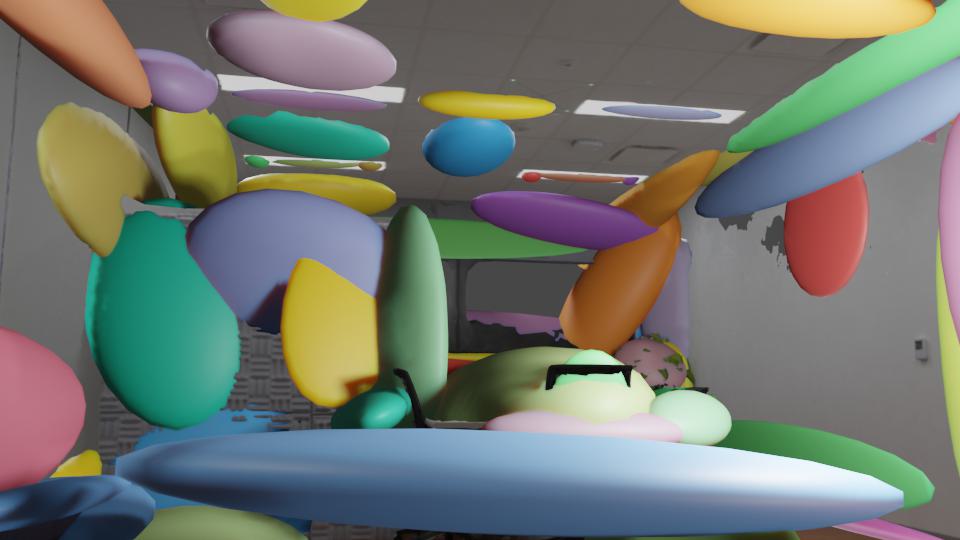}
  \end{subfigure}

  \begin{subfigure}{0.325\linewidth}
    \centering
    \includegraphics[width=\linewidth]{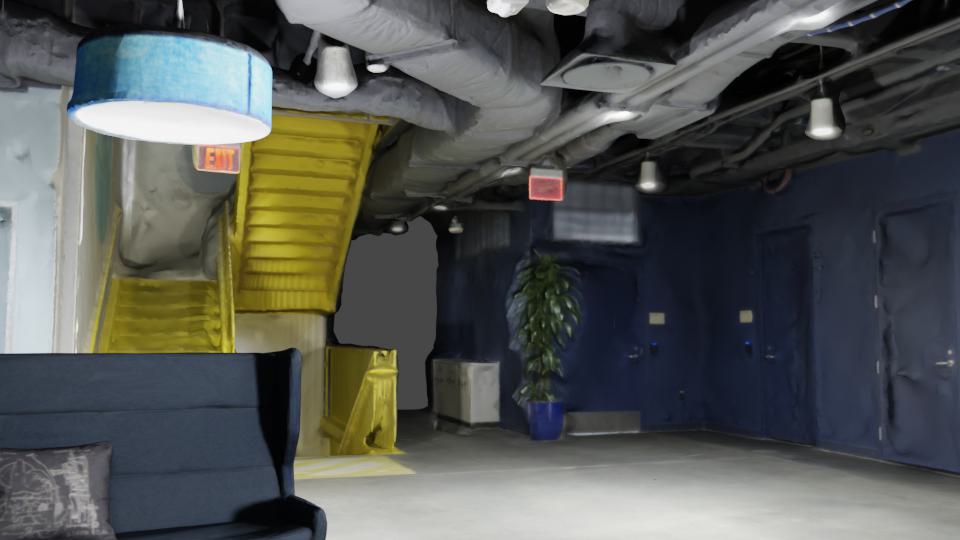}
  \end{subfigure}
  \begin{subfigure}{0.325\linewidth}
    \centering
    \includegraphics[width=\linewidth]{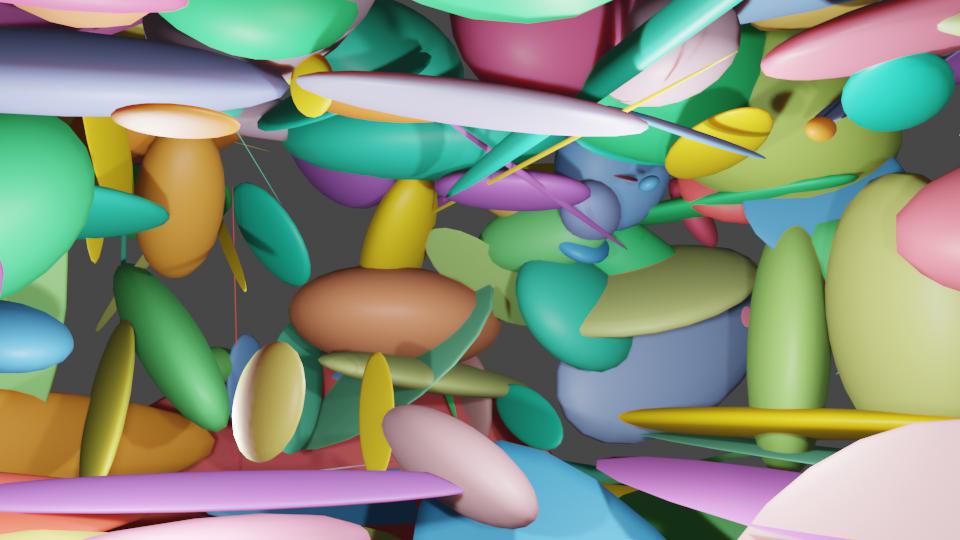}
  \end{subfigure}
  \begin{subfigure}{0.325\linewidth}
    \centering
    \includegraphics[width=\linewidth]{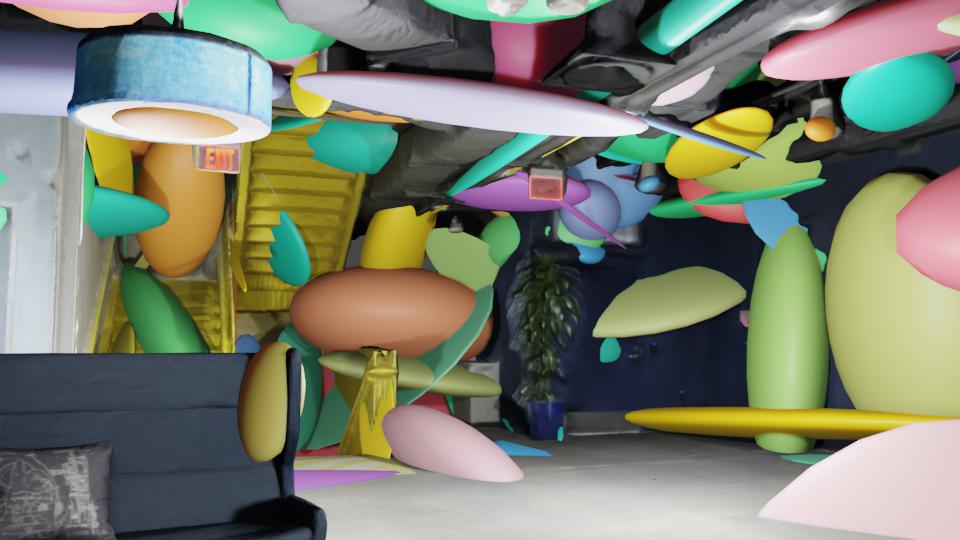}
  \end{subfigure}
  
  \vspace{-2mm}
    \caption{%
    Visualization of the learned Gaussian blobs for two scenes.
    We assign a random color for each Gaussian blob for better visibility.
     }
    \label{fig:supple_gauvis}
    \vspace{-4mm}
\end{figure}

All valid blurred pixels compose our ray dataset for the initialization stage.
In each iteration, we sample 25,600 pixels from the ray dataset, and generate the corresponding ray origin $\origin$, direction $\dir$, and the blur kernel size $k$.
We train the Gaussian parameters and Specular MLP using Adam \cite{KingmB2015} with a learning rate of 0.001, and leave other parameters as default.
We supervise the output color using the corresponding blurry color in the Gaussian pyramid using an L1 loss.
We find this small network converges quickly, thus we only train for 8,000 iterations, which takes around half an hour to finish on one NVIDIA A100 GPU.
We visualize the fitted preconvolved incident light field in \cref{fig:supple_gauinit}.
The reconstructed preconvolved light field well-represents the input with multiple blur levels.
We also visualize the fitted Gaussian blobs of two scenes in \cref{fig:supple_gauvis}.
We can see that some Gaussian blobs are aligned with the underlying objects (e.g. the lamp on the ceiling).

\subsection{Losses}

Recall that in our experiments, the final loss is a combination of several terms:
\begin{equation}
    \mathcal{L} = \mathcal{L}_\text{c} + \mathcal{L}_\text{prop} + \lambda_\text{dist} \mathcal{L}_\text{dist} + \lambda_\text{mono} \mathcal{L}_\text{mono} + \lambda_\text{norm} \mathcal{L}_\text{norm} \text{.}
\end{equation}
In this section, we follow the notation that $i = 1, ..., N$ is the sample point index along a ray.
We omit the ray index as each loss term has the same form for all rays.
The loss term is averaged over all rays within a training batch.

\paragraph{Reconstruction Loss}
The reconstruction loss $\mathcal{L}_\text{c}$ is the L1 norm between the predicted color $\mathbf{c}$ and the ground-truth color $\mathbf{c}_\text{gt}$:
\begin{equation}
    \mathcal{L}_\text{c} = \norm{\mathbf{c} - \mathbf{c}_\text{gt}}_1 \text{.}
\end{equation}
For the Eyeful Tower dataset, we compute the reconstruction loss in the Perceptual Quantizer (PQ) color space, as in VR-NeRF \cite{XuALGBKRPKBLZR2023}.
For other public datasets, we use the standard sRGB color space.

\paragraph{Proposal Loss and Distortion Loss}
The proposal loss $\mathcal{L}_\text{prop}$ supervises the density field of the proposal network to be consistent with that of the main NeRF.
The distortion loss $\mathcal{L}_\text{dist}$ is a regularization term for the density field of the main NeRF.
It consolidates the volumetric blending weights into as small a region as possible.
Please refer to \citet{mipnerf360} for the detailed definitions and explanations of both losses.

\paragraph{Normal Prediction Loss}
We encourage the predicted normals from the normal MLP to be consistent with the underlying geometry of NeRF.
For this, we use a normal prediction loss $\mathcal{L}_\text{norm}$ that supervises the normal $\normal'_i$ predicted for every sample point using the normal MLP and NeRF density gradient $\grad_i$:
\begin{equation}
    \mathcal{L}_\text{norm} = \frac{1}{N} \sum_{i} \norm{\normal'_i - \frac{-\grad_i}{\norm{\grad_i}}} \text{.}
\end{equation}
To compute the gradient of the density $\tau'$ with respect to the input world coordinate $\pos = (x, y, z)$, we could use the analytical gradient, which is natively supported by PyTorch \cite{pytorch}.
However, we model the density field using a hash-grid-based representation, which is prone to noisy gradients and has poor optimization performance \cite{neuralangelo}.
Therefore, we adopt a modified version of the numerical gradient from Neuralangelo \cite{neuralangelo}.
To compute the gradient along the $x$-axis, we use
\begin{equation}
	\nabla_x\tau' = \frac{\tau'(\pos + \eps_x) - \tau'(\pos - \eps_x)}{2\epsilon} \text{,}
	\label{eq:grad}
\end{equation}
where $\eps_x = (\epsilon, 0, 0)$.
The equations for computing the gradient along the $y$- and $z$-axes can be derived analogously.
Overall, $\nabla_\pos\tau$ involves sampling six additional points to query the density value.
Instead of predefining the schedule of the $\epsilon$ value during training, we compute a per-sample $\epsilon$ value that is consistent with the cone tracing radius at the sample location: $\epsilon = t \!\cdot\! r$.
Here, $t$ is the ray-marching distance of the sample point, and $r$ is the base radius of a pixel at unit distance along the ray.

\paragraph{Additional Losses}
For the Eyeful Tower dataset, we also deploy a depth supervision loss and an ``empty around camera'' loss, following VR-NeRF \cite{XuALGBKRPKBLZR2023}. 
For the depth loss, we supervise the NeRF depth with the depth from structure-from-motion mesh using L1 distance in the first 500 iterations.
For the ``empty around camera'' loss, we randomly sample 128 points in the unit sphere around training cameras, and regularize the density value to be zero.
This reduces the near-plane ambiguity as shown in FreeNeRF \cite{FreeNeRF}.
We set the weights of the depth loss and ``empty around camera'' loss to 0.1 and 10, respectively.

\section{Physical Interpretation of 3D Gaussians}
\begin{figure}[h]
  \centering
  
  \begin{subfigure}{0.32\linewidth}\hspace{4pt}
  	\caption*{Final}
  \end{subfigure}
  \hspace{-3pt}
  \begin{subfigure}{0.32\linewidth}
  	\caption*{Diffuse}
  \end{subfigure}
  \hspace{-3pt}
  \begin{subfigure}{0.32\linewidth}
  	\caption*{Specular}
  \end{subfigure}
  
  \begin{subfigure}{0.32\linewidth}
    \centering
    \includegraphics[width=\linewidth]{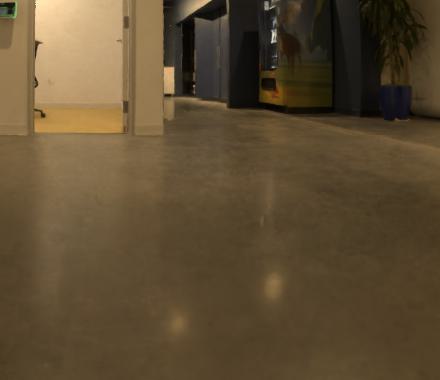}
  \end{subfigure}
  \hspace{-3pt}
  \begin{subfigure}{0.32\linewidth}
    \includegraphics[width=\linewidth]{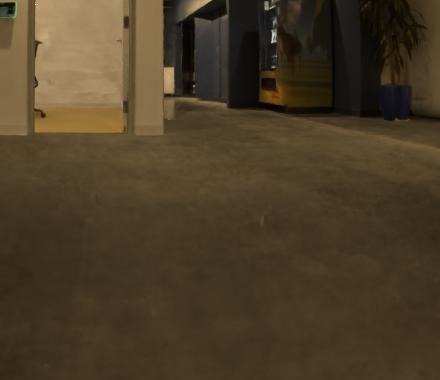}
  \end{subfigure}
  \hspace{-3pt}
  \begin{subfigure}{0.32\linewidth}
    \includegraphics[width=\linewidth]{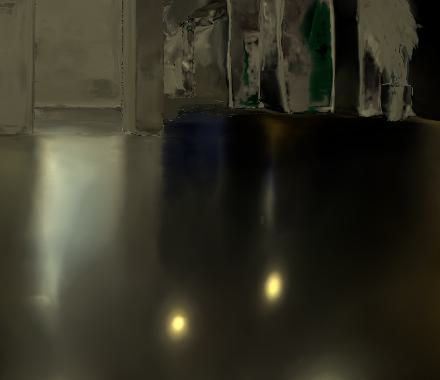}
  \end{subfigure}

  \vspace{-2mm}
    \caption{%
    	Our 3D Gaussians can model global illumination effects.
        This is evident on the floor, where the indirect light from the room is captured and represented through the specular component.
     }
    \label{fig:indirlight}
\end{figure}

Though a 3D Gaussian blob may appear similar to a point light source, we would like to emphasize that the 3D Gaussians do not represent explicit light sources, nor are they specifically designed for modeling direct light alone.
Instead, they serve as basis functions for representing the scene’s full 5D specular radiance field, including global illumination effects.
One example can be seen in \cref{fig:indirlight}.
This is analogous to how spherical Gaussians (SGs) represent a 2D environment map.

\begin{figure}[b]
    \vspace{-2mm}%
	\includegraphics[width=\linewidth,trim=14 7 16 4,clip]{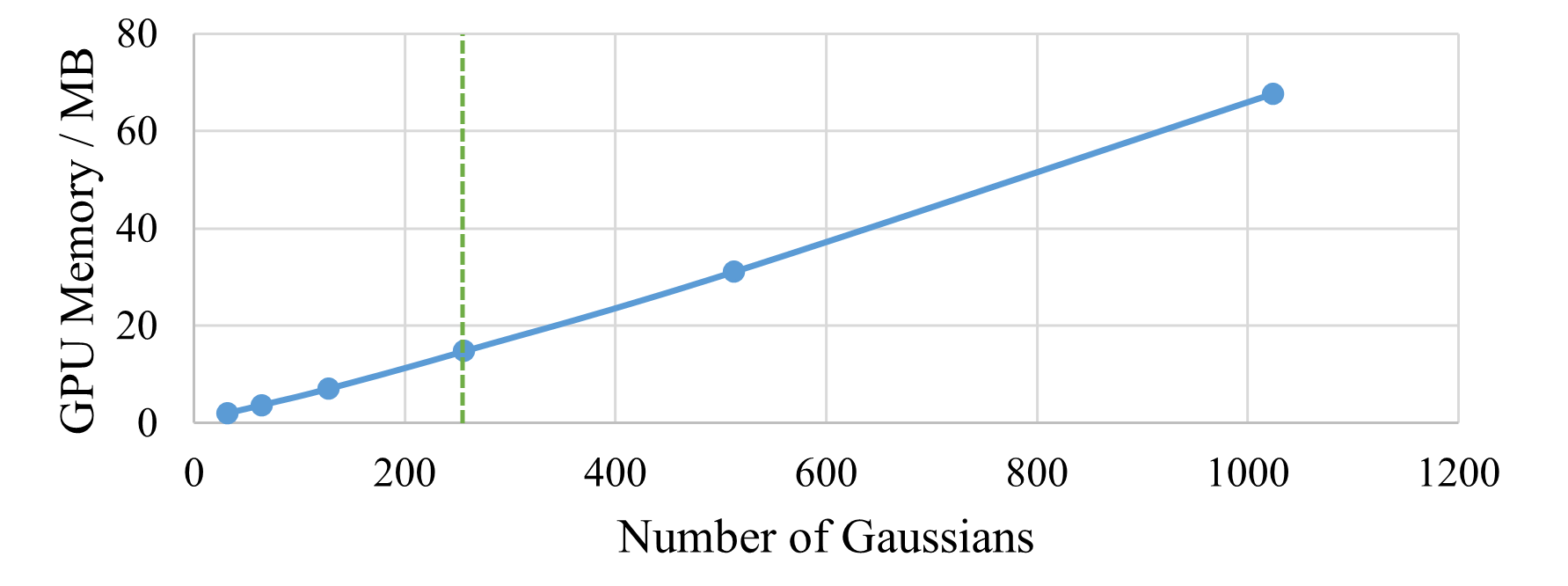}%
    \vspace{-2mm}%
	\caption{The GPU memory consumption of the Gaussian directional encoding and the Specular MLP with various number of Gaussians.
		We test GPU memory with a batch size of 12,800 rays.
		The green dashed line is the configuration used in our experiments.}
	\label{fig:supple_numgau_gpu}
\end{figure}

\section{Additional Experiments}

\subsection{Number of Gaussians}
We test the GPU memory usage of our Gaussian directional encoding and the specular MLP under a series of Gaussians, and visualize the results in \cref{fig:supple_numgau_gpu}.
We can see that our reflection model adds very little GPU memory overhead compared to the approximately 8\,GB of overall memory used for training the whole pipeline.

\begin{table}
\caption{\label{tab:shinyblender}%
    Quantitative comparisons on the Shiny Blender dataset \cite{VerbiHMZBS2022}. Our approach demonstrates comparable performance to Ref-NeRF since the dataset assumes perfect 2D lighting conditions.
}
\vspace{-2mm}
\begin{tabularx}{\linewidth}{p{2.2cm}|YYY}
\toprule
Methods 
 &
  \multicolumn{1}{c}{PSNR $\uparrow$} &
  \multicolumn{1}{c}{SSIM $\uparrow$} &
  \multicolumn{1}{c}{LPIPS $\downarrow$} \\
\midrule
Ours & 34.65 & 0.9615 & 0.0515 \\
Ref-NeRF~\cite{VerbiHMZBS2022} &  34.69 &     0.9619 &  0.0508  \\
\bottomrule
\end{tabularx}
\end{table}
\begin{figure}
  \centering
  
  \rotatebox{90}{\footnotesize \phantom{g}}
  \begin{subfigure}{0.31\linewidth}\hspace{4pt}
  	\caption*{Ground Truth}
  \end{subfigure}
  \hspace{-3pt}
  \begin{subfigure}{0.31\linewidth}
  	\caption*{Ours}
  \end{subfigure}
  \hspace{-3pt}
  \begin{subfigure}{0.31\linewidth}
  	\caption*{Ref-NeRF \cite{VerbiHMZBS2022}}
  \end{subfigure}
  
  \rotatebox{90}{\hspace{10mm} \footnotesize \textit{Ball}}
  \begin{subfigure}{0.31\linewidth}
    \centering
    \includegraphics[width=\linewidth]{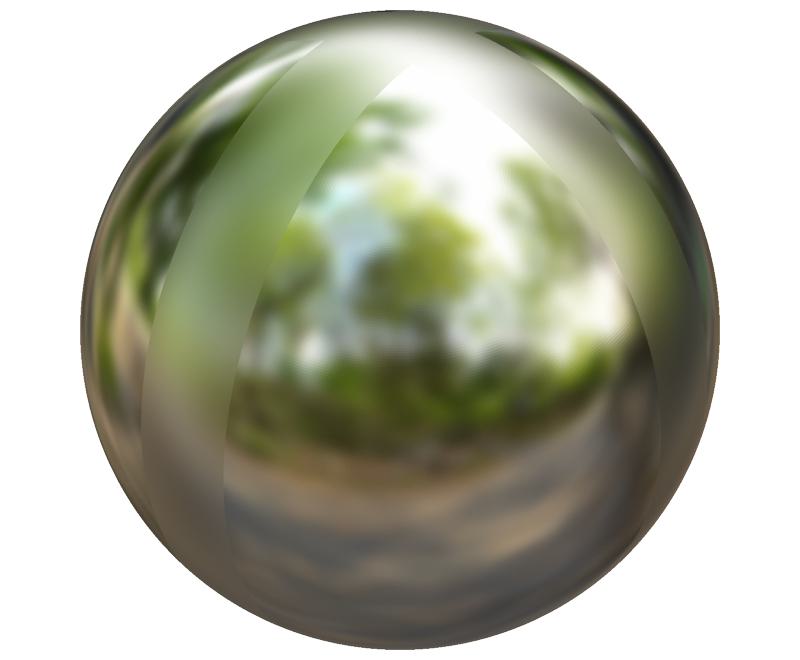}
  \end{subfigure}
  \hspace{-3pt}
  \begin{subfigure}{0.31\linewidth}
    \includegraphics[width=\linewidth]{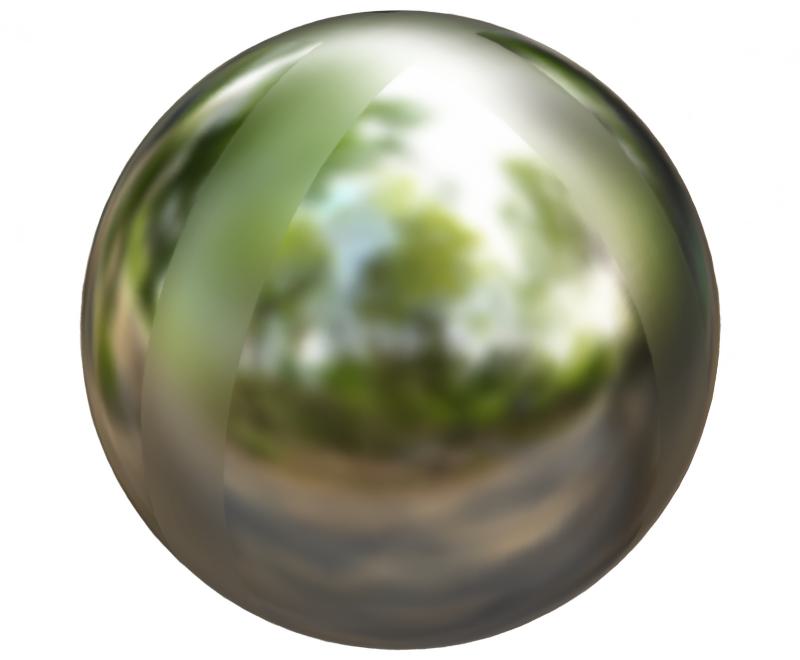}
  \end{subfigure}
  \hspace{-3pt}
  \begin{subfigure}{0.31\linewidth}
    \includegraphics[width=\linewidth]{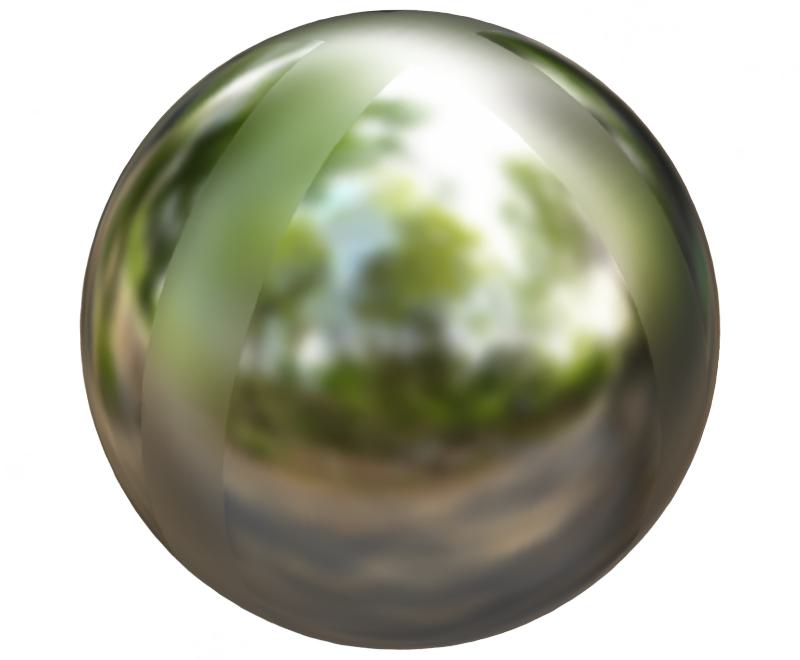}
  \end{subfigure}

  \rotatebox{90}{\hspace{11pt} \footnotesize \textit{Coffee} \phantom{g}}
  \begin{subfigure}{0.31\linewidth}
    \includegraphics[width=\linewidth]{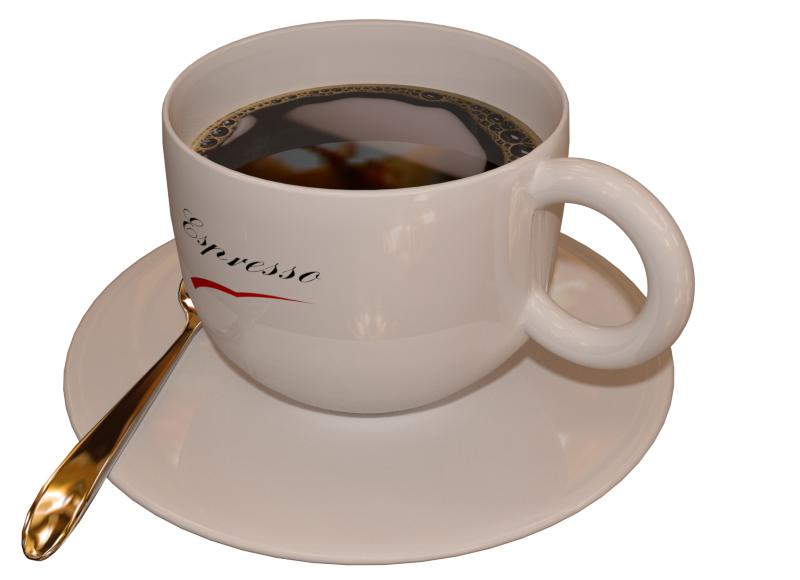}
  \end{subfigure}
  \hspace{-3pt}
  \begin{subfigure}{0.31\linewidth}
    \includegraphics[width=\linewidth]{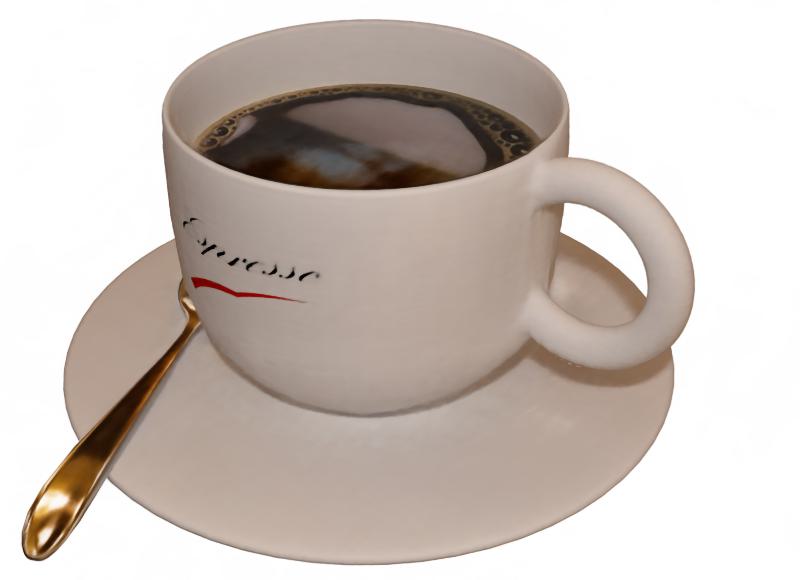}
  \end{subfigure}
  \hspace{-3pt}
  \begin{subfigure}{0.31\linewidth}
    \includegraphics[width=\linewidth]{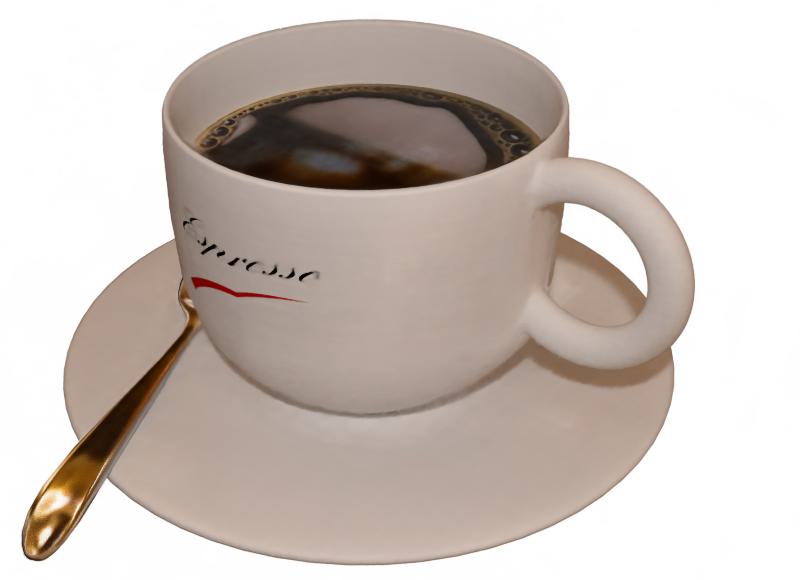}
  \end{subfigure}

    \caption{%
    	Qualitative comparisons of two example test views from Shiny Blender dataset~\cite{VerbiHMZBS2022}. 
     }
    \label{fig:shinyblender}
\end{figure}

\subsection{Shiny Blender Dataset}
We evaluate our method and the Ref-NeRF baseline on the Shiny Blender dataset \cite{VerbiHMZBS2022}.
We utilize a re-implementation of Ref-NeRF in NeRF-Factory \cite{nerfactory}.
To ensure a fair comparison, we adopt the same MLP backbone as used in NeRF-Factory.
We train both methods for 80,000 iterations for each scene.
The visual results are shown in \cref{fig:shinyblender} and the quantitative results are depicted in \cref{tab:shinyblender}.
Our method achieves comparable performance to Ref-NeRF, which is expected because all scenes in the dataset are lit by perfect 2D (far-field) environment light.
Our method outperforms Ref-NeRF under near-field lighting scenes as shown in the paper.

\subsection{Synthetic Dataset}
We compare our method with several baselines on the FIPT synthetic dataset \cite{RT_FIPT}.
In addition to the baselines described in the main paper, we also compare with FIPT \cite{RT_FIPT}, a state-of-the-art path-tracing-based inverse rendering approach.
We report the average PSNR, SSIM and LPIPS metrics for novel-view synthesis.
Since we have the ground-truth mesh for the synthetic dataset, we also report the mean angular error (MAE) used in Ref-NeRF \cite{VerbiHMZBS2022} for evaluating the estimated normal accuracy.
The results in \cref{tab:supple_synthetic_comp} show that our method achieves the best novel-view synthesis quality and geometry accuracy.
Interestingly, despite the use of ground-truth geometry for the physically based inverse rendering approach, the novel-view synthesis is worse than any NeRF-based baseline by a large margin.
This suggests that introducing a fully physically based rendering model may be a disadvantage when it comes to novel-view synthesis quality, at least compared to NeRF-like approaches that are tailored specifically for the view synthesis task.

\subsection{Additional Results}
\begin{figure*}
	\rotatebox[origin=c]{90}{\vspace{-3mm} \footnotesize \textit{Shiny1}}
	\begin{minipage}{0.183\textwidth}
		\begin{subfigure}{\textwidth}
			\caption*{GT Test Image}
			\includegraphics[width=\linewidth]{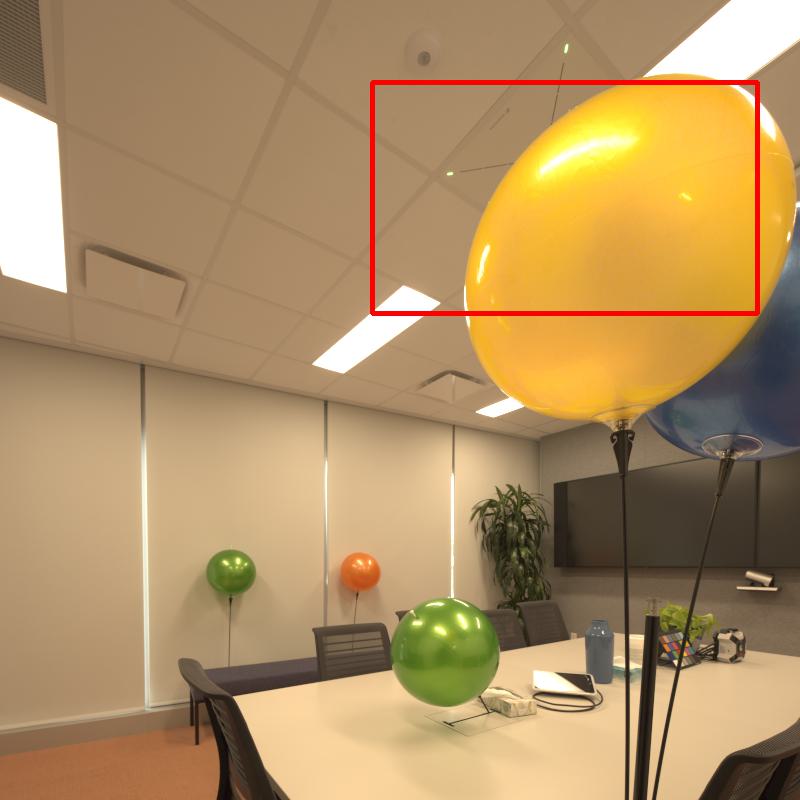}
		\end{subfigure}
	\end{minipage}
	\begin{minipage}{0.793\textwidth}
		\begin{subfigure}{0.19\linewidth}
			\caption*{GT \& SfM Normal}
			\includegraphics[width=\linewidth]{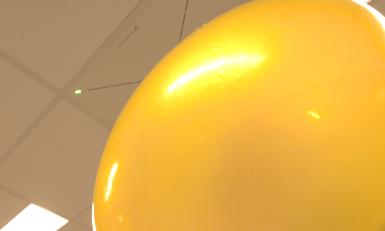}
		\end{subfigure}
		\begin{subfigure}{0.19\linewidth}
			\caption*{Ours}
			\includegraphics[width=\linewidth]{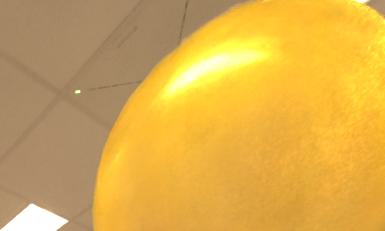}
		\end{subfigure}
		\begin{subfigure}{0.19\linewidth}
			\caption*{Ref-NeRF \cite{VerbiHMZBS2022}}
			\includegraphics[width=\linewidth]{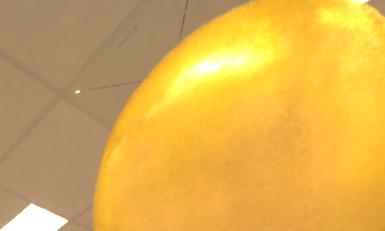}
		\end{subfigure}
		\begin{subfigure}{0.19\linewidth}
			\caption*{MS-NeRF \cite{YinQCR2023}}
			\includegraphics[width=\linewidth]{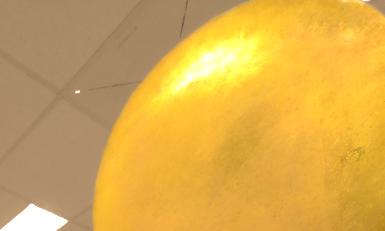}
		\end{subfigure}
		\begin{subfigure}{0.19\linewidth}
			\caption*{NeRF \cite{MildeSTBRN2020}}
			\includegraphics[width=\linewidth]{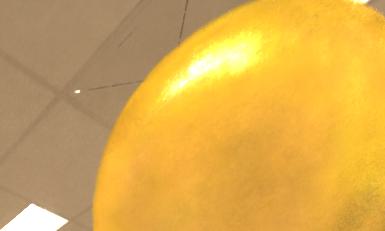}
		\end{subfigure} \\
		\begin{subfigure}{0.19\linewidth}
			\includegraphics[width=\linewidth]{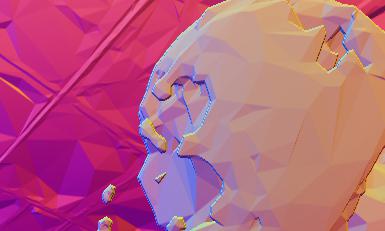}
		\end{subfigure}
		\begin{subfigure}{0.19\linewidth}
			\includegraphics[width=\linewidth]{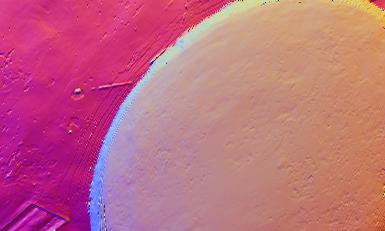}
		\end{subfigure}
		\begin{subfigure}{0.19\linewidth}
			\includegraphics[width=\linewidth]{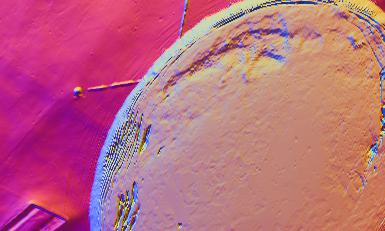}
		\end{subfigure}
		\begin{subfigure}{0.19\linewidth}
			\includegraphics[width=\linewidth]{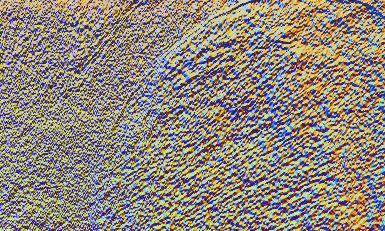}
		\end{subfigure}
		\begin{subfigure}{0.19\linewidth}
			\includegraphics[width=\linewidth]{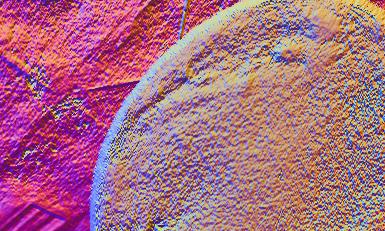}
		\end{subfigure}
	\end{minipage}
	
	\vspace{1mm}
	
	\rotatebox[origin=c]{90}{\footnotesize Eyeful Tower \textit{Apartment}}
	\begin{minipage}{0.183\textwidth}
		\begin{subfigure}{\textwidth}
			\includegraphics[width=\linewidth]{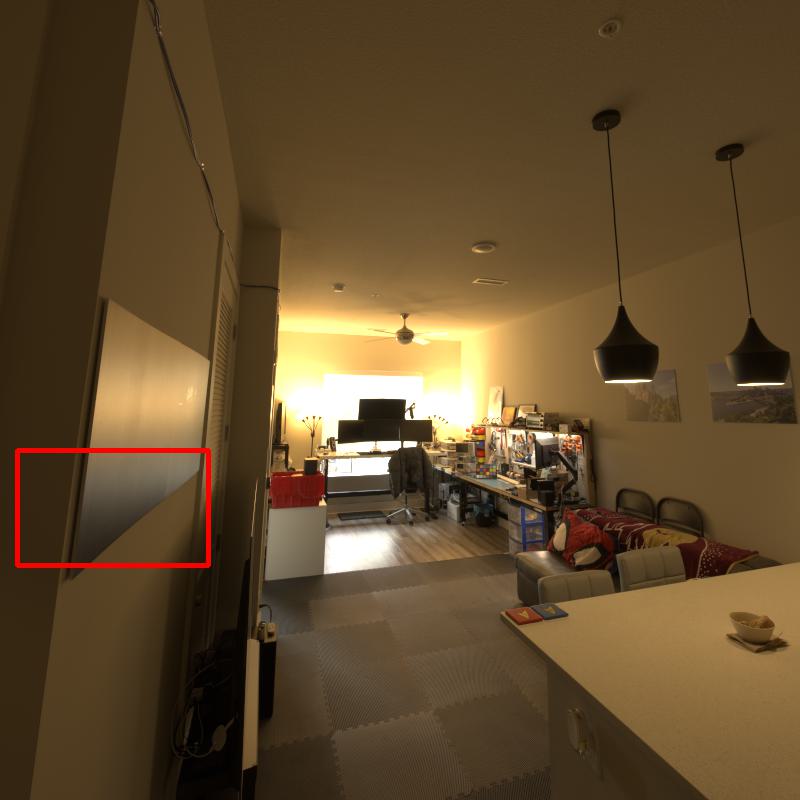}
		\end{subfigure}
	\end{minipage}
	\begin{minipage}{0.793\textwidth}
		\begin{subfigure}{0.19\linewidth}
			\includegraphics[width=\linewidth]{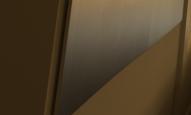}
		\end{subfigure}
		\begin{subfigure}{0.19\linewidth}
			\includegraphics[width=\linewidth]{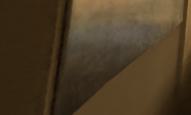}
		\end{subfigure}
		\begin{subfigure}{0.19\linewidth}
			\includegraphics[width=\linewidth]{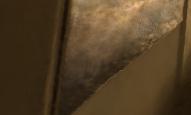}
		\end{subfigure}
		\begin{subfigure}{0.19\linewidth}
			\includegraphics[width=\linewidth]{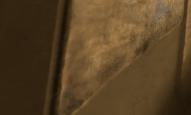}
		\end{subfigure}
		\begin{subfigure}{0.19\linewidth}
			\includegraphics[width=\linewidth]{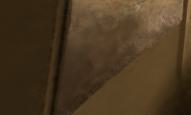}
		\end{subfigure} \\
		\begin{subfigure}{0.19\linewidth}
			\includegraphics[width=\linewidth]{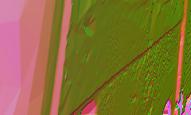}
		\end{subfigure}
		\begin{subfigure}{0.19\linewidth}
			\includegraphics[width=\linewidth]{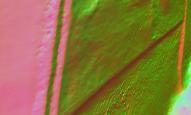}
		\end{subfigure}
		\begin{subfigure}{0.19\linewidth}
			\includegraphics[width=\linewidth]{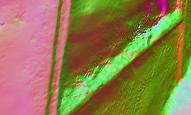}
		\end{subfigure}
		\begin{subfigure}{0.19\linewidth}
			\includegraphics[width=\linewidth]{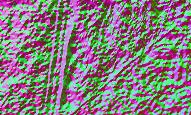}
		\end{subfigure}
		\begin{subfigure}{0.19\linewidth}
			\includegraphics[width=\linewidth]{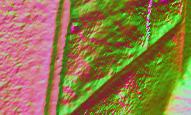}
		\end{subfigure}
	\end{minipage}
	
	\vspace{1mm}
	
	\rotatebox[origin=c]{90}{\footnotesize Eyeful Tower \textit{Office2}}
	\begin{minipage}{0.183\textwidth}
		\begin{subfigure}{\textwidth}
			\includegraphics[width=\linewidth]{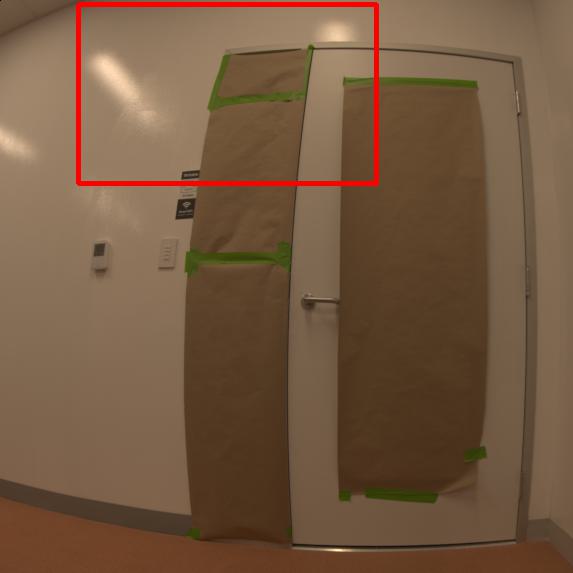}
		\end{subfigure}
	\end{minipage}
	\begin{minipage}{0.793\textwidth}
		\begin{subfigure}{0.19\linewidth}
			\includegraphics[width=\linewidth]{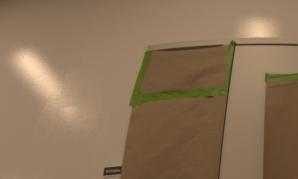}
		\end{subfigure}
		\begin{subfigure}{0.19\linewidth}
			\includegraphics[width=\linewidth]{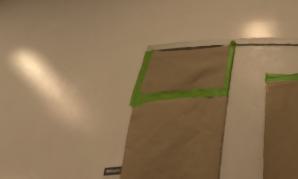}
		\end{subfigure}
		\begin{subfigure}{0.19\linewidth}
			\includegraphics[width=\linewidth]{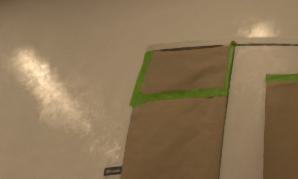}
		\end{subfigure}
		\begin{subfigure}{0.19\linewidth}
			\includegraphics[width=\linewidth]{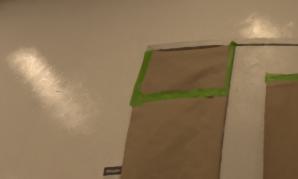}
		\end{subfigure}
		\begin{subfigure}{0.19\linewidth}
			\includegraphics[width=\linewidth]{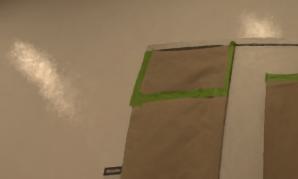}
		\end{subfigure} \\
		\begin{subfigure}{0.19\linewidth}
			\includegraphics[width=\linewidth]{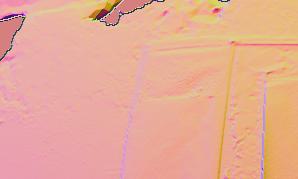}
		\end{subfigure}
		\begin{subfigure}{0.19\linewidth}
			\includegraphics[width=\linewidth]{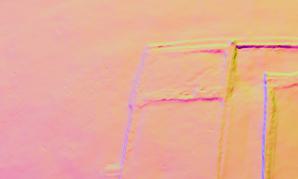}
		\end{subfigure}
		\begin{subfigure}{0.19\linewidth}
			\includegraphics[width=\linewidth]{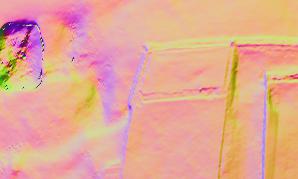}
		\end{subfigure}
		\begin{subfigure}{0.19\linewidth}
			\includegraphics[width=\linewidth]{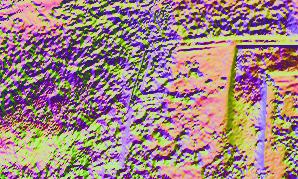}
		\end{subfigure}
		\begin{subfigure}{0.19\linewidth}
			\includegraphics[width=\linewidth]{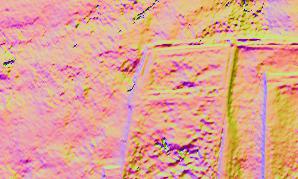}
		\end{subfigure}
	\end{minipage}
	
	\vspace{1mm}

    \rotatebox[origin=c]{90}{\footnotesize Eyeful Tower \textit{Office2}}
	\begin{minipage}{0.183\textwidth}
		\begin{subfigure}{\textwidth}
			\includegraphics[width=\linewidth]{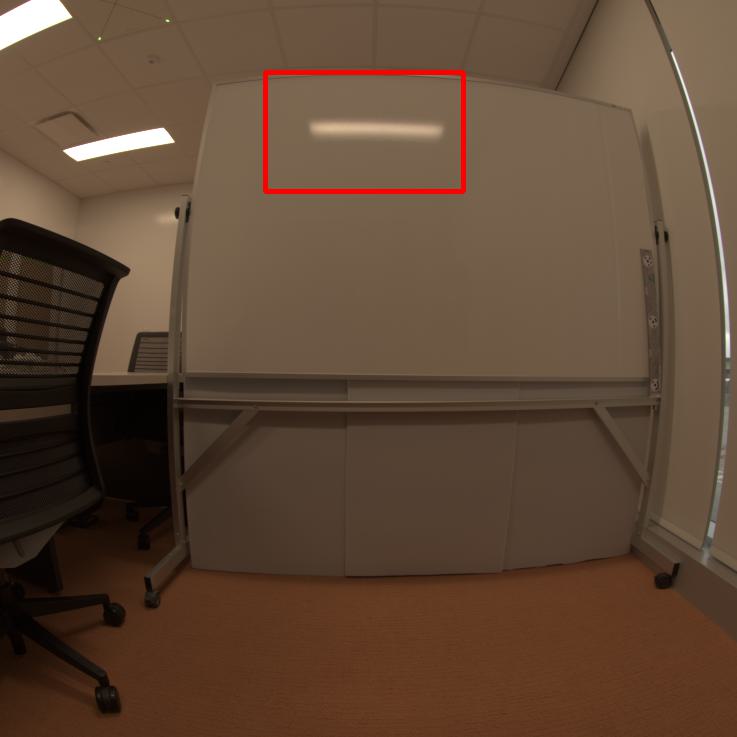}
		\end{subfigure}
	\end{minipage}
	\begin{minipage}{0.793\textwidth}
		\begin{subfigure}{0.19\linewidth}
			\includegraphics[width=\linewidth]{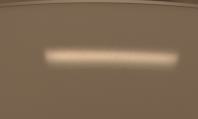}
		\end{subfigure}
		\begin{subfigure}{0.19\linewidth}
			\includegraphics[width=\linewidth]{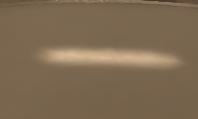}
		\end{subfigure}
		\begin{subfigure}{0.19\linewidth}
			\includegraphics[width=\linewidth]{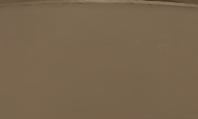}
		\end{subfigure}
		\begin{subfigure}{0.19\linewidth}
			\includegraphics[width=\linewidth]{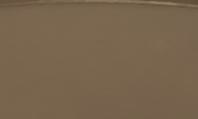}
		\end{subfigure}
		\begin{subfigure}{0.19\linewidth}
			\includegraphics[width=\linewidth]{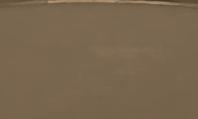}
		\end{subfigure} \\
		\begin{subfigure}{0.19\linewidth}
			\includegraphics[width=\linewidth]{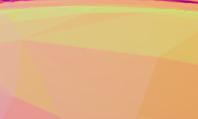}
		\end{subfigure}
		\begin{subfigure}{0.19\linewidth}
			\includegraphics[width=\linewidth]{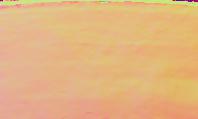}
		\end{subfigure}
		\begin{subfigure}{0.19\linewidth}
			\includegraphics[width=\linewidth]{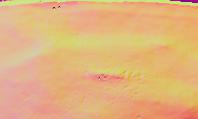}
		\end{subfigure}
		\begin{subfigure}{0.19\linewidth}
			\includegraphics[width=\linewidth]{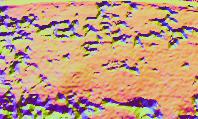}
		\end{subfigure}
		\begin{subfigure}{0.19\linewidth}
			\includegraphics[width=\linewidth]{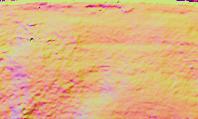}
		\end{subfigure}
	\end{minipage}
	
	\vspace{1mm}
 
    \rotatebox[origin=c]{90}{\footnotesize Eyeful Tower \textit{Workshop}}
	\begin{minipage}{0.183\textwidth}
		\begin{subfigure}{\textwidth}
			\includegraphics[width=\linewidth]{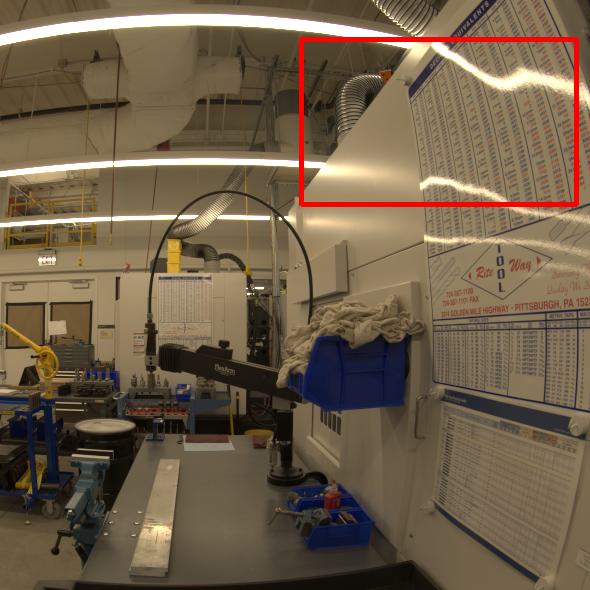}
		\end{subfigure}
	\end{minipage}
	\begin{minipage}{0.793\textwidth}
		\begin{subfigure}{0.19\linewidth}
			\includegraphics[width=\linewidth]{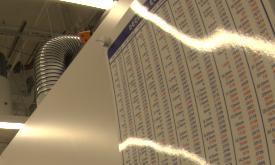}
		\end{subfigure}
		\begin{subfigure}{0.19\linewidth}
			\includegraphics[width=\linewidth]{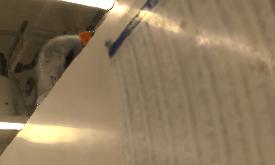}
		\end{subfigure}
		\begin{subfigure}{0.19\linewidth}
			\includegraphics[width=\linewidth]{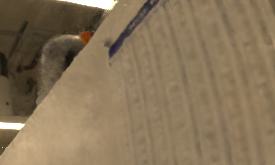}
		\end{subfigure}
		\begin{subfigure}{0.19\linewidth}
			\includegraphics[width=\linewidth]{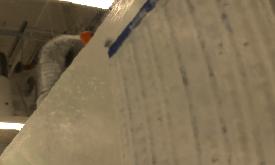}
		\end{subfigure}
		\begin{subfigure}{0.19\linewidth}
			\includegraphics[width=\linewidth]{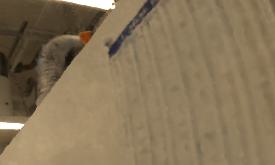}
		\end{subfigure} \\
		\begin{subfigure}{0.19\linewidth}
			\includegraphics[width=\linewidth]{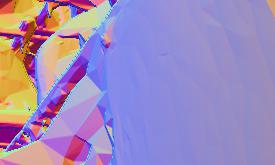}
		\end{subfigure}
		\begin{subfigure}{0.19\linewidth}
			\includegraphics[width=\linewidth]{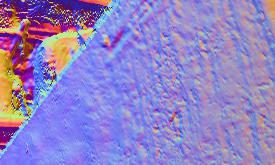}
		\end{subfigure}
		\begin{subfigure}{0.19\linewidth}
			\includegraphics[width=\linewidth]{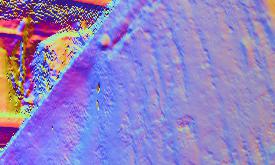}
		\end{subfigure}
		\begin{subfigure}{0.19\linewidth}
			\includegraphics[width=\linewidth]{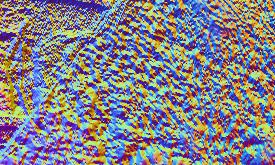}
		\end{subfigure}
		\begin{subfigure}{0.19\linewidth}
			\includegraphics[width=\linewidth]{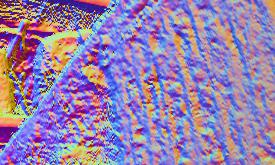}
		\end{subfigure}
	\end{minipage}

	\vspace{1mm}
 
    \rotatebox[origin=c]{90}{\footnotesize NISR \textit{LivingRoom2}}
	\begin{minipage}{0.183\textwidth}
		\begin{subfigure}{\textwidth}
			\includegraphics[width=\linewidth]{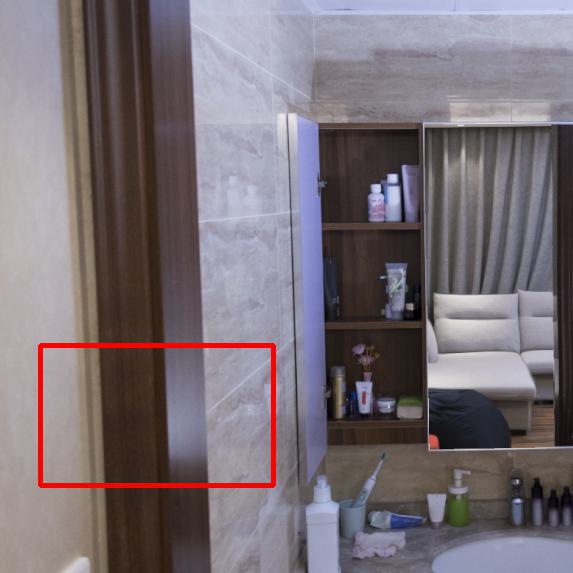}
		\end{subfigure}
	\end{minipage}
	\begin{minipage}{0.793\textwidth}
		\begin{subfigure}{0.19\linewidth}
			\includegraphics[width=\linewidth]{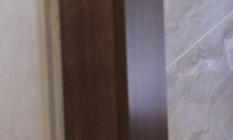}
		\end{subfigure}
		\begin{subfigure}{0.19\linewidth}
			\includegraphics[width=\linewidth]{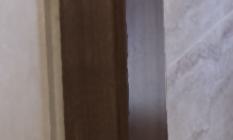}
		\end{subfigure}
		\begin{subfigure}{0.19\linewidth}
			\includegraphics[width=\linewidth]{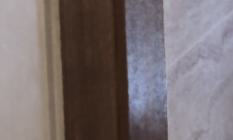}
		\end{subfigure}
		\begin{subfigure}{0.19\linewidth}
			\includegraphics[width=\linewidth]{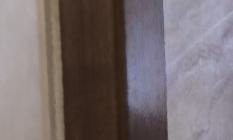}
		\end{subfigure}
		\begin{subfigure}{0.19\linewidth}
			\includegraphics[width=\linewidth]{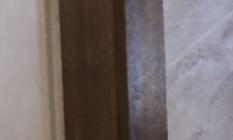}
		\end{subfigure} \\
		\begin{subfigure}{0.19\linewidth}
			\includegraphics[width=\linewidth]{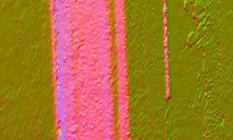}
		\end{subfigure}
		\begin{subfigure}{0.19\linewidth}
			\includegraphics[width=\linewidth]{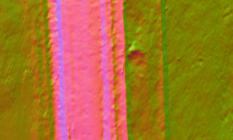}
		\end{subfigure}
		\begin{subfigure}{0.19\linewidth}
			\includegraphics[width=\linewidth]{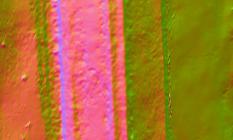}
		\end{subfigure}
		\begin{subfigure}{0.19\linewidth}
			\includegraphics[width=\linewidth]{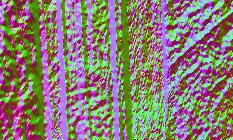}
		\end{subfigure}
		\begin{subfigure}{0.19\linewidth}
			\includegraphics[width=\linewidth]{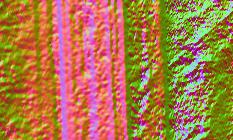}
		\end{subfigure}
	\end{minipage}
    \vspace{-2mm}
    \caption{\label{fig:supple_more_comp}%
        Comparisons of novel-view synthesis quality and normal map visualizations on the Eyeful Tower \cite{XuALGBKRPKBLZR2023} and NISR datasets \cite{WuXZBHTX2022}.
    }
    \vspace{-4mm}
\end{figure*}

\begin{figure*}
	\centering
	\begin{minipage}{0.14\textwidth}
		\centering
		\begin{subfigure}{2.34cm}
			\caption*{Test Image}
			\includegraphics[width=\linewidth]{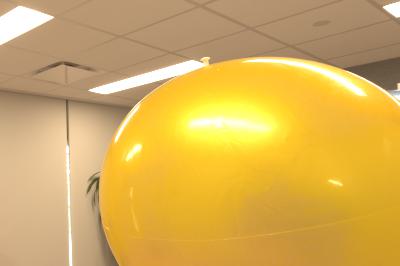}
		\end{subfigure}
	\end{minipage}
	\begin{minipage}{0.855\textwidth}
		\rotatebox{90}{\hspace{10pt} \centering\footnotesize Ours}
		\begin{subfigure}{2.34cm}
			\caption*{Final}
			\includegraphics[width=\linewidth]{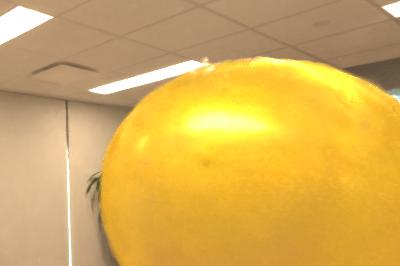}
		\end{subfigure}
		\hfill
		\begin{subfigure}{2.34cm}
			\caption*{Diffuse}
			\includegraphics[width=\linewidth]{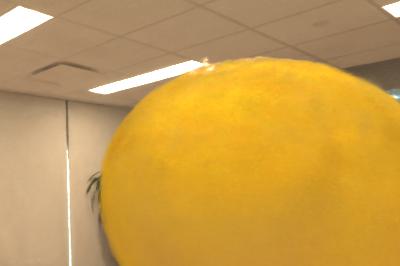}
		\end{subfigure}
		\hfill
		\begin{subfigure}{2.34cm}
			\caption*{Specular}
			\includegraphics[width=\linewidth]{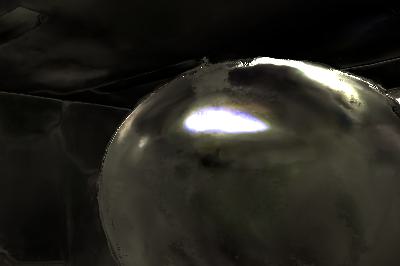}
		\end{subfigure}
		\hfill
		\begin{subfigure}{2.34cm}
			\caption*{Tint}
			\includegraphics[width=\linewidth]{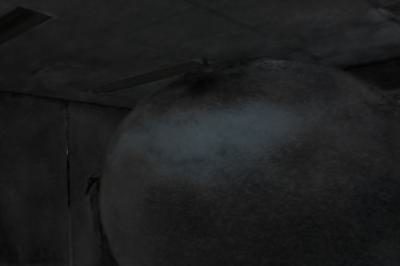}
		\end{subfigure}
		\hfill
		\begin{subfigure}{2.34cm}
			\caption*{Roughness}
			\includegraphics[width=\linewidth]{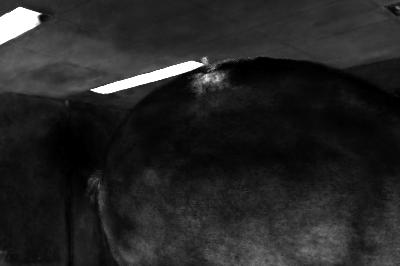}
		\end{subfigure}
		\hfill
		\begin{subfigure}{2.34cm}
			\caption*{Normal}
			\includegraphics[width=\linewidth]{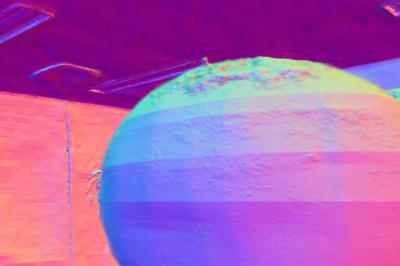}
		\end{subfigure} \\
		\rotatebox{90}{\hspace{3pt} \centering\footnotesize Ref-NeRF}
		\begin{subfigure}{2.34cm}
			\includegraphics[width=\linewidth]{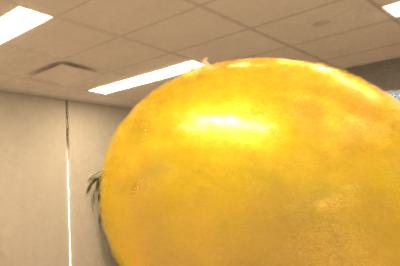}
		\end{subfigure}
		\hfill
		\begin{subfigure}{2.34cm}
			\includegraphics[width=\linewidth]{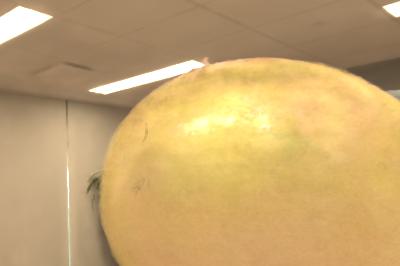}
		\end{subfigure}
		\hfill
		\begin{subfigure}{2.34cm}
			\includegraphics[width=\linewidth]{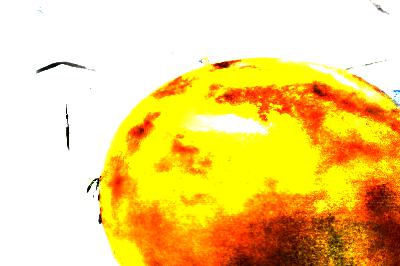}
		\end{subfigure}
		\hfill
		\begin{subfigure}{2.34cm}
			\includegraphics[width=\linewidth]{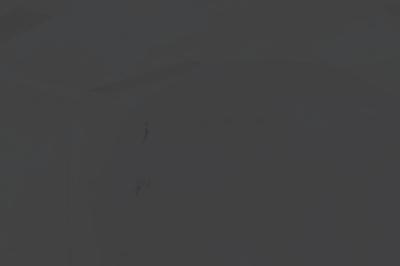}
		\end{subfigure}
		\hfill
		\begin{subfigure}{2.34cm}
			\includegraphics[width=\linewidth]{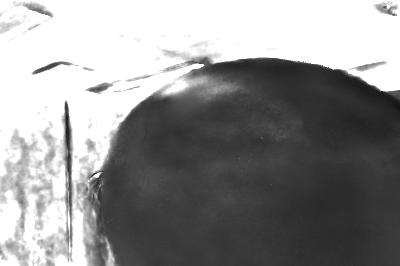}
		\end{subfigure}
		\hfill
		\begin{subfigure}{2.34cm}
			\includegraphics[width=\linewidth]{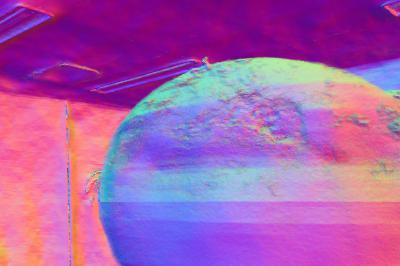}
		\end{subfigure}
	\end{minipage}

	\vspace{1mm}

	\begin{minipage}{0.14\textwidth}
		\centering
		\begin{subfigure}{2.34cm}
			\caption*{Test Image}
			\includegraphics[width=\linewidth]{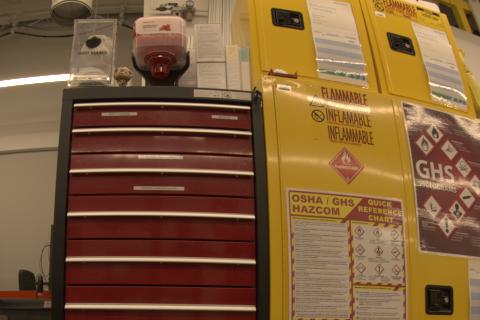}
		\end{subfigure}
	\end{minipage}
	\begin{minipage}{0.855\textwidth}
		\rotatebox{90}{\hspace{10pt} \centering\footnotesize Ours}
		\begin{subfigure}{2.34cm}
			\includegraphics[width=\linewidth]{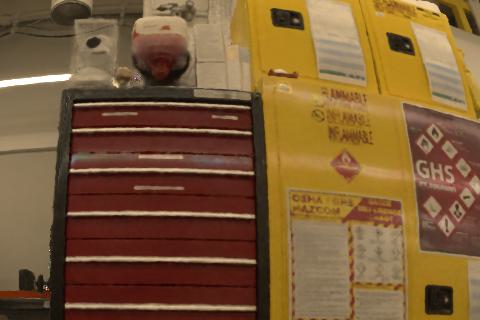}
		\end{subfigure}
		\hfill
		\begin{subfigure}{2.34cm}
			\includegraphics[width=\linewidth]{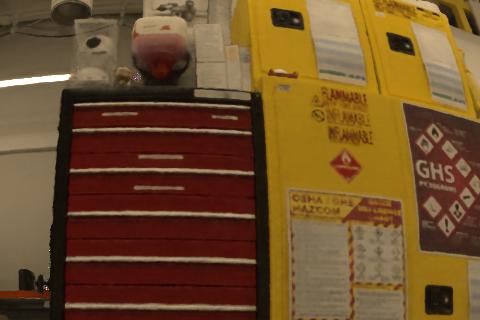}
		\end{subfigure}
		\hfill
		\begin{subfigure}{2.34cm}
			\includegraphics[width=\linewidth]{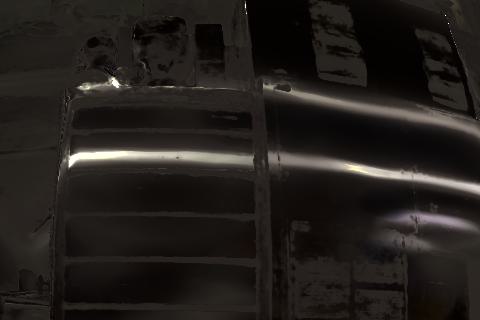}
		\end{subfigure}
		\hfill
		\begin{subfigure}{2.34cm}
			\includegraphics[width=\linewidth]{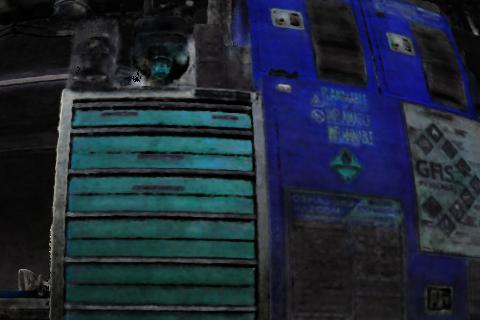}
		\end{subfigure}
		\hfill
		\begin{subfigure}{2.34cm}
			\includegraphics[width=\linewidth]{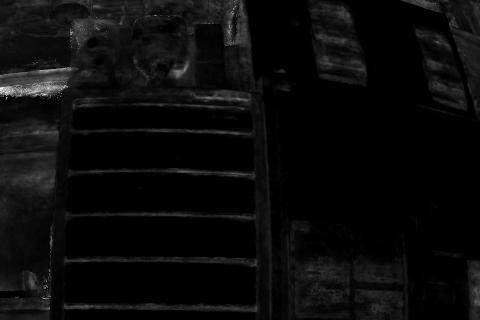}
		\end{subfigure}
		\hfill
		\begin{subfigure}{2.34cm}
			\includegraphics[width=\linewidth]{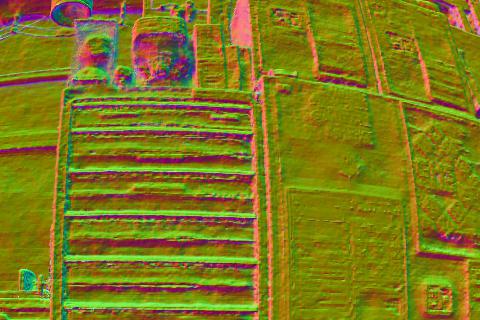}
		\end{subfigure} \\
		\rotatebox{90}{\hspace{3pt} \centering\footnotesize Ref-NeRF}
		\begin{subfigure}{2.34cm}
			\includegraphics[width=\linewidth]{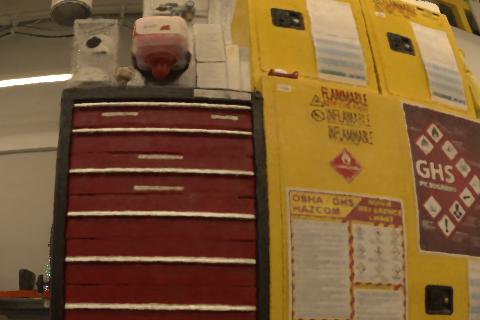}
		\end{subfigure}
		\hfill
		\begin{subfigure}{2.34cm}
			\includegraphics[width=\linewidth]{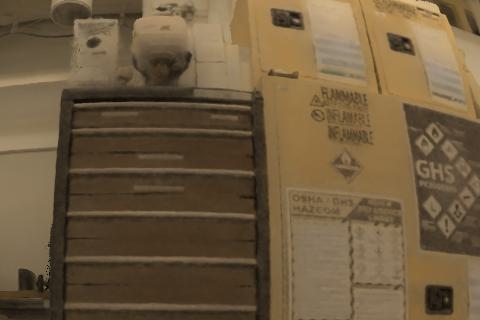}
		\end{subfigure}
		\hfill
		\begin{subfigure}{2.34cm}
			\includegraphics[width=\linewidth]{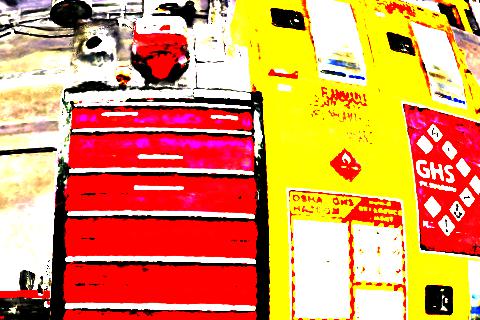}
		\end{subfigure}
		\hfill
		\begin{subfigure}{2.34cm}
			\includegraphics[width=\linewidth]{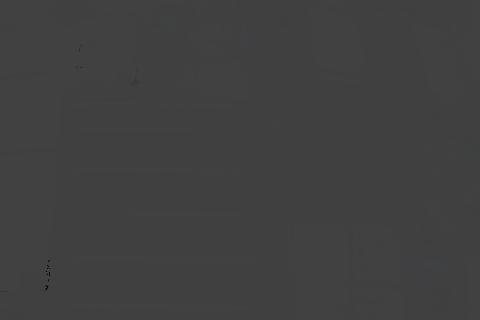}
		\end{subfigure}
		\hfill
		\begin{subfigure}{2.34cm}
			\includegraphics[width=\linewidth]{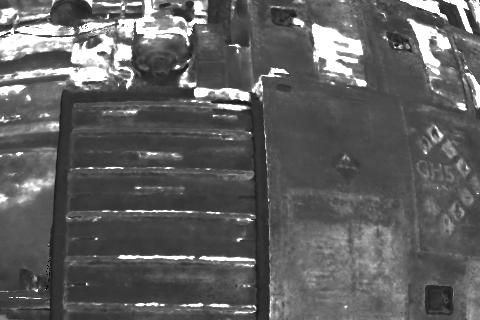}
		\end{subfigure}
		\hfill
		\begin{subfigure}{2.34cm}
			\includegraphics[width=\linewidth]{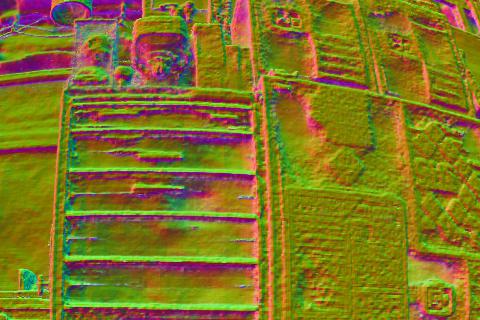}
		\end{subfigure}
	\end{minipage}
	
	\vspace{1mm}

\begin{minipage}{0.14\textwidth}
		\centering
		\begin{subfigure}{2.34cm}
			\caption*{Test Image}
			\includegraphics[width=\linewidth]{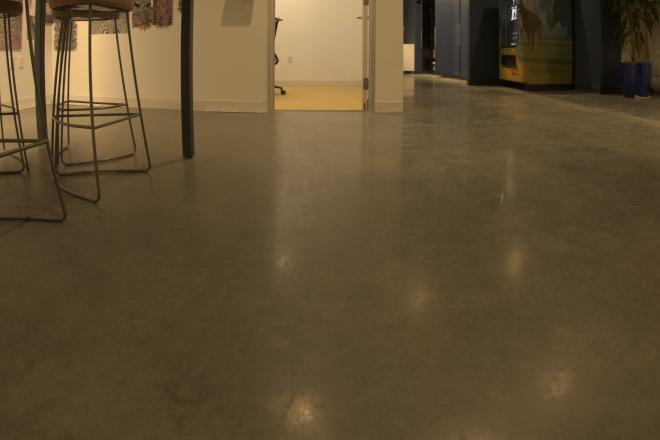}
		\end{subfigure}
	\end{minipage}
	\begin{minipage}{0.855\textwidth}
		\rotatebox{90}{\hspace{10pt} \centering\footnotesize Ours}
		\begin{subfigure}{2.34cm}
			\includegraphics[width=\linewidth]{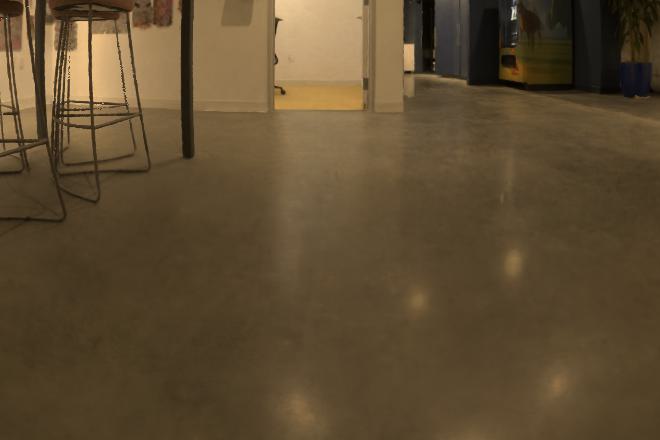}
		\end{subfigure}
		\hfill
		\begin{subfigure}{2.34cm}
			\includegraphics[width=\linewidth]{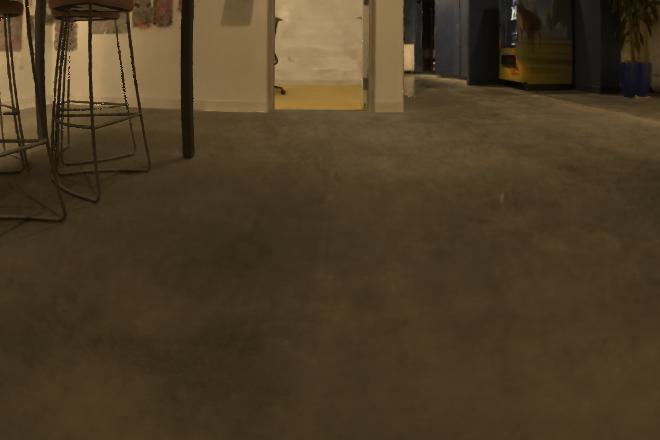}
		\end{subfigure}
		\hfill
		\begin{subfigure}{2.34cm}
			\includegraphics[width=\linewidth]{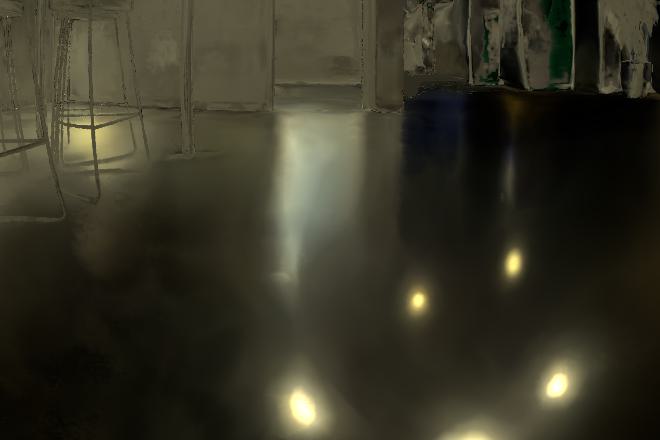}
		\end{subfigure}
		\hfill
		\begin{subfigure}{2.34cm}
			\includegraphics[width=\linewidth]{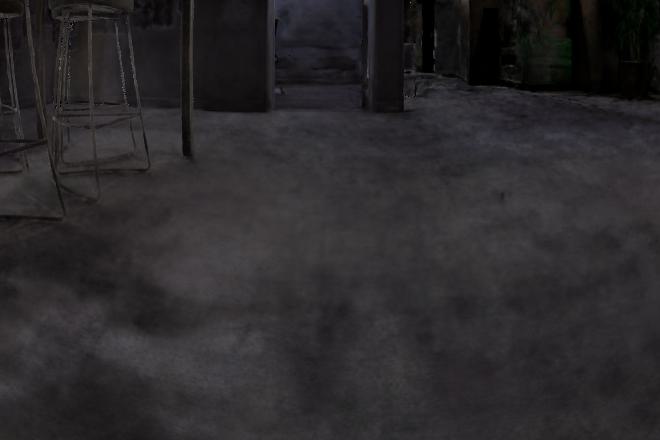}
		\end{subfigure}
		\hfill
		\begin{subfigure}{2.34cm}
			\includegraphics[width=\linewidth]{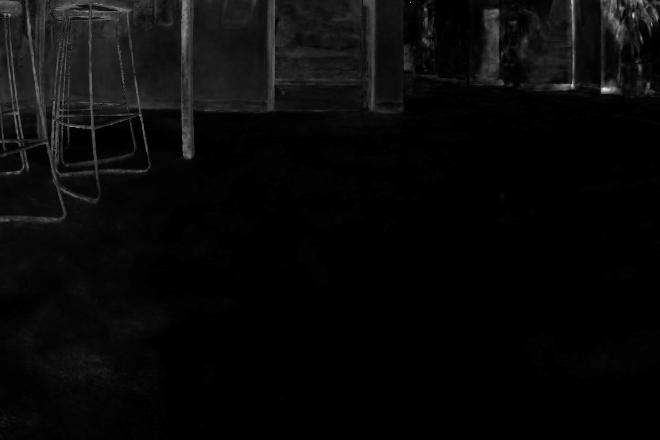}
		\end{subfigure}
		\hfill
		\begin{subfigure}{2.34cm}
			\includegraphics[width=\linewidth]{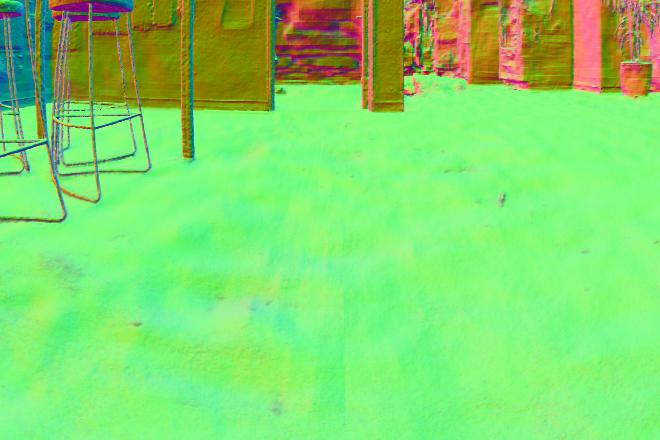}
		\end{subfigure} \\
		\rotatebox{90}{\hspace{3pt} \centering\footnotesize Ref-NeRF}
		\begin{subfigure}{2.34cm}
			\includegraphics[width=\linewidth]{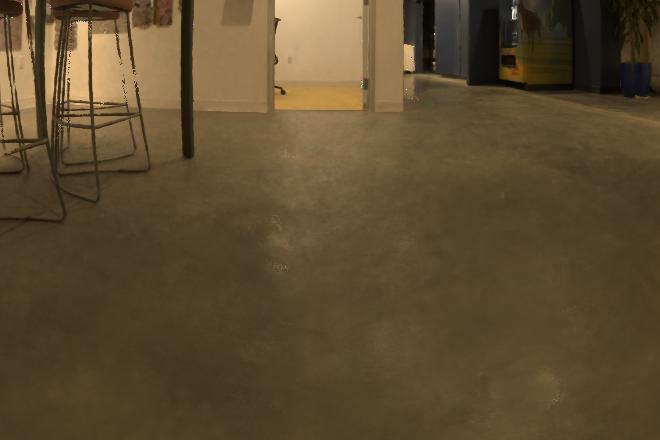}
		\end{subfigure}
		\hfill
		\begin{subfigure}{2.34cm}
			\includegraphics[width=\linewidth]{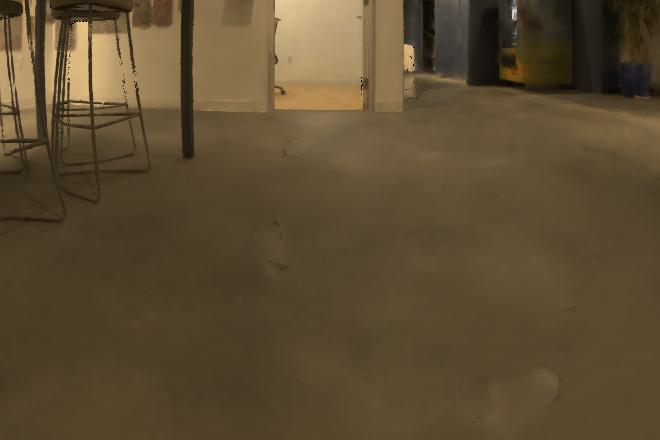}
		\end{subfigure}
		\hfill
		\begin{subfigure}{2.34cm}
			\includegraphics[width=\linewidth]{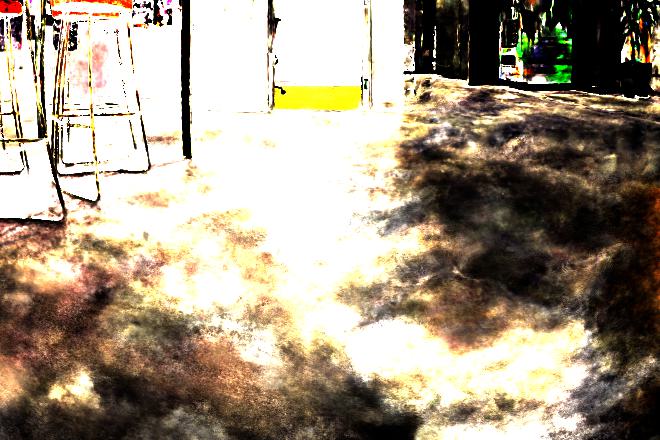}
		\end{subfigure}
		\hfill
		\begin{subfigure}{2.34cm}
			\includegraphics[width=\linewidth]{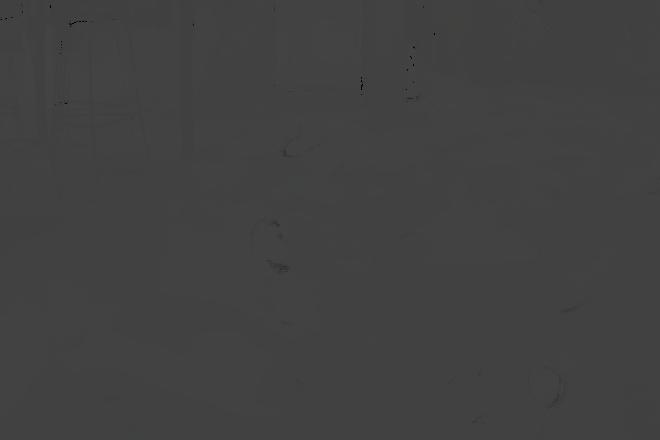}
		\end{subfigure}
		\hfill
		\begin{subfigure}{2.34cm}
			\includegraphics[width=\linewidth]{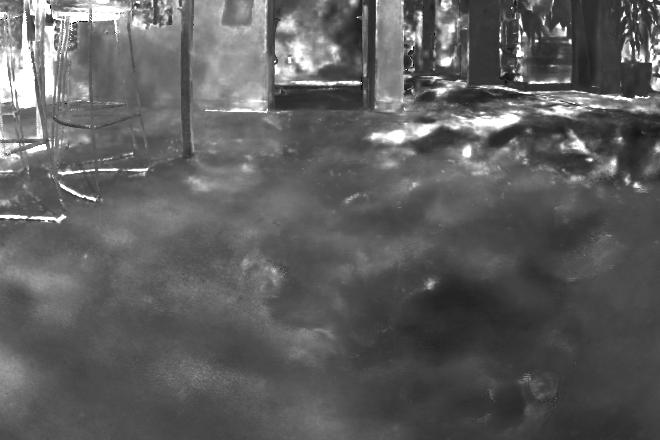}
		\end{subfigure}
		\hfill
		\begin{subfigure}{2.34cm}
			\includegraphics[width=\linewidth]{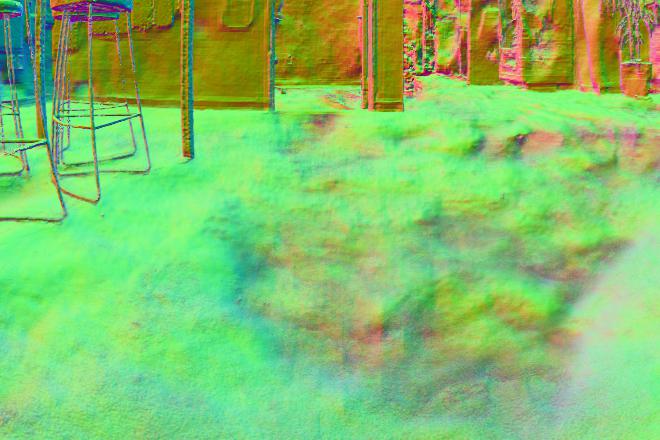}
		\end{subfigure}
	\end{minipage}
	
	\vspace{1mm}

 \begin{minipage}{0.14\textwidth}
		\centering
		\begin{subfigure}{2.34cm}
			\caption*{Test Image}
			\includegraphics[width=\linewidth]{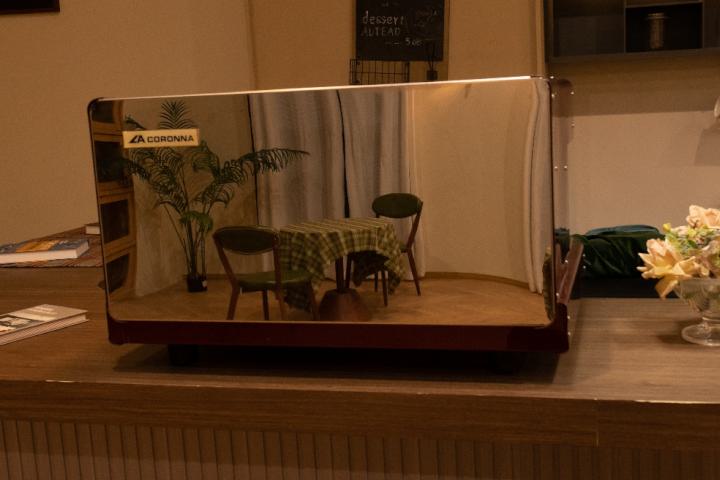}
		\end{subfigure}
	\end{minipage}
	\begin{minipage}{0.855\textwidth}
		\rotatebox{90}{\hspace{10pt} \centering\footnotesize Ours}
		\begin{subfigure}{2.34cm}
			\includegraphics[width=\linewidth]{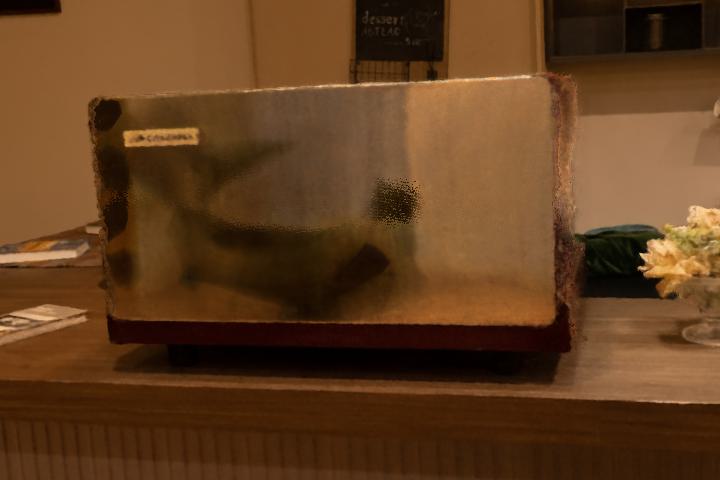}
		\end{subfigure}
		\hfill
		\begin{subfigure}{2.34cm}
			\includegraphics[width=\linewidth]{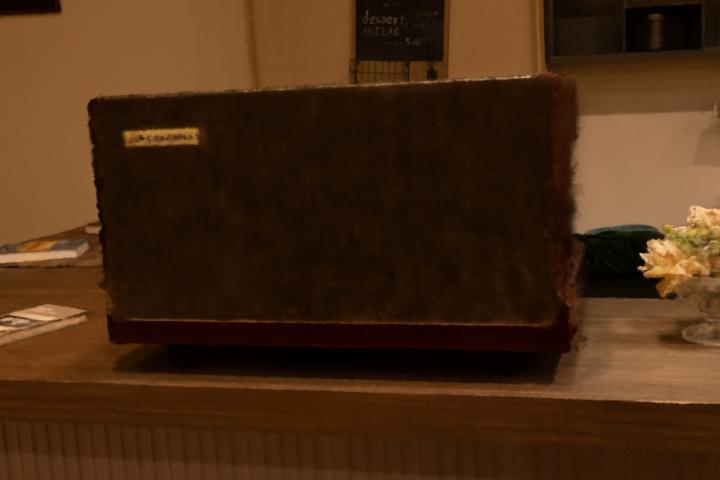}
		\end{subfigure}
		\hfill
		\begin{subfigure}{2.34cm}
			\includegraphics[width=\linewidth]{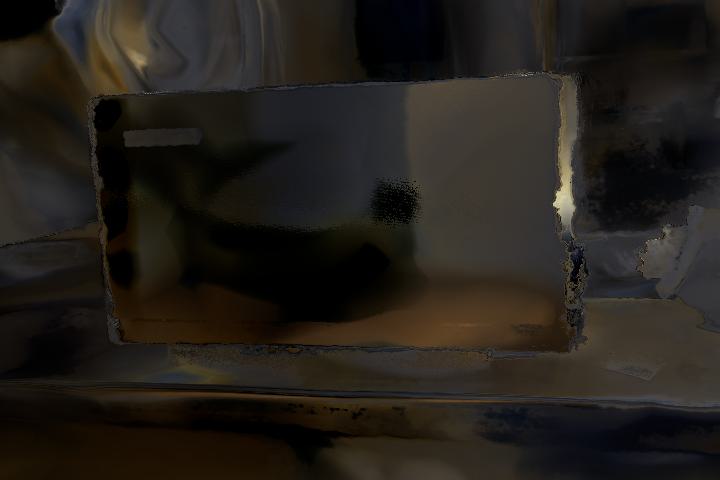}
		\end{subfigure}
		\hfill
		\begin{subfigure}{2.34cm}
			\includegraphics[width=\linewidth]{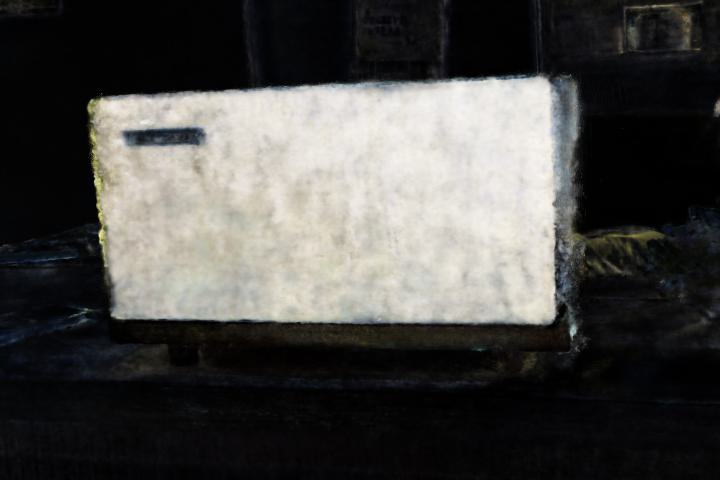}
		\end{subfigure}
		\hfill
		\begin{subfigure}{2.34cm}
			\includegraphics[width=\linewidth]{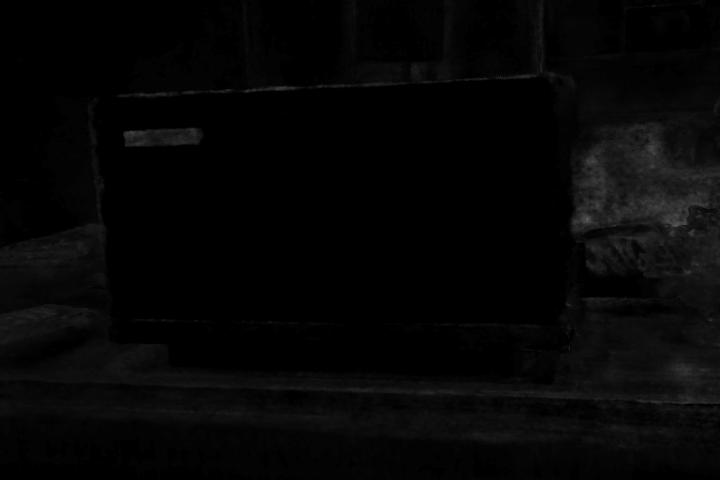}
		\end{subfigure}
		\hfill
		\begin{subfigure}{2.34cm}
			\includegraphics[width=\linewidth]{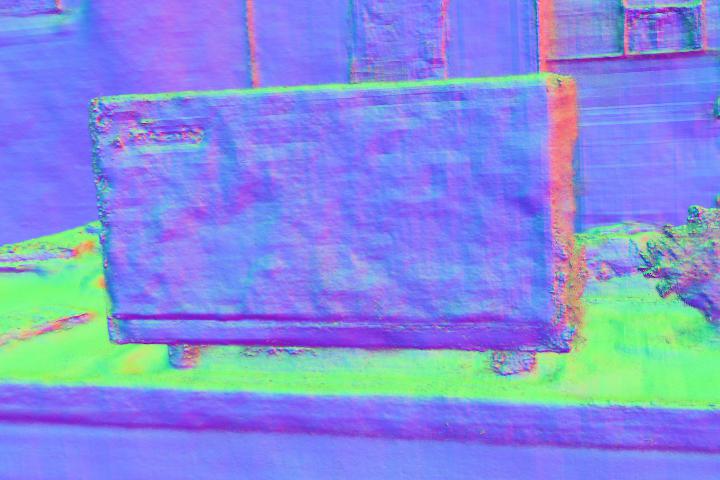}
		\end{subfigure} \\
		\rotatebox{90}{\hspace{3pt} \centering\footnotesize Ref-NeRF}
		\begin{subfigure}{2.34cm}
			\includegraphics[width=\linewidth]{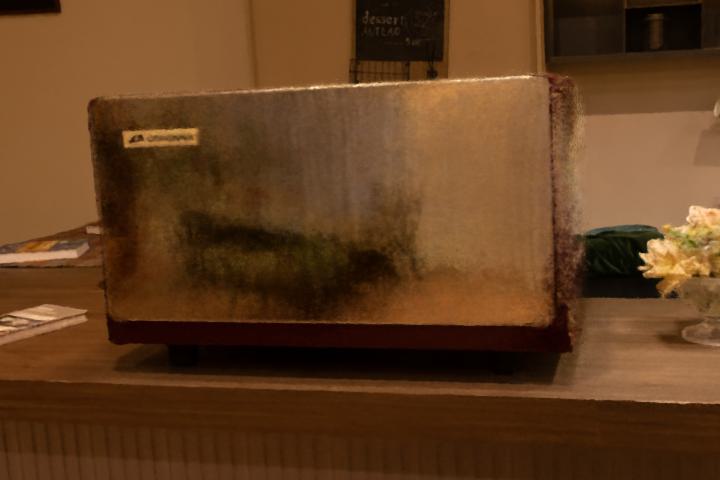}
		\end{subfigure}
		\hfill
		\begin{subfigure}{2.34cm}
			\includegraphics[width=\linewidth]{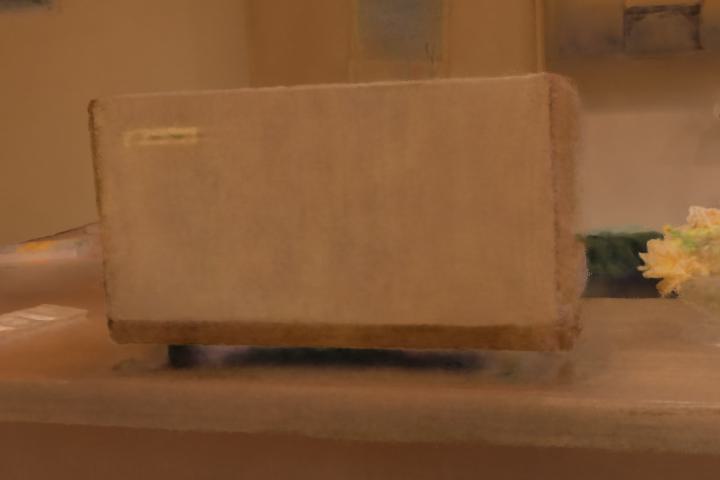}
		\end{subfigure}
		\hfill
		\begin{subfigure}{2.34cm}
			\includegraphics[width=\linewidth]{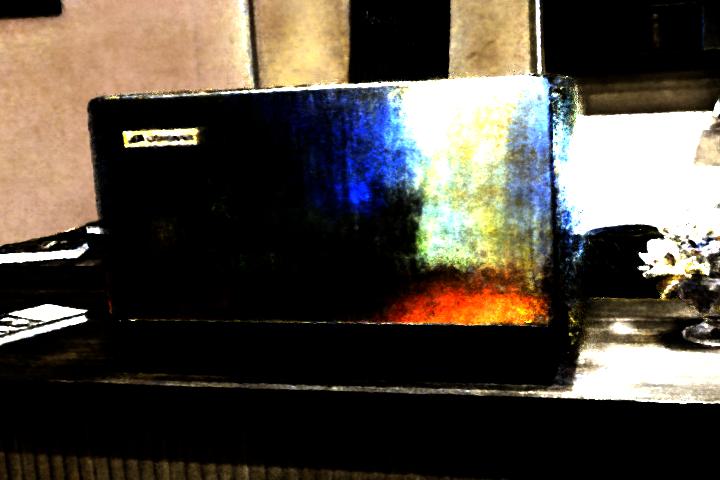}
		\end{subfigure}
		\hfill
		\begin{subfigure}{2.34cm}
			\includegraphics[width=\linewidth]{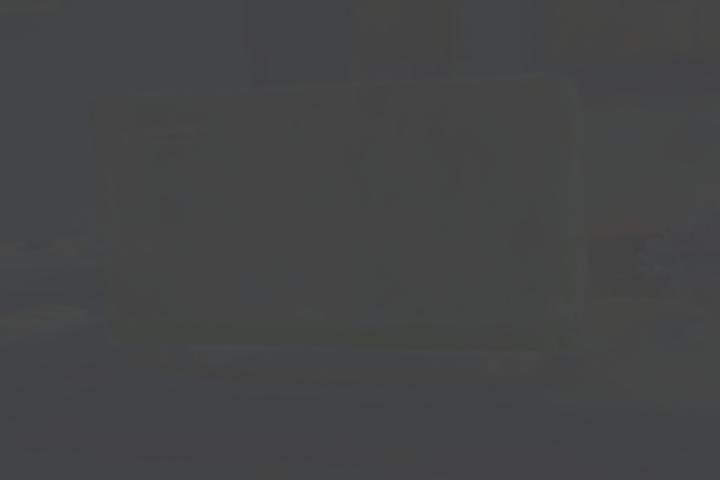}
		\end{subfigure}
		\hfill
		\begin{subfigure}{2.34cm}
			\includegraphics[width=\linewidth]{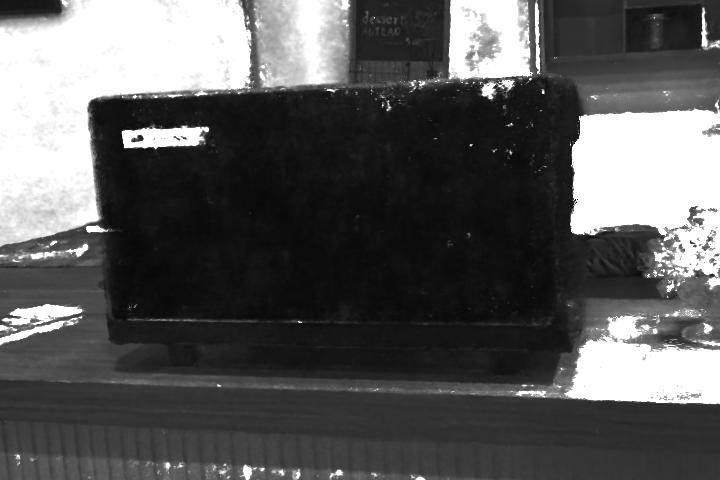}
		\end{subfigure}
		\hfill
		\begin{subfigure}{2.34cm}
			\includegraphics[width=\linewidth]{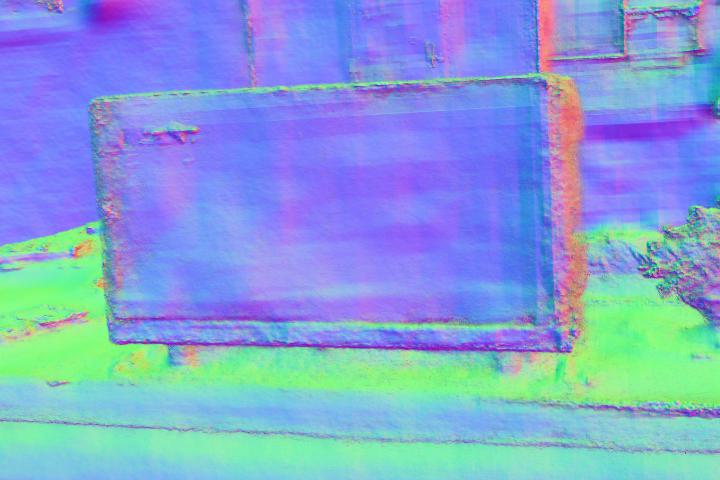}
		\end{subfigure}
	\end{minipage}
	
	\vspace{1mm}
	
 \begin{minipage}{0.14\textwidth}
		\centering
		\begin{subfigure}{2.34cm}
			\caption*{Test Image}
			\includegraphics[width=\linewidth]{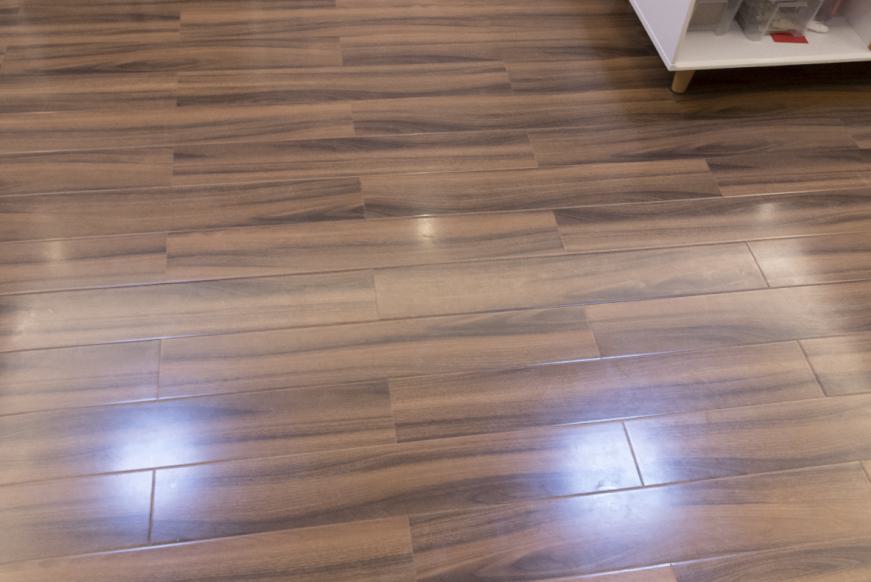}
		\end{subfigure}
	\end{minipage}
	\begin{minipage}{0.855\textwidth}
		\rotatebox{90}{\hspace{10pt} \centering\footnotesize Ours}
		\begin{subfigure}{2.34cm}
			\includegraphics[width=\linewidth]{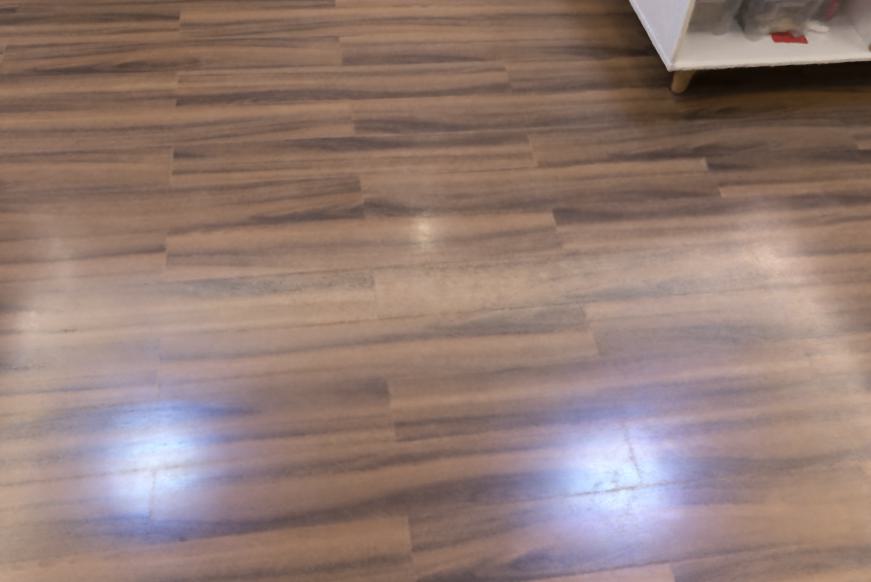}
		\end{subfigure}
		\hfill
		\begin{subfigure}{2.34cm}
			\includegraphics[width=\linewidth]{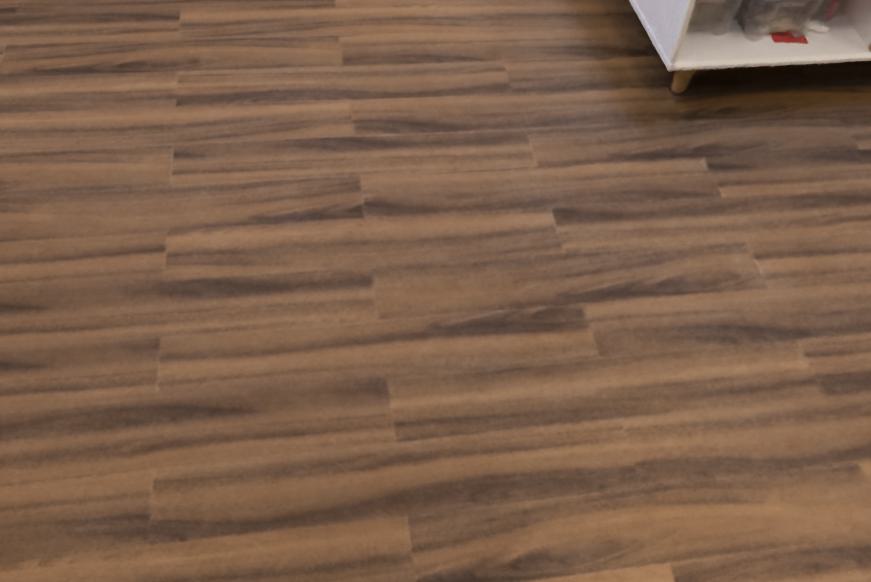}
		\end{subfigure}
		\hfill
		\begin{subfigure}{2.34cm}
			\includegraphics[width=\linewidth]{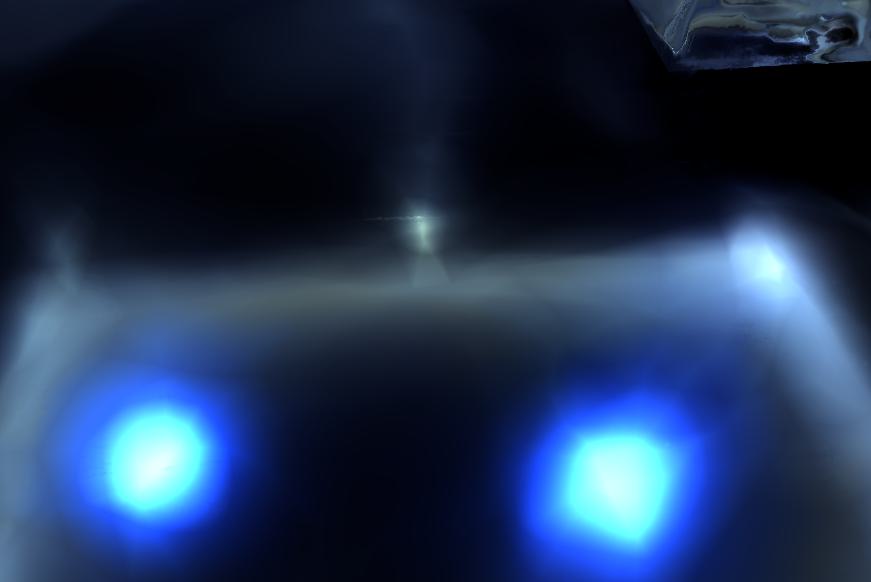}
		\end{subfigure}
		\hfill
		\begin{subfigure}{2.34cm}
			\includegraphics[width=\linewidth]{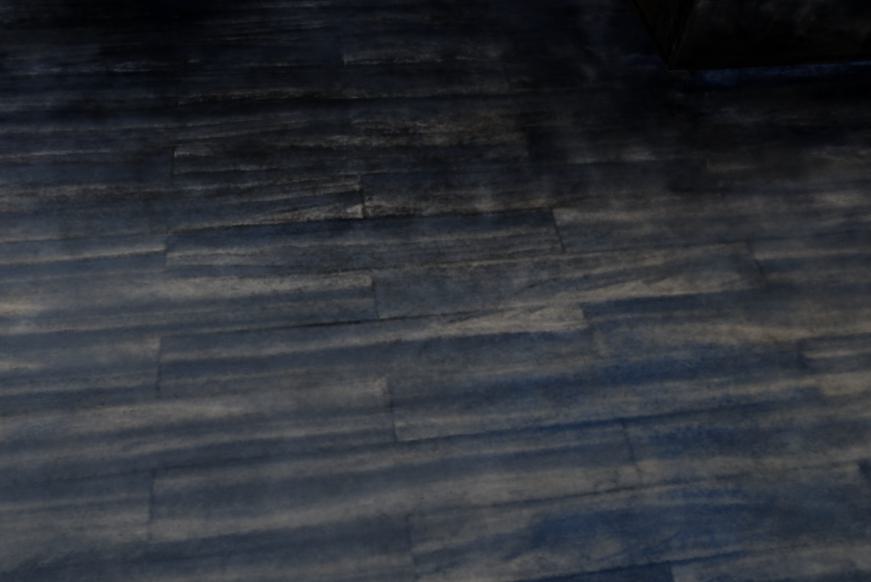}
		\end{subfigure}
		\hfill
		\begin{subfigure}{2.34cm}
			\includegraphics[width=\linewidth]{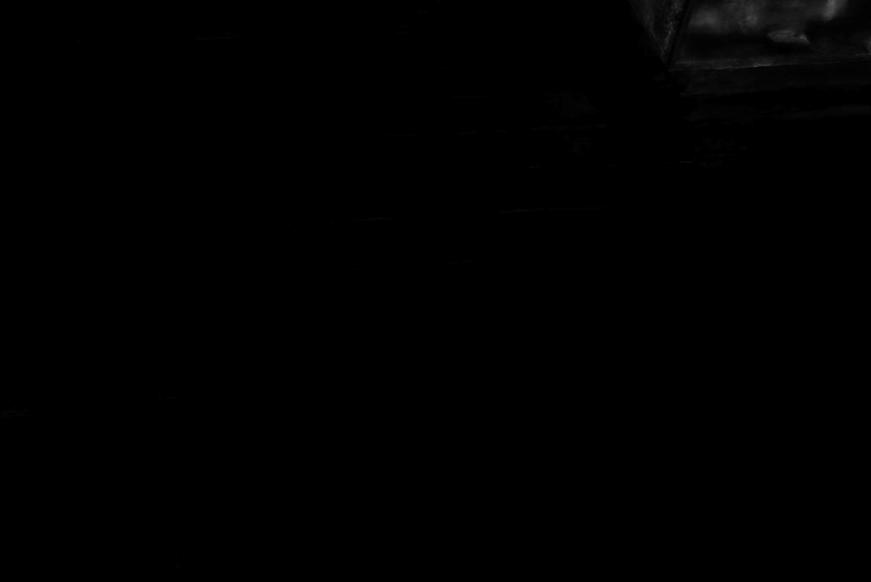}
		\end{subfigure}
		\hfill
		\begin{subfigure}{2.34cm}
			\includegraphics[width=\linewidth]{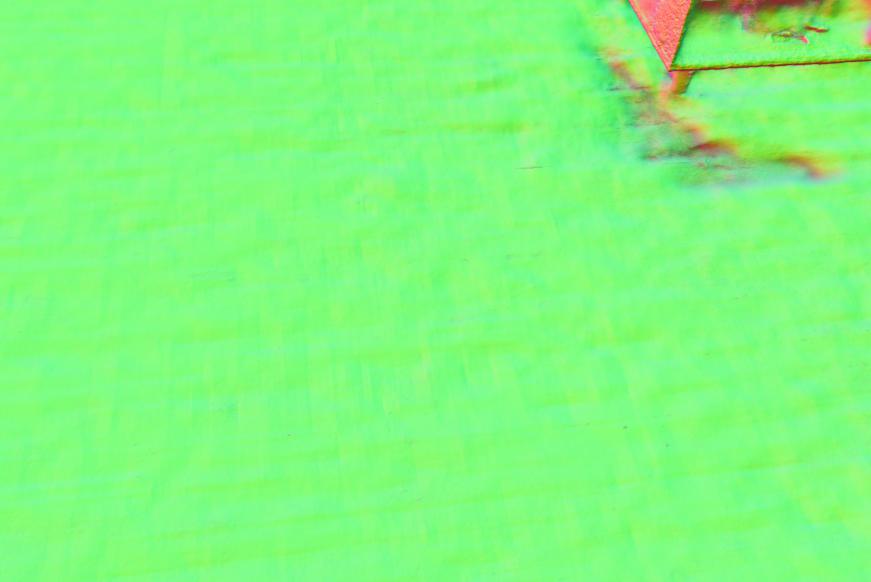}
		\end{subfigure} \\
		\rotatebox{90}{\hspace{3pt} \centering\footnotesize Ref-NeRF}
		\begin{subfigure}{2.34cm}
			\includegraphics[width=\linewidth]{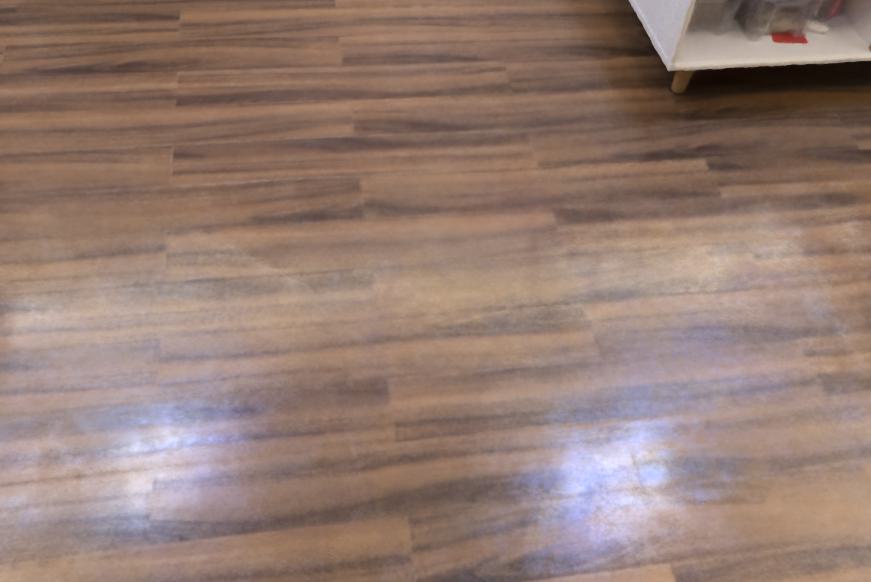}
		\end{subfigure}
		\hfill
		\begin{subfigure}{2.34cm}
			\includegraphics[width=\linewidth]{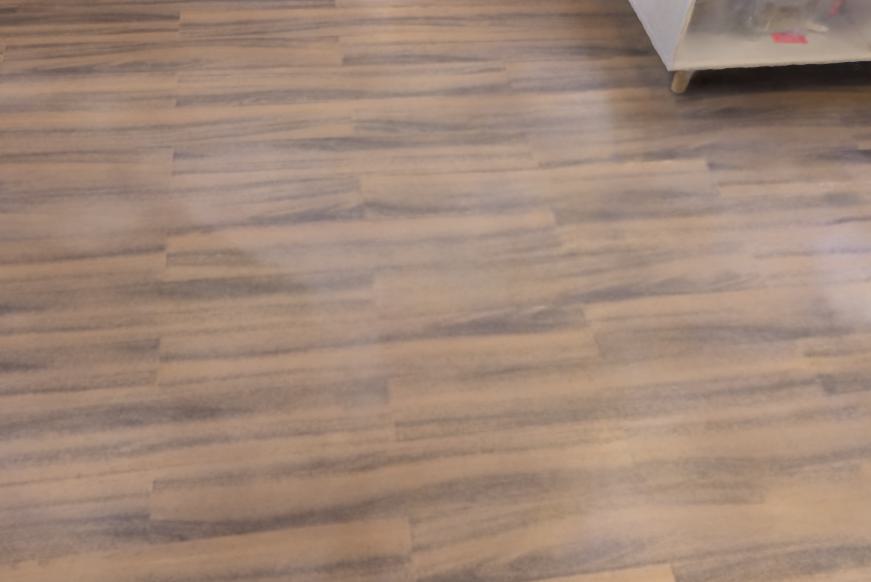}
		\end{subfigure}
		\hfill
		\begin{subfigure}{2.34cm}
			\includegraphics[width=\linewidth]{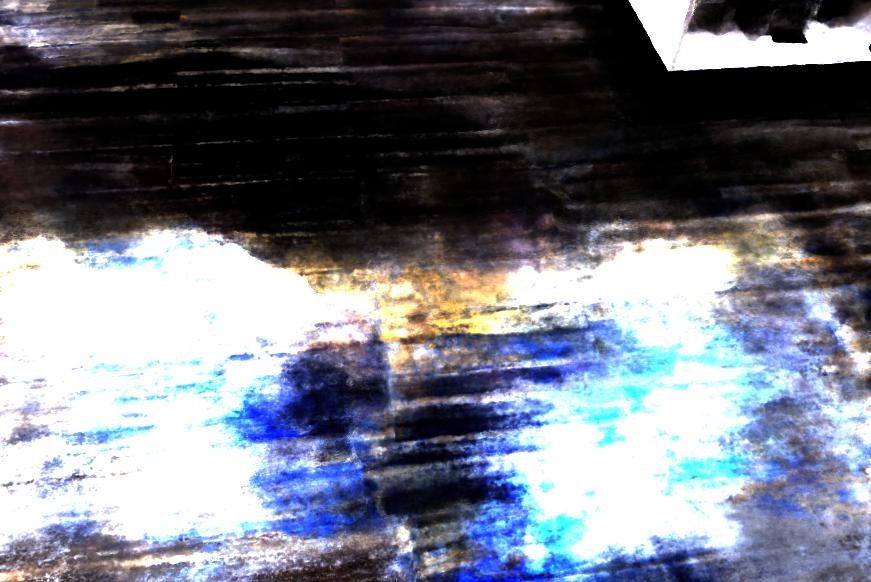}
		\end{subfigure}
		\hfill
		\begin{subfigure}{2.34cm}
			\includegraphics[width=\linewidth]{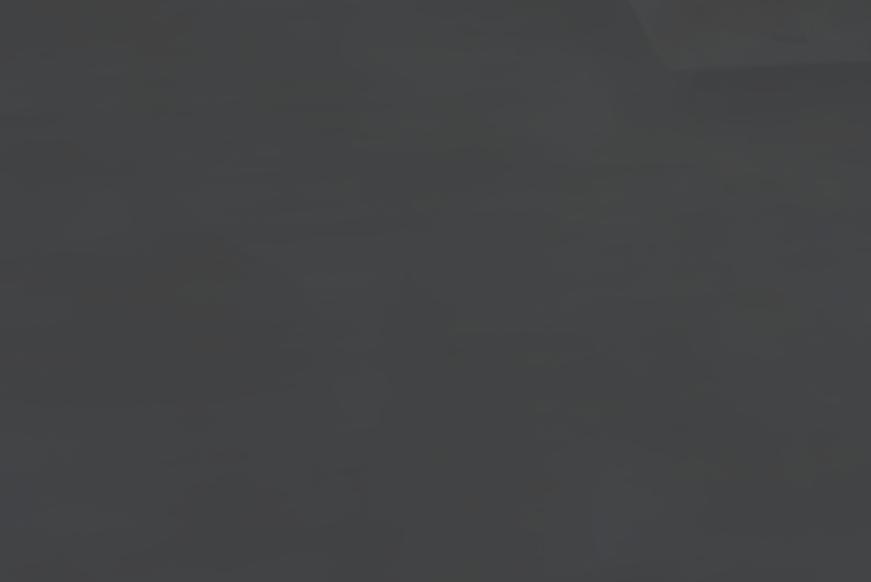}
		\end{subfigure}
		\hfill
		\begin{subfigure}{2.34cm}
			\includegraphics[width=\linewidth]{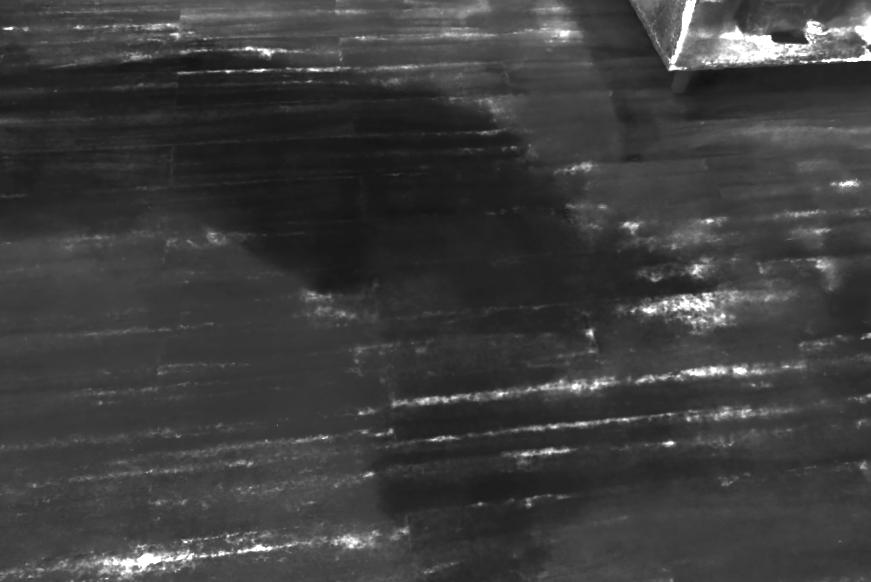}
		\end{subfigure}
		\hfill
		\begin{subfigure}{2.34cm}
			\includegraphics[width=\linewidth]{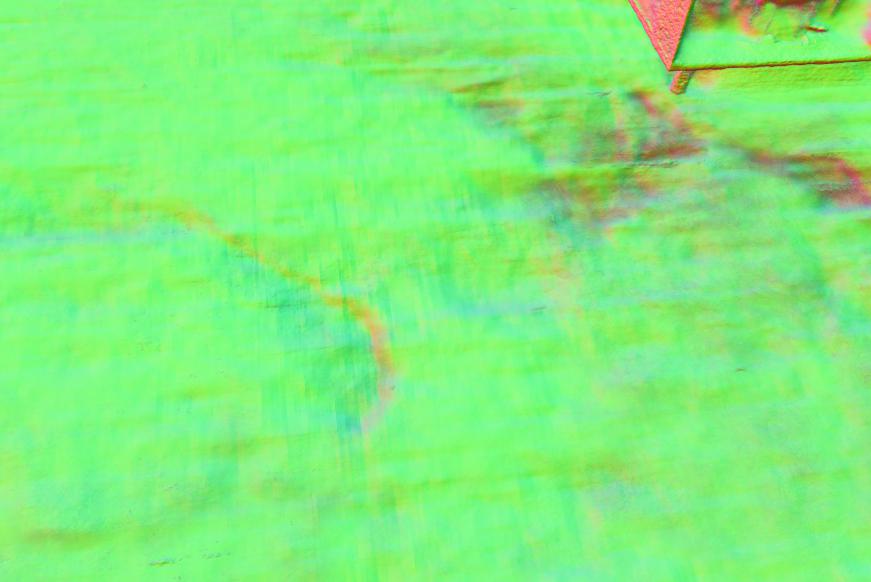}
		\end{subfigure}
	\end{minipage}
    \vspace{-2mm}
    \caption{\label{fig:supple_more_decomp}%
        Additional results for intermediate component visualizations of our approach compared to Ref-NeRF \cite{VerbiHMZBS2022}  on the Eyeful Tower \cite{XuALGBKRPKBLZR2023} and NISR datasets \cite{WuXZBHTX2022}.
        Our approach produces more accurate decompositions and normal maps.
    }
    \vspace{-4mm}
\end{figure*}

We show additional comparisons and decomposition results in \cref{fig:supple_more_comp} and \cref{fig:supple_more_decomp}.
Our method achieves the best visual quality as well as the predicted normal quality for specular reflections.

\begin{table}[t]
\caption{
\label{tab:supple_synthetic_comp}
Quantitative comparisons of novel-view synthesis and geometry quality on the FIPT synthetic dataset.
Our method achieves the best view synthesis quality, and is most accurate in terms of geometry.
We highlight the best numbers in \textbf{bold}.
}
\vspace{-1mm}
\begin{tabularx}{\linewidth}{p{2cm}|YYYY}
\toprule
\small Methods & \small PSNR$\uparrow$   & \small SSIM$\uparrow$   & \small LPIPS$\downarrow$ & \small MAE$^\circ$$\downarrow$   \\
\midrule
\small Ours                           & \textbf{32.043} & \textbf{0.8657} & 0.1266 & \textbf{16.09} \\
\small NeRF~\cite{MildeSTBRN2020}     & 31.621 & 0.8586 & 0.1325 & 34.16 \\
\small Ref-NeRF~\cite{VerbiHMZBS2022} & 31.952 & 0.8650 & \textbf{0.1250} & 18.76 \\
\small MS-NeRF~\cite{YinQCR2023}      & 31.441 & 0.8534 & 0.1345 & 42.19 \\
\small FIPT~\cite{RT_FIPT}            & 28.322 & 0.6922 & 0.1379 & 0$^\dagger$ \\
\bottomrule
\end{tabularx}\\[3pt]
\centering\small$^\dagger$Note that FIPT uses the ground-truth geometry.
\end{table}

\end{document}